\documentclass[10pt,twocolumn,letterpaper]{article}

\usepackage[pagenumbers]{cvpr} % To force page numbers, e.g. for an arXiv version

\usepackage[utf8]{inputenc} % allow utf-8 input
\usepackage[T1]{fontenc}    % use 8-bit T1 fonts
\usepackage{url}            % simple URL typesetting
\usepackage{booktabs}       % professional-quality tables
\usepackage{amsfonts}       % blackboard math symbols
\usepackage{nicefrac}       % compact symbols for 1/2, etc.
\usepackage{microtype}      % microtypography
\usepackage{xcolor}         % colors
\usepackage{graphicx}
\usepackage{xcolor}
\usepackage{amssymb}
\usepackage{mathrsfs}
\usepackage{comment}
\usepackage{multirow}
\usepackage{soul}
\usepackage[export]{adjustbox}
\usepackage[accsupp]{axessibility}
\usepackage[scr=boondoxo]{mathalfa}
\usepackage{amsfonts}
\usepackage{amsmath}
\usepackage{comment}

\newcommand{\R}{\mathbb{R}}

\newcommand{\jac}[1]{\mbox{\text{Jac}}\!\left( #1\right)}
\newcommand{\hess}[1]{\mbox{\textbf{Hess}}\left( #1\right)}

\newcommand\norm[1]{\left\lVert#1\right\rVert}
\newcommand{\gt}[1]{\textbf{#1}}

\usepackage{makecell}

\usepackage{graphicx,wrapfig}
\usepackage[nopar]{lipsum}

\definecolor{cvprblue}{rgb}{0.21,0.49,0.74}
\usepackage[pagebackref,breaklinks,colorlinks,citecolor=cvprblue]{hyperref}

\def\paperID{5084} % *** Enter the Paper ID here
\def\confName{CVPR}
\def\confYear{2024}

\title{Neural Implicit Morphing of Face Images}

\author{\normalsize
  Guilherme Schardong$^{1,}$\thanks{These authors contributed equally to this work.\\Project page: \scriptsize{\url{https://schardong.github.io/ifmorph}}}
  \quad
  \normalsize Tiago Novello$^{2,*}$
  \quad
  \normalsize Hallison Paz$^2$
  \quad
  \normalsize Iurii Medvedev$^1$
  \quad
  \normalsize Vinícius da Silva$^3$ \\
  \quad
  \normalsize Luiz Velho$^2$
  \quad
  \normalsize Nuno Gonçalves$^{1,4}$ \\ \\
  $^1$ \small{Institute of Systems and Robotics, University of Coimbra} \\
  $^2$ \small{Institute of Pure and Applied Mathematics} \\
  $^3$ \small{Tecgraf, Pontifical Catholic University of Rio de Janeiro} \\
  $^4$ \small{Portuguese Mint and Official Printing Office}
}

\begin{document}

\maketitle

\begin{abstract}
  \vspace{-0.45cm}
  \textbf{Face morphing} is a problem in computer graphics with numerous artistic and forensic applications. It is challenging due to variations in pose, lighting, gender, and ethnicity. This task consists of a \textbf{warping} for feature alignment and a \textbf{blending} for a seamless transition between the warped images.
  We propose to leverage \textbf{coord-based neural networks} to represent such warpings and blendings of face images. During training, we exploit the smoothness and flexibility of such networks by combining energy functionals employed in classical approaches without discretizations. Additionally, our method is \textbf{time-dependent}, allowing a continuous warping/blending of the images.
  During morphing inference, we need both direct and inverse transformations of the time-dependent warping. The first (second) is responsible for warping the target (source) image into the source (target) image. Our neural warping stores those maps in a single network dismissing the need for inverting them.
  The results of our experiments indicate that our method is competitive with both classical and generative models under the lens of image quality and face-morphing detectors. Aesthetically, the resulting images present a seamless blending of diverse faces not yet usual in the literature.
\end{abstract}

\vspace{-0.4cm}
\section{Introduction}
\vspace{-0.1cm}

\textit{Image warping} is a continuous transformation mapping points of the image support to points in a second domain. The process of warping an image has applications ranging from correcting image distortions caused by lens or sensor imperfections~\cite{glasbey1998review} to creating distortions for artistic/scientific purposes~\cite{carroll2010image}.
Warping finds a special application in creating \textit{image morphings}~\cite{gomes1999warping}, where it is used to align corresponding features. By gradually aligning the image features using the warping, we obtain a smooth transition between them.

We assume the warpings to be parameterized by smooth maps. Besides obtaining smooth transitions,
this allows us to use its derivatives to constrain the deformation, such as approximating it as a minimum of a \textit{variational problem}.
Feature alignment can be specified using \textit{landmarks} to establish correlations between two images.

In this work, we use \textit{coord-based neural networks}, which we call \textit{neural warpings}, to parameterize image warpings.
This approach enables us to calculate the derivatives in closed form, eliminating the need for discretization. We also employ a time parameter, to represent smooth transitions.
By incorporating the derivatives into the loss function, we can regularize the network and easily add constraints by summing additional terms.
To train a neural warping, we propose a \textit{loss function} consisting of two main terms. First, a \textit{data constraint} ensures that the warping fits the given keypoint correspondences. Second, we \textit{regularize} the neural warping using the \textit{thin-plate} energy to minimize distortions.

We use neural warping to model \textit{time-dependent} morphings of face images, thus aligning the image features over time. Afterward, we explore the flexibility of coord-based neural networks to define three blending techniques.
First, we blend the aligned image warpings in the \textit{signal domain} using point-wise interpolation.
Second, we propose to blend the image warpings in the \textit{gradient-domain} of the signals. For this, we introduce another neural network to represent the morphing and train it to satisfy the corresponding variational problem.
If the target faces have different semantics, we cannot adequately blend the warped images in the signal/gradient domain; therefore, we propose a third option: blending using generative methods.
In other words, we propose to use a \textit{generative mixing}: we embed the image warpings in a \textit{latent space} of some generative model, then we interpolate the resulting embedding and project it back to the image space.
We present experiments using Diffusion Auto-encoders (diffAE)~\cite{preechakul:2022}.

Our contributions can be summarized as follows:

\begin{itemize}
    \item The introduction of a time-dependent \textbf{neural warping} which encodes in a single network the \textit{direct} and \textit{inverse} transformations needed to align two images along time.
    We use the warping to transport the images and their derivatives from the initial states to intermediate times.

    \item The neural network is \textbf{smooth}, both in space and time, enabling the use of its derivatives in the loss function. We~exploit it to define an implicit regularization using the \textit{thin-plate} energy which penalizes distortions.
    Thus, the landmarks follow a path that minimizes this energy instead of a straight line, as in classical~approaches.

    \item The neural warping model is \textbf{compact}. We achieved accurate warping using a MLP composed of a single hidden layer with $128$ neurons, although our ablation studies indicate that smaller networks would work for specific~cases.

    \item We blend the resulting aligned image warpings to define a time-dependent \textbf{morphing}, distinguishing it from current methods that focus on a single blend.
    For the case of blending in the gradient-domain, we use another neural network (\textbf{neural morphing}).
    For the \textbf{generative morphing}, we embed the warpings in a latent space, interpolate the resulting curves, and project it back to image space.
\end{itemize}

\vspace{-0.3cm}
\section{Related Works}
\label{sec:related-works}
\vspace{-0.2cm}

The first algorithms for face morphing were simple \textit{cross-dissolves}, i.e., pixel interpolation between target images~\cite{wolberg1998image}. However, the resulting morphings are substandard unless the images are aligned, resulting in artifacts. To overcome~this, \textit{mesh-based} alignment was used before the interpolation stage, shifting the complexity to the image alignment. \citet{beier1992feature} further refined the process using line correspondences and an interface to align them.
\citet{liao2014automating} exploited halfway domains, \textit{thin-plate} splines, and \textit{structural similarity} to create a discrete vector field to warp the images.

The above morphing approaches are landmark-based, as is ours. Recently, generative methods, such as StyleGANs~\cite{karras:2019:stylegan,karras:2020:stylegan,karras:2021:stylegan} and diffAE~\cite{preechakul:2022}, have also been used to interpolate between faces.
In contrast to these methods, ours is \textit{smooth} in both time and space, as we have a differentiable curve tracking the path of each image point during warping. Moreover, our approach exploits the recent \textit{implicit neural representations}, which employ coord-based neural networks~\cite{sitzmann2020implicit} to parameterize the images. Hence, we eliminate the need for interpolation and image resampling. This approach has also been used in the context of generative models~\cite{anokhin2021image} and multiresolution image representation~\cite{paz2023mr}.

Furthermore, by implicitly representing the images, we obtain their \textit{derivatives in closed form} through automatic differentiation, which is not possible with previous landmark and generative approaches. This allows efficient use of the gradient during the training/analysis. Moreover, composing the warping and images results in the warped images with gradients given by the product of the warping Jacobian and the image gradient.

An important step in our warping is the incorporation of the time variable as input of the neural warping. Combined with the above advantages, this enables the creation of continuous, smooth, and compact warpings.
This also allows us to constrain the landmark paths over time by minimizing distortions, unlike classical methods.

Regarding StyleGANs and diffusion models, StyleGANs create a latent space of images. Thus, the blending between two faces is an interpolation of the corresponding projected codes in the latent space. It produces high-quality images, although their embedding is not necessarily invertible. Therefore there is no guarantee that the blendings will be strictly of the desired faces~\cite{preechakul:2022}.

\noindent On the other hand, diffAE uses a learnable encoder to discover the high-level semantics of the image and \textit{denoising the implicit diffusion model}~\cite{song2020denoising} to decode and model stochastic variations.
Unlike StyleGANs that depend on error-prone inversion, diffAE encodes the image without an additional optimization step. The outputs of the target images are close to the originals, which is desirable for blending.

Additionally, StyleGANs may not satisfy the property of blending the target faces over time since features of other faces (from the training dataset) can appear in the intermediate frames (Fig~\ref{fig:morphing_comparisons}).
We note no such problem using the generative blending of diffAE.
That is why we use it as an example of neural blending in our framework.

Note that generative models do not align image features over time, as they do not model any warping of the image domains. Instead, they perform a \textit{generative blending} between the images.
Furthermore, such models
rely on latent code interpolations, and while they can blend the target images, they lack temporal coherence (see the video in the supp. mat.).

Also, these models consider the face images to be aligned by placing the eyes and mouths at fixed locations in the image support. Thus restricting face interpolation to a specific case, where eyes and mouths are fixed over~time.

Our morphing approach does not suffer from the said issues, since it disentangles the warping from the blending, thus allowing for different blendings, such as Poisson image blending and generative blending. For instance, the output of our neural warping can serve as input for a generative blending, enabling faces in different positions, ensuring temporal coherence, and tracking the path of each point in the image support over time (see Fig~\ref{fig:roto-translations} and the video in supp. mat.).

Morphing enables the creation of synthetic faces remarkably similar to real ones,  known as ``face-morphing attack''. These techniques have captured the attention of the biometrics community, resulting in a body of works dedicated to detecting such attacks~\cite{ferrara2014magic,morph_bench_1}. Our method has the potential to generate new datasets, enhancing the effectiveness of these detection systems.
In biometrics, the production and identification of morphed images are primarily concerned with images that comply with the International Civil Aviation Organization (ICAO) standards~\cite{ICAO_requirements_1, facing_2}. Morphing can create images that merge the biometric identifiers of multiple individuals, resulting in a facial image that could match several people. Such images in official identification documents pose a significant threat, as they undermine the fundamental principle of biometric verification: one document should correspond to an unique identity.

\section{Methodology}
% \vspace{-0.18cm}
\subsection{Background and Notation}
% \vspace{-0.18cm}
We represent an \textit{image} by a function $\text{I}:\Omega\subset \R^2\to \mathcal{C}$, where $\Omega$ is the image \textit{support} and $\mathcal{C}$ is the \textit{color space}, and parameterize it using a (coord-based) neural network $\text{I}_\theta:\R^2\to \mathcal{C}$ with parameters $\theta$.
To train the \textit{neural image} $\text{I}_\theta$ such that it approximates $\text{I}$, we can optimize $\int_\Omega \left(\text{I}-\text{I}_\theta\right)^2dx$.
This work explores \textit{coord-based neural networks} to morph \textit{neural images} using a novel neural \textit{warping} approach.

We assume that a coord-based network is a \textit{sinusoidal} multilayer perceptron (MLP)~\cite{lapedes1987nonlinear,parascandolo2016taming,sitzmann2020implicit} $f_\theta(p):\R^n\to \R^m$ defined as the composition $f_\theta(x)\!=\!W_{d}\circ f_{d-1}\circ \cdots \circ f_{0}(x)+b_{d}$ of $d$ \textit{sinusoidal layers} $f_i(x_i)\!=\!\sin(W_ix_i\!+\!b_i)\!=\!x_{i+1}$, where $W_i\in\R^{n_{i+1}\times {n_i}}$ are the weight matrices, and $b_i\!\in\!\R^{n_{i+1}}$ are the biases. The union of these parameters defines~$\theta$.
The integer $d$ is the \textit{depth} of $f_\theta$ and $n_i$ are the layers \textit{widths}.

The MLP $f_\theta$ is smooth because its layers are composed of smooth maps, and we can compute its derivatives in closed form using automatic differentiation. This property plays an important role in our method since it allows using derivatives for implicit regularization of the warpings and morphings.

\subsection{Neural Morphing}
% \vspace{-0.18cm}
This section introduces the \textit{neural morphing} of two images. It consists of a \textit{neural warping} to align the features of the image and a \textit{ neural blending} of the resulting warped images.

Specifically, let $\text{I}_0, \text{I}_1:\R^2\to \mathcal{C}$ be two neural images, we represent their \textit{neural morphing} using a (time-dependent) neural network
$\mathscr{I}:\R^2\times [0,1]\to \mathcal{C}$  subject to $\mathscr{I}(\cdot, i)\!=\!\text{I}_i(\cdot)$, for $i\!=\!0,1$.
Thus, for each $t$ we have an image $\mathscr{I}(\cdot,t)$, and varying $t$ results in a video interpolating $\text{I}_i$.
To define the morphing~$\mathscr{I}$, we \textbf{disentangle} the spatial deformation (\textit{warping}), used to align the corresponding \textit{features} of $\text{I}_i$ along the time, from the \textit{blending} of the resulting warped~images.

For the warping, we use pairs of \textit{landmarks} {$\{p_j,q_j\}$}, with $j$ being the \textit{landmark index}, sampled from the domains of $\text{I}_0$ and $\text{I}_1$ providing feature correspondences.
Then, we seek a warping $\gt{T}\!\!:\!\R^2\!\!\times\!\! [-1,1]\!\to\! \R^2$ satisfying the \textit{data constraints}:
\begin{itemize}
    \item The curves $\gt{T}(p_j,t)$ and $\gt{T}(q_j,t-1)$, with $t\in[0,1]$, has $p_j$ and $q_j$ as end points;
    \item For each $t\in(0,1)$, we require $\gt{T}(p_j,t)=\gt{T}(q_j,t-1)$.
\end{itemize}
Thus, the values $\text{I}_0(p_j)$ and $\text{I}_1(q_j)$ can be blended along the path $\gt{T}(p_j,t)$.
In points $x\neq p_j$, we employ the well-known \textit{thin-plate} energy to force the transformations to minimize deformation.
The resulting network $\gt{T}$ deforms $\text{I}_i$ along the time resulting in the \textit{warpings} $\mathscr{I}_i\!:\!\R^2\!\times\! [0,1]\!\to\! \mathcal{C}$ defined as:
\vspace{-0.5cm}
\begin{align}\label{e-wapings}
    \mathscr{I}_i(x,t):=\text{I}_i\big(\gt{T}(x,i-t)\big).
\end{align}
\vspace{-0.5cm}

\begin{figure}[hh]
  \centering
  \includegraphics[width=\columnwidth]{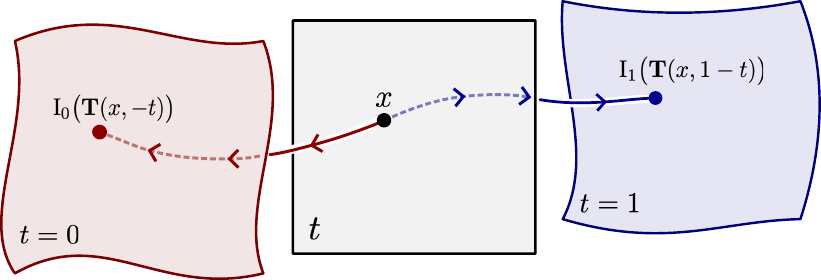}
  % \vspace{-0.2cm}
  \caption{Schematic illustration of the neural warping $\gt{T}$ being used to aligning the initial images $\text{I}_i$}\label{f-warping}
\end{figure}

Fig ~\ref{f-warping} illustrates the warpings~$\mathscr{I}_i$.
Given a point $(x,t)$, to evaluate $x$ in image $\text{I}_i$ we move it to time $t=i$, for $i\!=\!0,1$, which is done by $x_i:=\gt{T}(x, i-t)$.
Note that for $x_0$ and $x_1$, we need the inverse and direct transformations of $\gt{T}$ (in red/blue) since it employs negative and positive time values.

Then we obtain the image values by evaluating~$\text{I}_i(x_i)$.
Moreover, we can move a vector $v_i$ at $x_i$ to $x$, at time $t$, considering the product $v_i\cdot \jac{\gt{T}(x,i-t)}$, where $\text{Jac}$ is the Jacobian. In Section~\ref{s-blending}, we use such property and consider $v_i=\nabla \text{I}_i(x_i)$ to blend the images in the \textit{gradient~domain}.

We blend the resulting aligned warpings $\mathscr{I}_i$ to define the desired morphing $\mathscr{I}:\R^2\times [0,1]\to \mathcal{C}$.
We consider three blending approaches: a simple linear interpolation $\mathscr{I}\!=\!\!(1\!-\!t)\mathscr{I}_0\!+\! t\mathscr{I}_1$, blending in the \textit{gradient domain} using the Poisson equation, and \textit{generative blending} using diffAE.
Section~\ref{s-blending} presents these approaches in detail.

The following steps summarize the procedure of morphing two images $\text{I}_i$:
\begin{itemize}
    \item Extract \textbf{key points} $\{p_j,q_j\}$ in the domains of the face images $\text{I}_0$ and $\text{I}_1$, providing feature correspondence.

    \item Define and train the \textbf{neural warping} $\gt{T}:\R^2\times \R\to \R^2$ to align the key points $\{p_j,q_j\}$ while penalizing distortions using the thin-plate energy. This produces the image warpings $\mathscr{I}_i$ that align the features of $\text{I}_i$ along time;

    \item Blend $\mathscr{I}_i$ to define the \textbf{morphing} $\mathscr{I}\!:\R^2\!\times \R\!\to\!\mathcal{C}$ of~$\text{I}_i$. We~consider two representations for $\mathscr{I}$. First, we use a sinusoidal MLP and exploit its flexibility to train in the \textit{gradient domain}. Second, we embed $\mathscr{I}_i$ in the latent space of diffAE resulting in two curves, then $\mathscr{I}$ is given by interpolating these curves and projecting back to image~space.
\end{itemize}

\subsection{Neural warping}
\vspace{-0.18cm}
This section presents the \textit{neural warping}, a neural network that aligns features of the target images along time.
Precisely, we model it using a sinusoidal MLP $\gt{T}:\R^2\times [-1,1]\to \R^2$, and require the following properties:
\begin{itemize}
    \item $\gt{T}(\cdot, 0)$ is the \textit{identity} (Id);
    \item  For each $t\!\in\![-1,1]$, we have that $\gt{T}_{-t}$ is the \textit{inverse} of~$\textbf{T}_t$.
\end{itemize}
The corresponding deformation of an image $\text{I}:\R^2\to \mathcal{C}$ by $\gt{T}$ is defined using $\mathscr{I}(\cdot, t)=\text{I}\circ \gt{T}(\cdot, -t)$ which uses the inverse $\gt{T}_{-t}$ of $\gt{T}_{t}$. That is one of the reasons we require the inverse property.
In fact, if $\gt{T}$ holds such a property, there is no need to invert the \textit{direct} warp $\gt{T}_{t}$, which is a difficult task in general.
For simplicity, we say that $\mathscr{I}$ is a \textit{warping} of $\text{I}$.
Note that at $t=0$, we have $\mathscr{I}(\cdot, 0)=\text{I}$ because $\gt{T}(\cdot, 0)=\text{Id}$. Thus, $\mathscr{I}$ evolves the initial image $\text{I}$ along time.

We could avoid using the inverse map $\gt{T}_{-t}$ by employing a sampling $\{\text{I}_{ij}\}$ of $\text{I}$ on a regular grid $\{x_{ij}\}$ of the image support. Then, $\{\text{I}_{ij}\}$ are samples of the warped image $\text{I}\circ \gt{T}_{-t}$ at points $\{\gt{T}_t(p_{ij})\}$. However, this approach has the drawbacks of resampling $\text{I}\circ \gt{T}_{-t}$ in a new regular grid which can result in \textit{holes} and relies on interpolation techniques.
Our method avoids such problems since it will be trained to fit the property $\gt{T}_t\circ \gt{T}_{-t}=\text{Id}$ for $t\in[-1,1]$.

Observe that, for each $t$, the map $\gt{T}_{t}$ approximates a \textit{diffeomorphism} since it is a smooth sinusoidal MLP with an inverse also given by a sinusoidal MLP $\gt{T}_{-t}$ since $\textbf{T}_{t}\circ \textbf{T}_{-t}\!\!=\!\text{Id}$.

\vspace{-0.35cm}
\subsubsection{Loss function}
\vspace{-0.18cm}
Let $\text{I}_0, \text{I}_1\!\!:\!\!\R^2\!\!\to\! \mathcal{C}$ be neural images and $\{p_j,q_j\}$ be the \textit{source} and \textit{target} points sampled from the supports of $\text{I}_0$~and~$\text{I}_1$ that provide feature correspondences.
Let $\gt{T}\!:\!\R^2\!\times\! [-1,1]\!\to\! \R^2$ be a sinusoidal MLP, we train its parameters $\theta$ so that $\gt{T}$ approximates a warping aligning the key points $p_j$ and $q_j$ along time. For this, we use the following loss functional:

\vspace{-0.45cm}
\begin{align}
    \mathscr{L}(\theta) = \mathscr{W}(\theta)+\mathscr{D}(\theta)+\mathscr{T}(\theta).
\end{align}
Where $\mathscr{W}(\theta)$, $\mathscr{D}(\theta)$, $\mathscr{T}(\theta)$ are the \textit{warping}, \textit{data}, and \textit{thin-plate} constraints.
%WARPING CONSTRAINT
$\mathscr{W}(\theta)$ requires the network $\gt{T}$ to satisfy the identity and inverse properties of the warping definition,
\vspace{-0.15cm}\small{
\begin{align}
    \mathscr{W}\!(\theta)\!\!=\!\!\!\underbrace{\int\limits_{\R^2}\!\!\norm{\gt{T}(x,0)\!-\!x}^2\!\!dx}_{\text{Identity constraint}} \!+\!\!\!\! \underbrace{\int\limits_{\R^2\!\times\! [-1,1]}\!\!\!\!\!\!\!\!\!\norm{\gt{T}\big(\gt{T}(x,t),-t\big)\!\!-\!x}^2\!\!\!dxdt}_{\text{Inverse constraint}}.
\end{align}}
\normalsize
The \textit{identity} constraint forces $\gt{T}_0=\text{Id}$ and, the \textit{inverse} constraint asks for $\gt{T}_{-t}$ to be the inverse of $\gt{T}_t$ for all $t\in \R$.

%DATA CONSTRAINT
The \textit{data constraint} $\mathscr{D}(\theta)$ is responsible for forcing $\gt{T}$ to move the source points $p_j$ to the target points $q_j$ such that their paths match along time.
For this, we simply consider:
\vspace{-0.15cm}
\begin{align}\label{e-data-constraint}
    \mathscr{D}(\theta)=\sum\limits_{j}\int_{[0,1]}\norm{\gt{T}(p_j,t)-\gt{T}(q_j,t-1)}^2dt.
\end{align}
Note that $\mathscr{D}$ is asking for $\gt{T}(p_j,1)=q_j$ and $\gt{T}(q_j,-1)=p_j$ because at the same time $\mathscr{W}$ is forcing the identity property. Moreover, it forces $\gt{T}(p_j,t)=\gt{T}(q_j,t-1)$ along time, thus, as observed at the beginning of this section, this is the required property for the key points $\{p_j,q_j\}$ be aligned along time.
Since we assume $\gt{T}$ to be a sinusoidal MLP, the resulting warping provides a smooth deformation that moves the source points to the target points.

\pagebreak

%THIN-PLATE CONSTRAINT
However, $\mathscr{D}$ does not add restrictions on points other than the source and target points.
Even assuming $\gt{T}$ to be smooth the resulting warping would need some regularization, such as minimizing distortions.
For this, we propose a \textit{regularization} which penalizes distortions of the transformations $\gt{T}_t$ using the well-known \textit{thin-plate} energy~\cite{bookstein1989principal,glasbey1998review}:
\vspace{-0.15cm}
\begin{align}\label{e-TPS_regularization}
    \mathscr{T}(\theta) = \int_{\R^2\times [-1,1]} \norm{\hess{\gt{T}}(x,t)}_F^2 dxdt.
\end{align}
$\mathscr{T}$ regularizes $\gt{T}$ and works as a bending energy term penalizing deformation, at each point $(x,t)$, based on the derivatives of $\gt{T}$. This helps eliminate global effects that may arise from considering only data and warping constraints.
It is important to note that we have incorporated the time variable into the thin-plate energy $\mathscr{T}$.

\begin{figure}[hh]
  \centering
  \includegraphics[width=0.19\columnwidth]{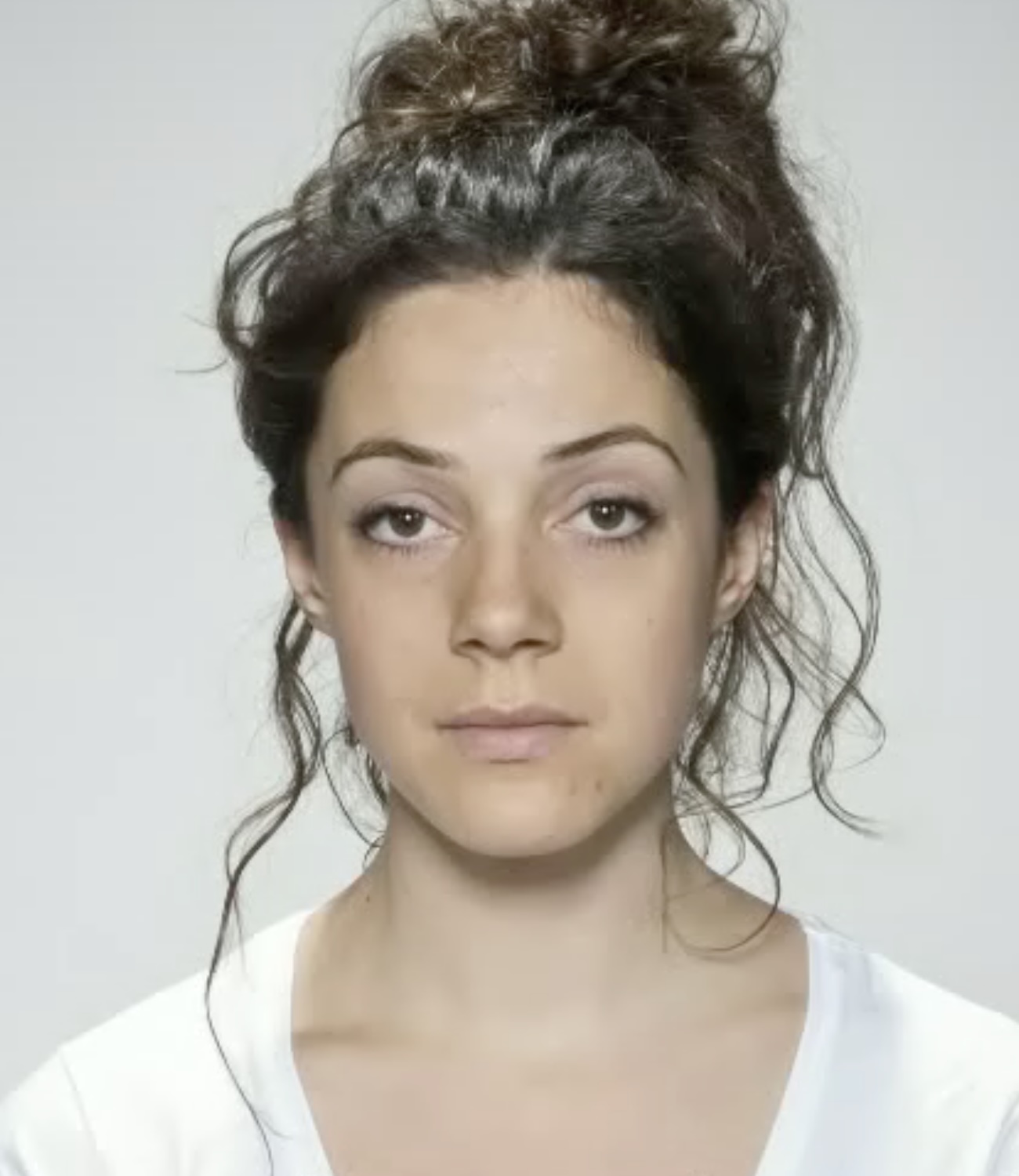}
  \includegraphics[width=0.19\columnwidth]{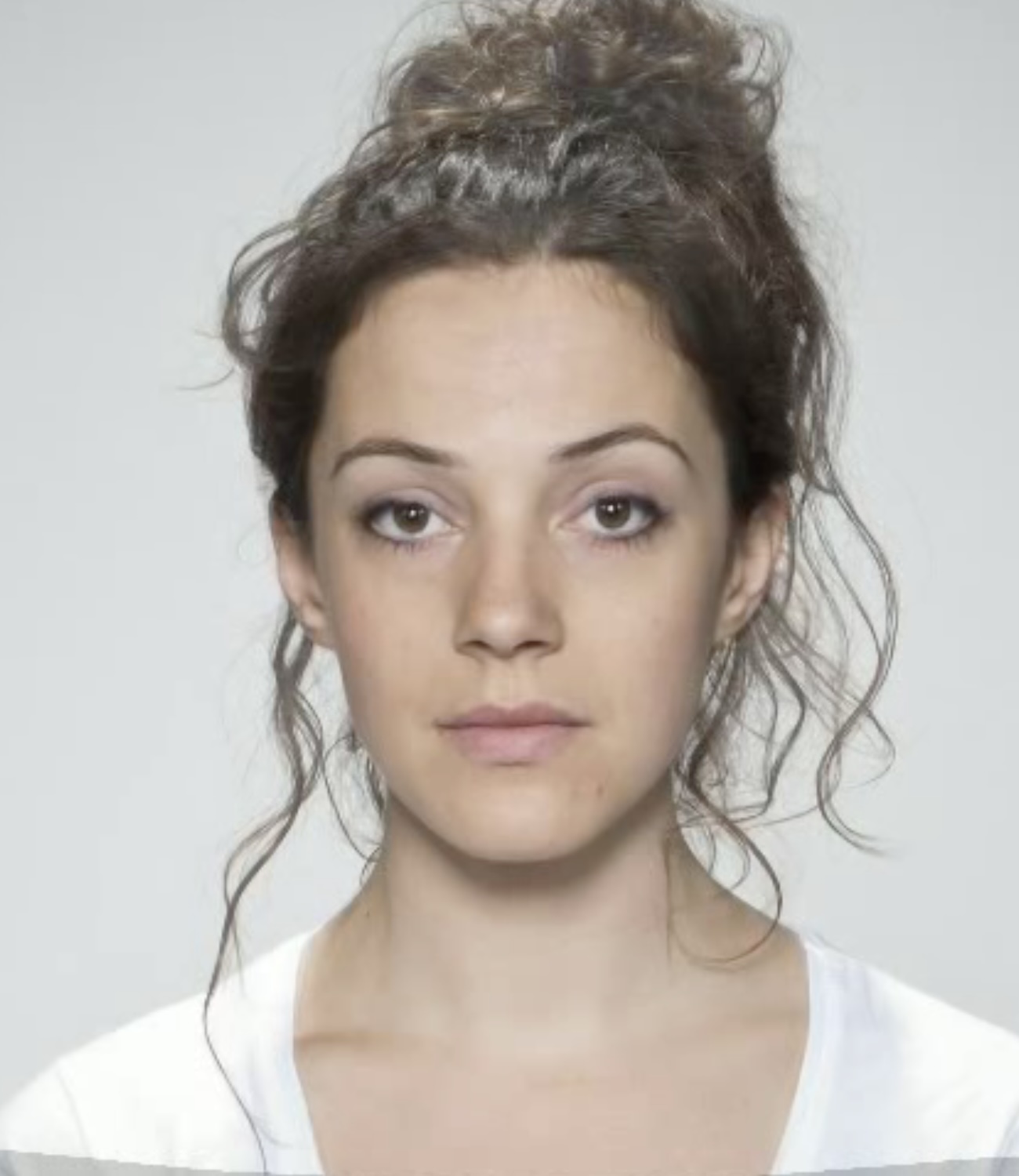}
  \includegraphics[width=0.19\columnwidth]{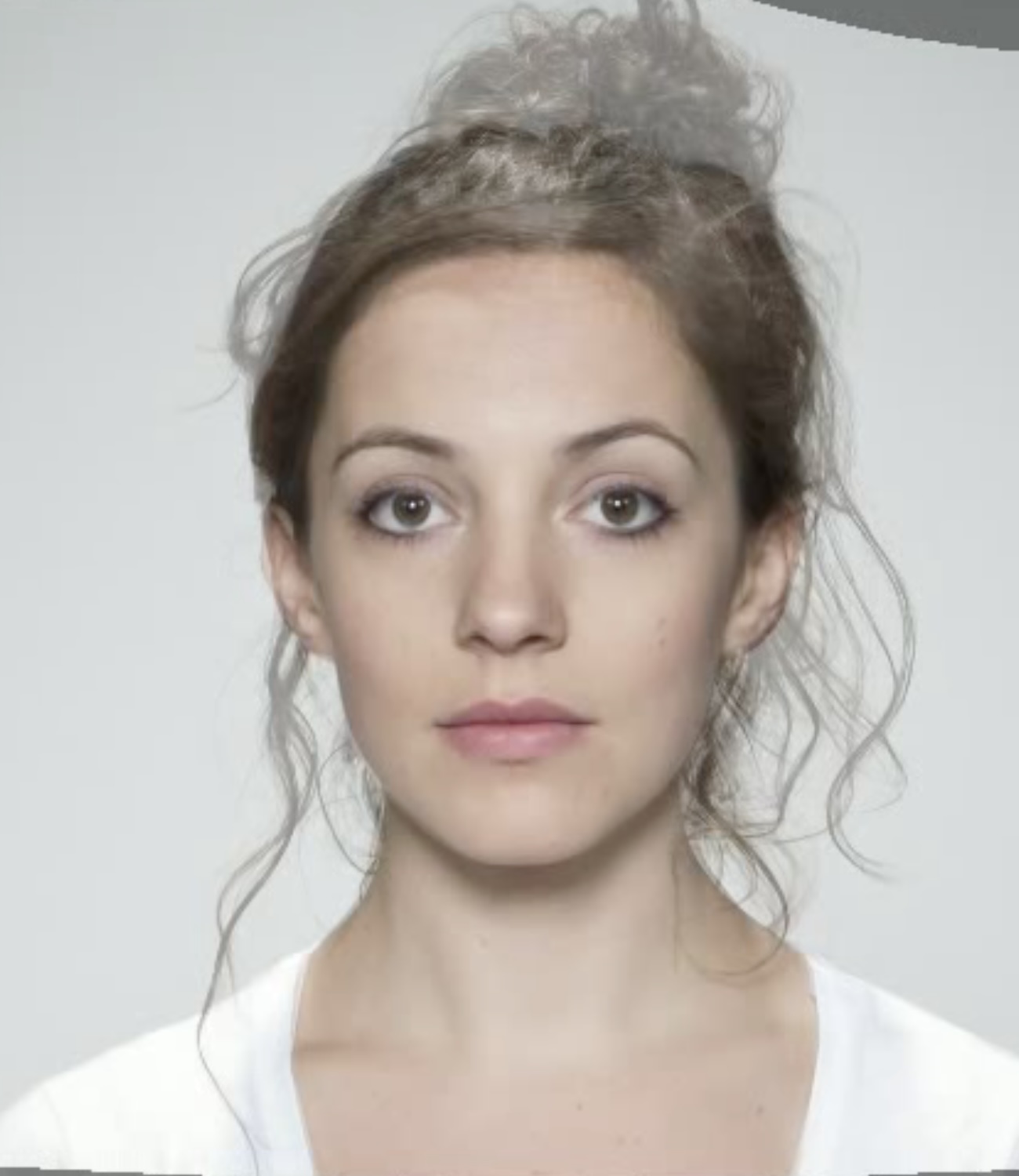}
  \includegraphics[width=0.19\columnwidth]{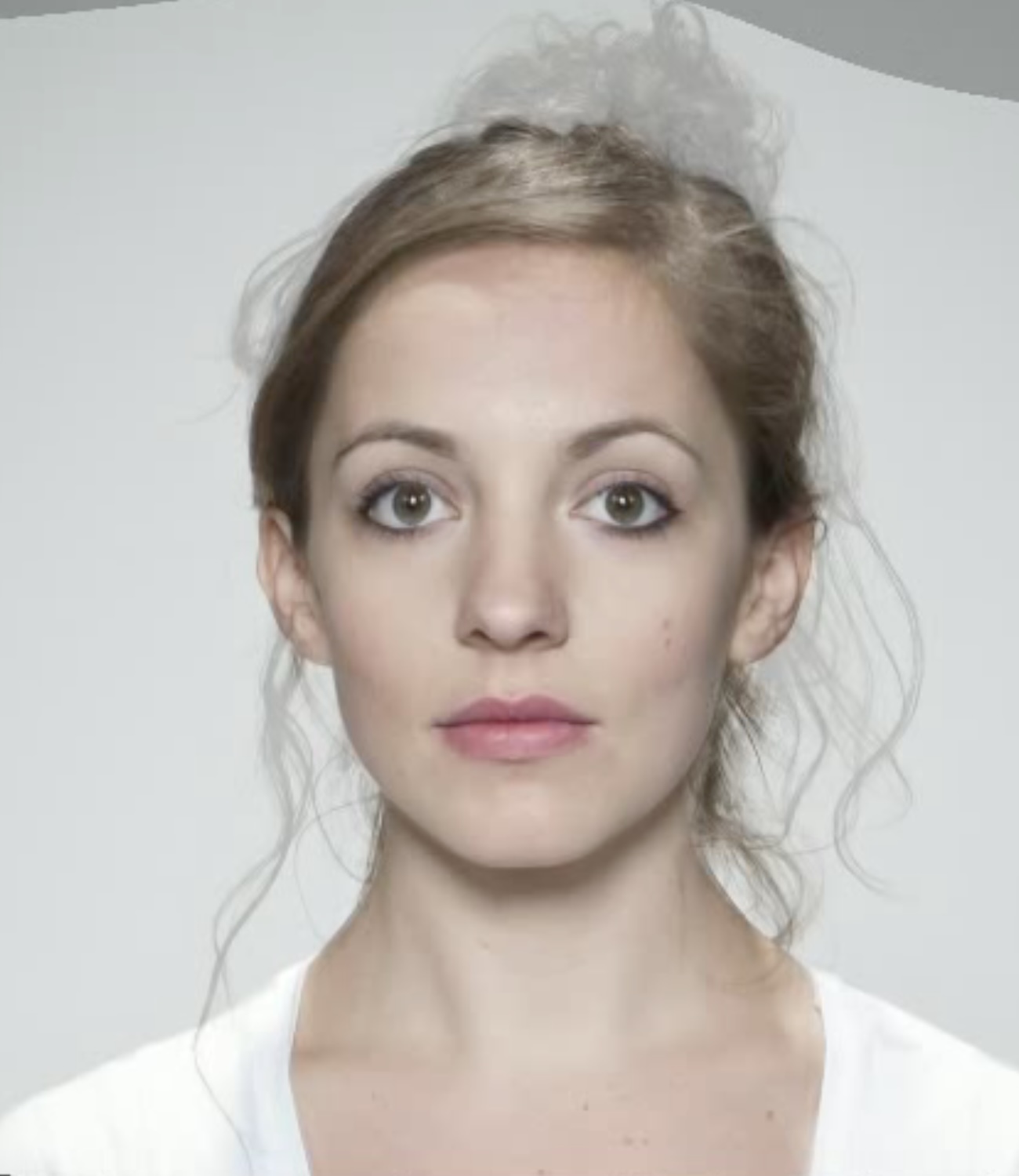}
  \includegraphics[width=0.19\columnwidth]{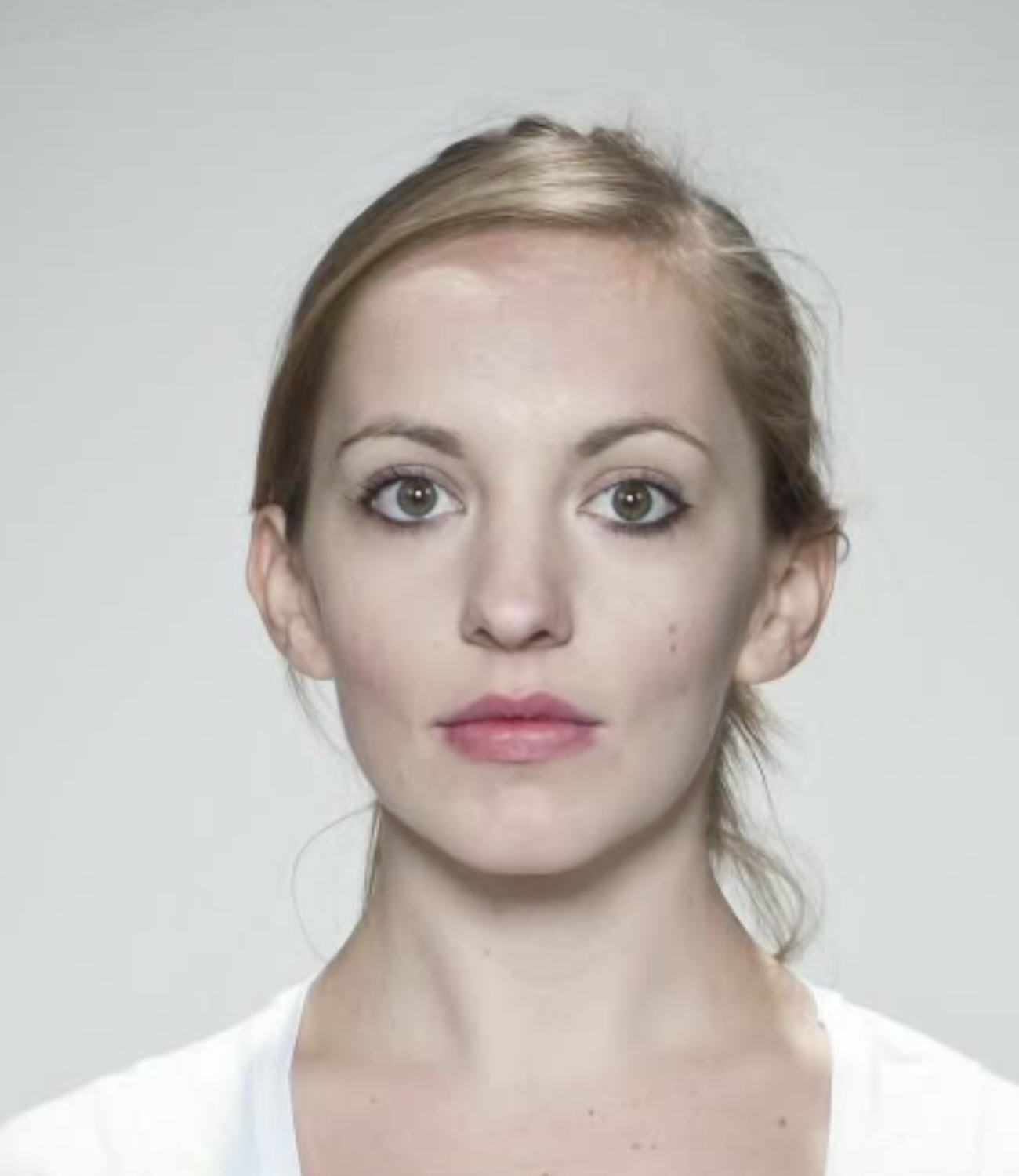}
  \caption{A neural warping $\gt{T}$ continuously aligning two face images along time. We use $\gt{T}$ to create their aligned warpings $\mathscr{I}_i$. The morphing $(1-t)\mathscr{I}_0+t\mathscr{I}_1$ was sampled at $t=0,0.25, 0.5, 0.75, 1$.}
  \label{f-warping_two_imgs}
\end{figure}

By using a sinusoidal MLP to model $\gt{T}$ and training it with $\mathscr{W}$ while regularizing with the thin-plate energy, we achieve robust warpings, see Fig~\ref{f-warping_two_imgs} for an alignment between two images, for more detail see the experiments in Sec~\ref{s-experiments}.

\begin{figure}[h!]
  \centering
  \includegraphics[width=0.21\columnwidth]{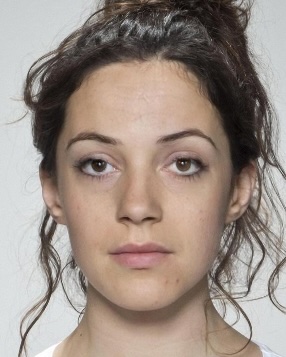}
  \includegraphics[width=0.21\columnwidth]{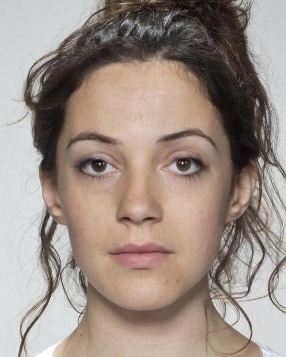}
  \includegraphics[width=0.21\columnwidth]{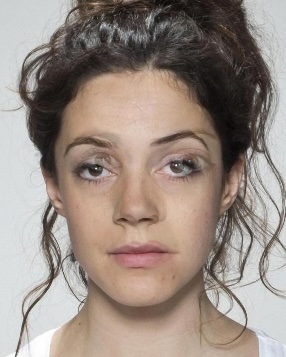}
  \includegraphics[width=0.21\columnwidth]{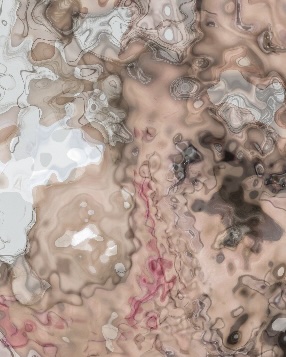}
  \caption{Loss term impact experiment. From the left: results without the inverse, identity, data, and thin-plate constraints.}
  \label{fig:loss-ablation}
\end{figure}

Additionally, we perform experiments to assess the impact of each term  $\mathscr{W},\,\mathscr{D},\,\mathscr{T}$ to understand their importance during the training of $\gt{T}$.
We found out that the thin-plate constraint $\mathscr{T}$ is crucial. Also, as expected without the data constraint $\mathscr{D}$ we can not align the image features. The warping constraint has less influence, acting mostly on finer details. That was an interesting finding implying that the warping properties are being enforced by $\mathscr{T}$. This is probably due to the fact that $\mathscr{D}$ forces such property along the feature paths and $\mathscr{T}$ asks for the deformation to be minimized in $\R^2\times[-1,1]$.
Fig~\ref{fig:loss-ablation} illustrates the experiment.

\vspace{-0.4cm}
\subsection{Neural Blending}
\label{s-blending}
\vspace{-0.18cm}
Let $\text{I}_i\!:\!\R^2\!\to\! \mathcal{C}$ be two neural images and $\gt{T}:\R^2\times \R\!\to\! \R^2$ be a neural warping aligning their features.
Specifically, the images $\text{I}_i$ are deformed by $\gt{T}$ along time and Eq.~\ref{e-wapings} gives the corresponding warpings $\mathscr{I}_i(x,t)=\text{I}_i\big(\gt{T}(x,i-t)\big)$.
Then, we blend $\mathscr{I}_i$ or their derivatives to construct a morphing $\mathscr{I}:\R^2\times\R\to \mathcal{C}$ of the initial images $\text{I}_i$.
A naive blending approach could be defined directly from $\mathscr{I}_i$ by interpolating using $\mathscr{I}(x,t)=(1-t)\mathscr{I}_0(x, t)+t\mathscr{I}_1(x, t).$
Thus, at $t=0$ and $t=1$, we obtain $\mathscr{I}_0$ and $\mathscr{I}_1$, respectively (See Fig~\ref{f-warping_two_imgs}).
Note that $\mathscr{I}$ is a smooth function both in time and space.

\vspace{-0.4cm}
\subsubsection{Blending in the gradient domain}
\label{s-blending-grad-domain}
\vspace{-0.18cm}
Interpolating $\text{I}_i$ does not allow us to keep parts of one of the images unchanged during the morphing, e.g. the complement region of the face.
To address these issues, inspired by the \textit{Poisson image editing} technique~\cite{perez2003poisson}, we propose to blend $\text{I}_i$ by solving a \textit{boundary value problem} in $\R^2\times \R$ to handle smooth animations and model $\mathscr{I}$ by a neural network.

We use the Jacobians $\jac{\mathscr{I}_i}$ of the warpings $\mathscr{I}_i$ to train $\mathscr{I}$.
We restrict the morphing support to $S\!\!=\!\![-1,1]^2\!\!\times\![0,1]$, with $[-1,1]^2$ representing the image domain and $[0,1]$ is the time interval.
Let $\Omega\subset S$ be an open set used for blending~$\mathscr{I}_i$, such as the interior of the face path, and let $\mathscr{I}^*\!\!:\!S\!\to\! \R$ be a known function on $S-\Omega$ (it could be either $\mathscr{I}_0$ or $\mathscr{I}_1$).
Finally, let $U$ be a matrix field obtained by blending $\jac{ \mathscr{I}_i}$, for example, $U=(1-t)\jac{ \mathscr{I}_0}+t \jac{ \mathscr{I}_1}$.
A common way to extend $\mathscr{I}^*$ to $\Omega$ is by solving:
\vspace{-0.2cm}
\begin{align}\label{e-gradient-interpolant}\small
\!\min \!\!\int_{\Omega} \!\!\!\norm{{\jac{\mathscr{I}}\!-\!U}}^2\!dxdt \text{ subject to } \mathscr{I}|_{S-\Omega}\!=\!\mathscr{I}^*|_{S-\Omega}.
\end{align}
\vspace{-0.4cm}

\noindent We propose to use this variational problem to define the following loss function to train the parameters $\theta$ of $\mathscr{I}$.

\vspace{-0.4cm}
\begin{align}\label{e-blending-no-grad}\small
\mathscr{M}(\theta)\!\!=\!\!\underbrace{\int_{\Omega} \norm{{\jac{\mathscr{I}}-U}}^2dxdt}_{\mathscr{C}(\theta)} + \underbrace{\int_{S-\Omega} \!\!\!\!\!(\mathscr{I}-\mathscr{I}^*)^2dxdt}_{\mathscr{B}(\theta)}.
\end{align}

\vspace{-0.2cm}

\noindent The \textit{cloning term} ${\mathscr{C}(\theta)}$ fits $\mathscr{I}$ to the primitive of $U$ in $\Omega$, and the \textit{boundary constraint} $\mathscr{B}(\theta)$ forces $\mathscr{I}\!=\!\mathscr{I}^*$ in $S\!-\!\Omega$.
Thus, $\mathscr{M}$ trains $\mathscr{I}$ to \textit{seamless clone} the primitive of $U$ to $\mathscr{I}^*$ in $\Omega$.
Unlike classical approaches that rely on pixel manipulation, seamless cloning operates on the image gradients.

Since the images $\text{I}_i$ contain faces and $\gt{T}$ aligns their features, we define $\Omega$ as the path of the facial region over time.
Specifically, let $\Omega_0$ be the region containing the face in $\text{I}_0$, define $\Omega$ by warping $\Omega_0$ along time using $\gt{T}$, i.e., $\Omega=\cup_{t\in[0,1]}\gt{T}_t(\Omega_0)$. Note that the deformation of $\Omega_0$ uses the direct deformation $\gt{T}_t$ while the warped image $\mathscr{I}_0$ uses the inverse~$\gt{T}_{-t}$. The use of both inverse/direct deformations encoded in our neural warping avoids the need to compute inverses at inference time.
Finally, for each~$t$, $\gt{T}$ aligns the faces $\text{I}_i$ in the region $\gt{T}_t(\Omega_0)$. Thus, $\mathscr{M}$ trains $\mathscr{I}$ to morph the face in $\text{I}_0$ into the face in $\text{I}_1$ while cloning the result to $\mathscr{I}_0$ on $S-\Omega$.

Besides choosing $U$ as a linear interpolation of $\jac{ \mathscr{I}_i}$, which we call the \textit{averaged seamless cloning} case, we could choose
$U\!~\!\!=\!\!~\!\jac{ \mathscr{I}_1} \text{ and } \mathscr{I}^*\!\!=\!\mathscr{I}_0$.
So, the resulting loss function $\mathscr{M}$ forces $\mathscr{I}$
to \textit{seamless clone} the face $\mathscr{I}_1$ to the corresponding region of $\mathscr{I}_0$.

\begin{figure}[hh]
  \centering
  \includegraphics[width=0.225\columnwidth]{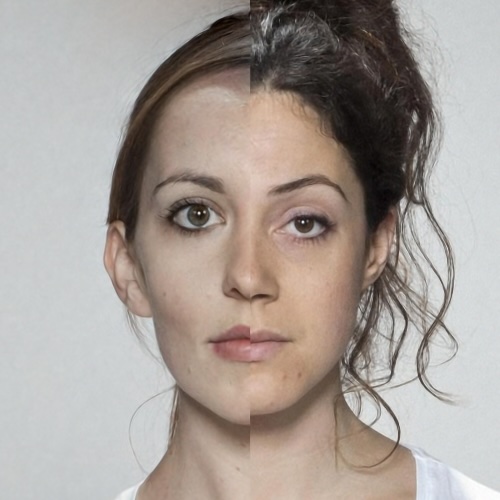}
  \includegraphics[width=0.225\columnwidth]{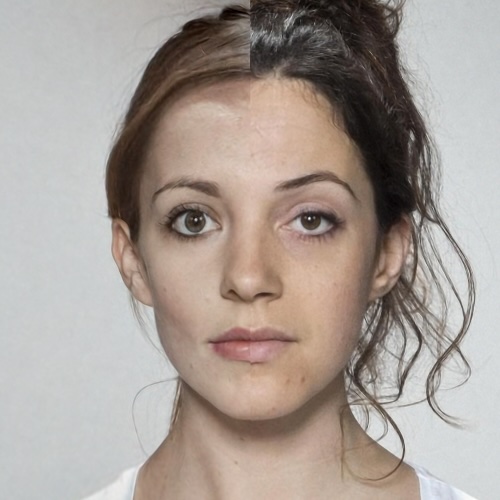}
  \includegraphics[width=0.225\columnwidth]{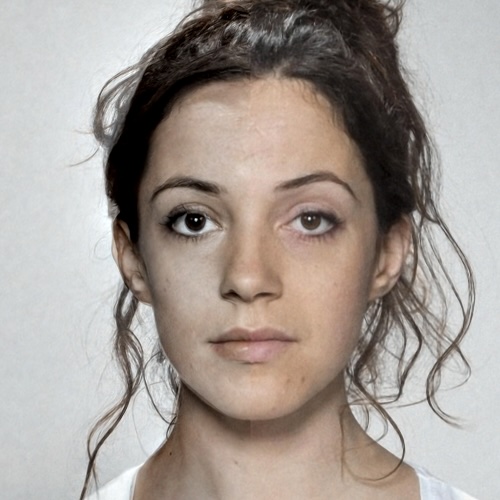}
  \includegraphics[width=0.225\columnwidth]{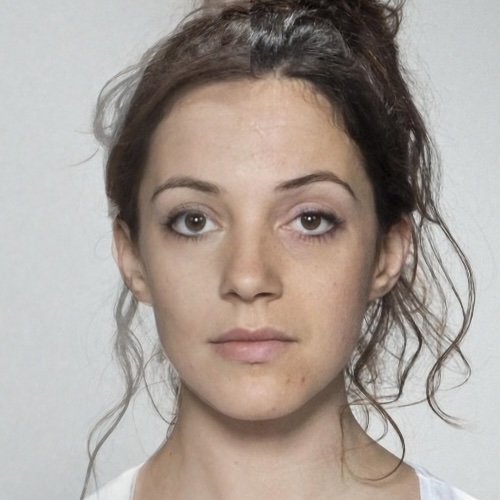}
  \\
  \vspace{0.05cm}
  \includegraphics[width=0.225\columnwidth]{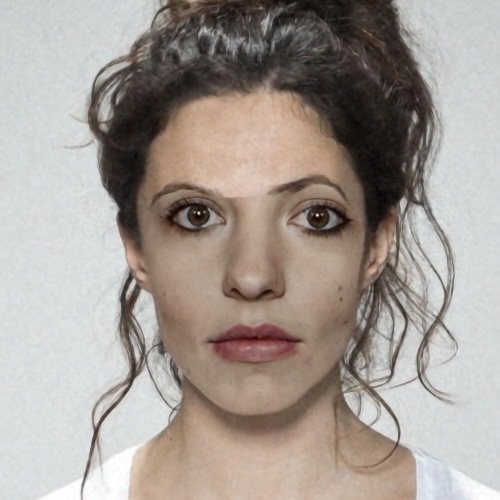}
  \includegraphics[width=0.225\columnwidth]{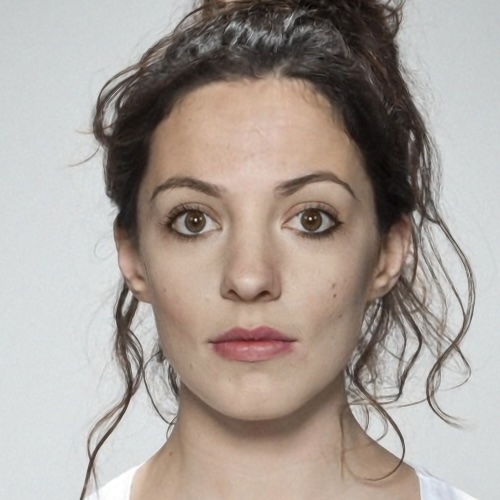}
  \includegraphics[width=0.225\columnwidth]{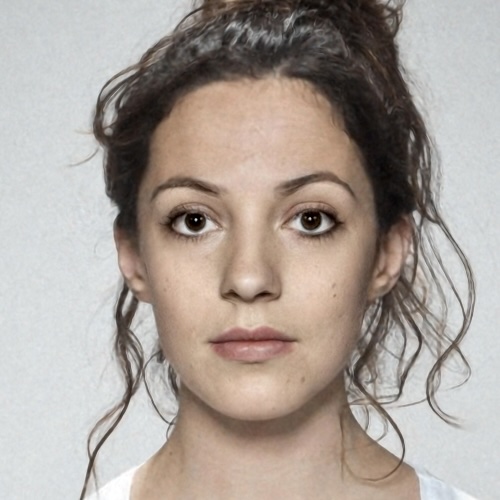}
  \includegraphics[width=0.225\columnwidth]{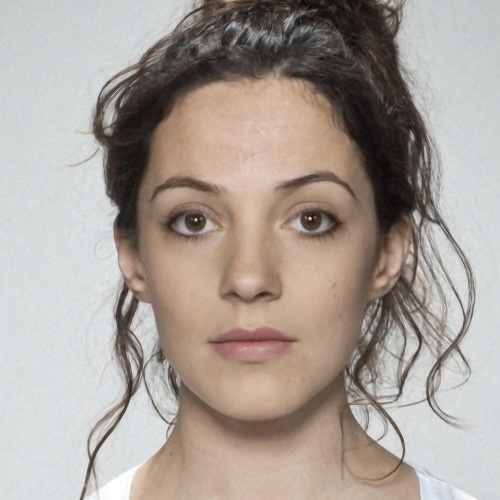}\\
  \footnotesize{\text{\,\,\,\,No warping} \quad \text{seamless cloning} \quad \text{average cloning} \quad \text{mixed cloning}}
  \vspace{-0.2cm}
  \caption{Comparing different neural blendings of two faces $\text{I}_i$.
    Line 1/2 shows examples of cloning the half-space region of $\text{I}_1$ into $\text{I}_0$. In Column 1 we do not align the image landmarks, the remaining columns use our neural warping for the alignment. Column~2 uses $U \!\!=\! \!\jac{\mathscr{I}_1}$ and $\mathscr{I}^*\!\!=\!\!\mathscr{I}_0$ in the neural blending. Columns~3 and 4 applies the mixed and normal seamless clone respectively.
    }
  \label{f-neural_blending_two_imgs}
\end{figure}

It may be desirable to combine features of~$\mathscr{I}_i$, however an interpolation of $\jac{\mathscr{I}_i}$ can lead to loss of details. To avoid it, we extend the approach in \cite{perez2003poisson}, which allows mixing the features of both images. At each $(x,t)$, we retain the stronger of the variations in the warpings by choosing $U\!\!=\!\! \jac{\mathscr{I}_0}$ if $\norm{\jac{\mathscr{I}_0}}\!>\!\norm{\jac{\mathscr{I}_1}}$, and $U\!\! =\!\! \jac{\mathscr{I}_1}$, otherwise. The resulting loss function $\mathscr{M}$ forces $\mathscr{I}$ to learn a \textit{mixed seamless clone} of $\mathscr{I}_i$.
Fig~\ref{f-neural_blending_two_imgs} shows examples of neural blending.

\vspace{-0.8cm}
\subsubsection{Blending using generative models}
\vspace{-0.18cm}
Generative models may be used to interpolate faces. However, they do not ensure feature alignment, only provide a blending of the images.
To overcome this issue, we use our neural warping to align the face features and a generative blending to combine the resulting warped images over time. Sec.~\ref{sec:quantitative-results} presents experiments with this approach.

Specifically, let $\text{I}_i$ be neural images representing two faces and $\gt{T}$ be a neural warping aligning their features.
Again, the images $\text{I}_i$ are deformed by $\gt{T}$ along time resulting in the image warpings $\mathscr{I}_i$. Recall that, for each $t\in[0,1]$, we have that the faces in $\mathscr{I}_0(t)$ and $\mathscr{I}_1(t)$ have their features aligned.
Let $\mathscr{E}$ and $\mathscr{D}$ be the \textit{encoder} and \textit{decoder} of a generative model.
We embed $\mathscr{I}_i$ in the latent space which results in the \textit{code curves} $\mathscr{c}_i(t)=\mathscr{E}\big(\mathscr{I}_i(\cdot, t)\big)$.
Then, we interpolate the curves directly in the latent space and the desired generative morphing is given by projecting the resulting curve to the image space using the decoder $\mathscr{D}$:
\vspace{-0.3cm}
\begin{align}
    \mathscr{I}(\cdot, t):=\mathscr{D}\Big((1-t)\mathscr{c}_0(t) + t\mathscr{c}_1(t)\Big).
\end{align}
\vspace{-0.5cm}

With the \textit{generative morphing} $\mathscr{I}$ we have the feature correspondence along time and their path explicitly. We use it to improve the temporal coherence in generative approaches.

In practice, we employ diffAE~\cite{preechakul:2022} since, unlike GANs that depend on error-prone inversion, it encodes the input and produces high-quality output without an optimization step. Moreover, the output of the target images is close to the originals, i.e. $\mathscr{I}(i)\approx\text{I}_i$, which is desirable for the morphing task.
Also, note that to blend images using diffAE we have to interpolate between two-part codes with a semantic and a stochastic part.

\begin{figure}[h!]
  \centering
  \includegraphics[width=\columnwidth]{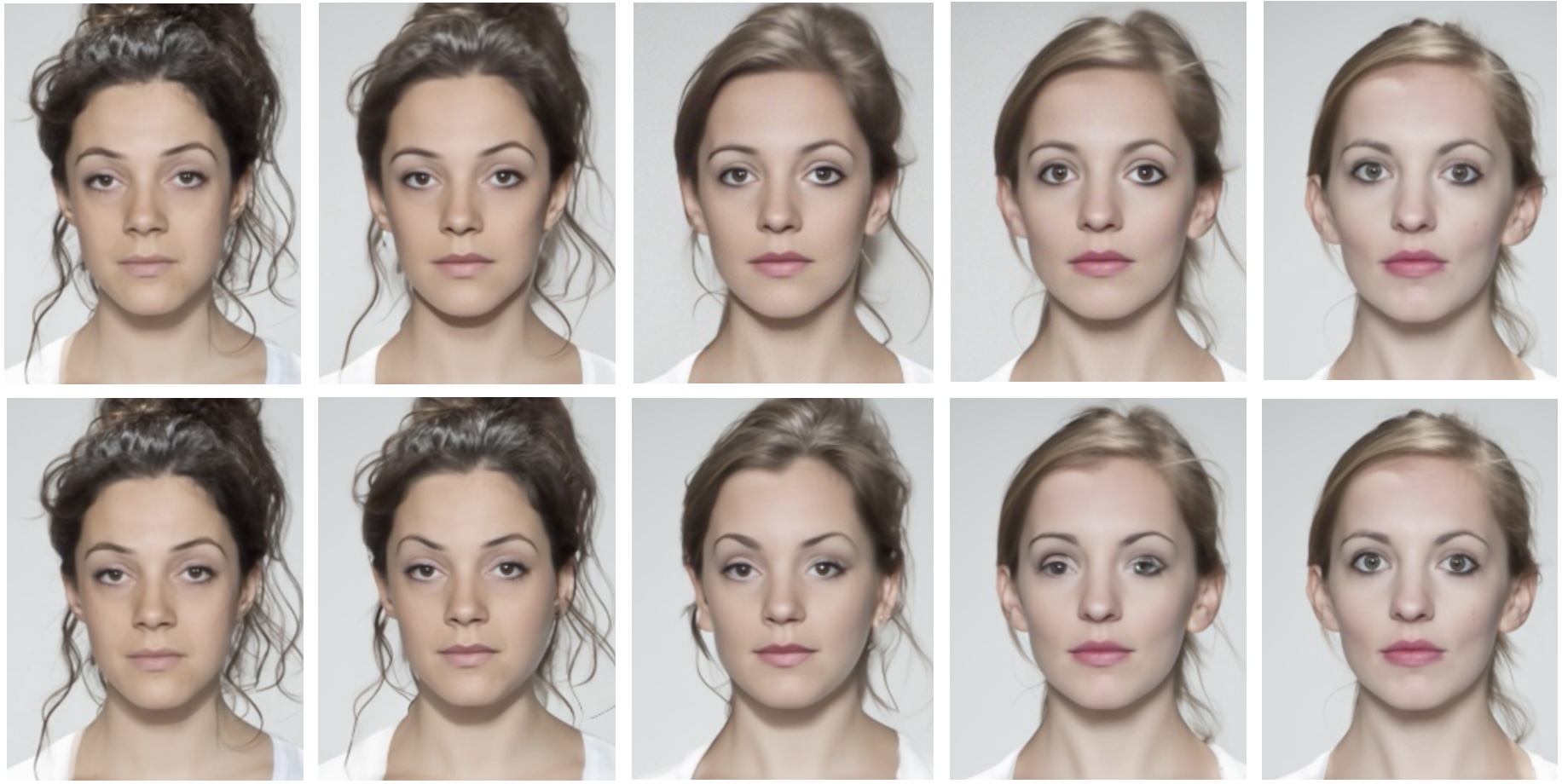}
  \vspace{-0.6cm}
  \caption{Generative morphing. Line 1 presents a morphing between two faces using the generative morphing (neural warping + diffAE). Line 2 shows the results of diffAE using no alignment.}
  \label{fig:diffae_pure_warp}
\end{figure}

Fig~\ref{fig:diffae_pure_warp} shows a comparison between the generative morphing and a pure diffAE applied to $\text{I}_i$.
Line 1 presents samples of the generative morphing $\mathscr{I}(\cdot, t)$.
In Line 2, we simply interpolate between the codes of $\text{I}_i$.
Note that the generative morphing offers smoother transitions between corresponding features; see the video in the supplementary material.

\vspace{-0.3cm}

This experiment does not employ the pre-processing step of fixing the eyes and mouth in the image support. This step is common in generative approaches and relies on DLib~\cite{kazemi2014cvpr,sagonas2016} to detect facial features. For the experiment using this alignment, refer to Fig~\ref{fig:morphing_comparisons}. However, such dependence on generative models forces the eyes and mouths to remain fixed in the image support over time. Hence, we cannot morph between roto-translated images.

\vspace{-0.2cm}
\section{Experiments and Discussions}
\label{s-experiments}
\vspace{-0.2cm}
In the experiments, we used small sinusoidal MLPs consisting of a single hidden layer with $128$ neurons to parametrize the neural warpings.
However, our ablation study indicated that smaller networks also works, see the supp. material.
This shows that our representation is compact and robust for time-dependent warpings. The network initialization follows the definitions in \cite{sitzmann2020implicit}. Additionally we use DLib~\cite{kazemi2014cvpr,sagonas2016} for landmark detection. For the experiments, StyleGAN3 was fine-tuned with images from the FRLL dataset for $312$ epochs, while diffAE was used directly from the authors' repository (model FFHQ256, autoencoding only).

\vspace{-0.1cm}
\subsection{Qualitative comparisons}
\label{sec:qualitative-results}
\vspace{-0.18cm}

\begin{figure}[hh]
  \text{\footnotesize Neural warping + linear blending [Ours]}\\
  \includegraphics[width=0.19\columnwidth]{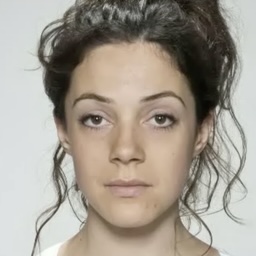}
  \includegraphics[width=0.19\columnwidth]{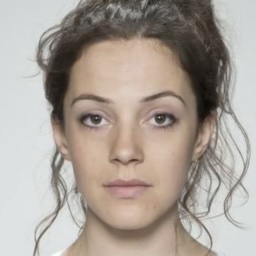}
  \includegraphics[width=0.19\columnwidth]{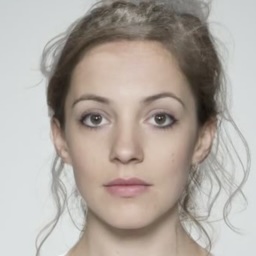}
  \includegraphics[width=0.19\columnwidth]{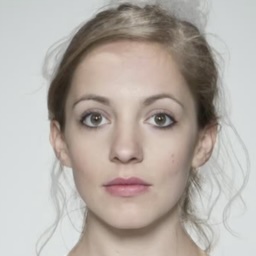}
  \includegraphics[width=0.19\columnwidth]{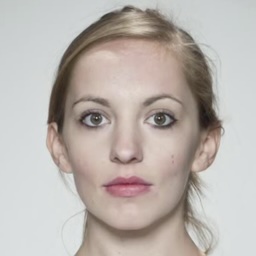}
  \text{\footnotesize FFHQ alignment + diffAE}\\
  \includegraphics[width=0.19\columnwidth]{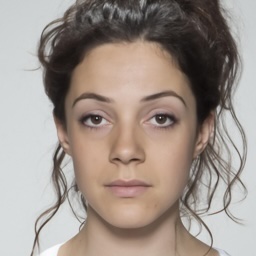}
  \includegraphics[width=0.19\columnwidth]{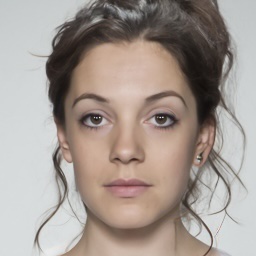}
  \includegraphics[width=0.19\columnwidth]{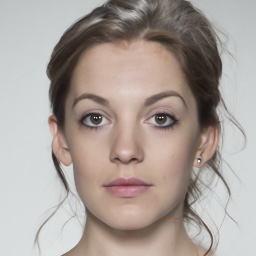}
  \includegraphics[width=0.19\columnwidth]{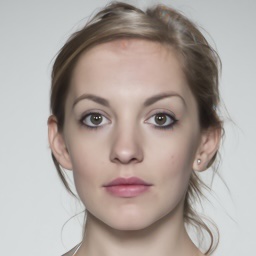}
  \includegraphics[width=0.19\columnwidth]{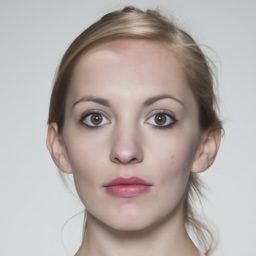}
  \text{\footnotesize Neural warping + diffAE (generative morphing) [Ours]}\\
  \includegraphics[width=0.19\columnwidth]{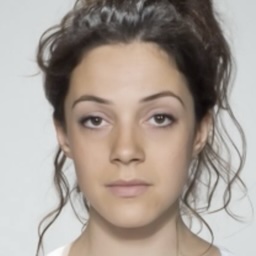}
  \includegraphics[width=0.19\columnwidth]{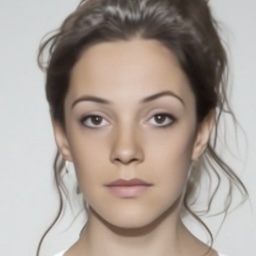}
  \includegraphics[width=0.19\columnwidth]{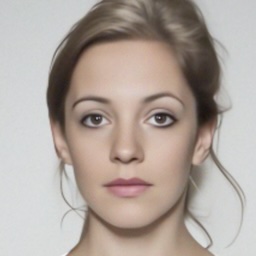}
  \includegraphics[width=0.19\columnwidth]{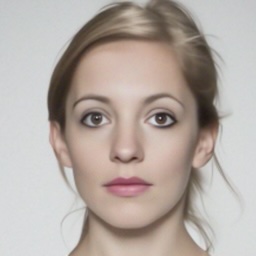}
  \includegraphics[width=0.19\columnwidth]{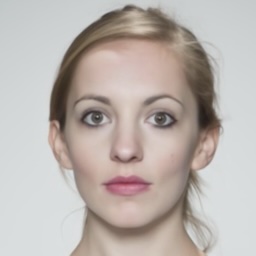}
  \text{\footnotesize FFHQ alignment + StyleGAN3}\\
  \includegraphics[width=0.19\columnwidth]{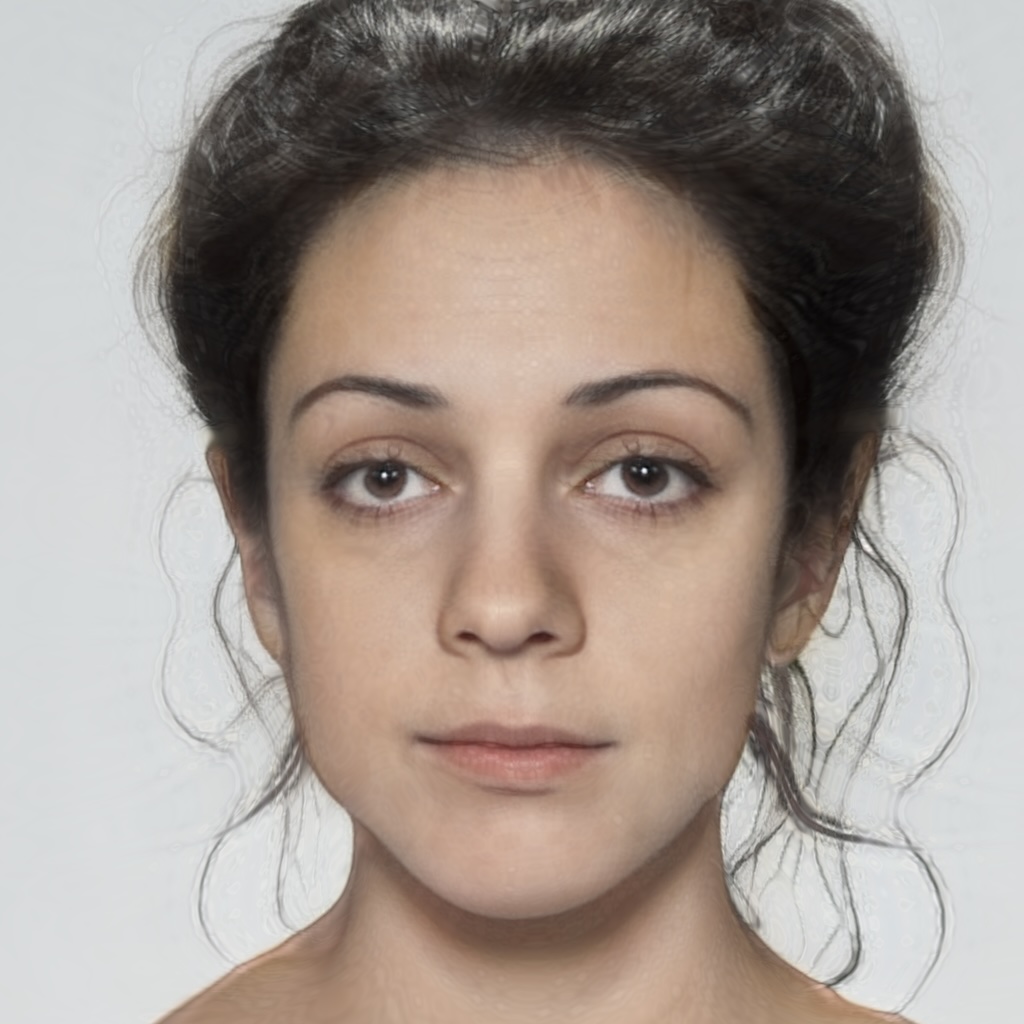}
  \includegraphics[width=0.19\columnwidth]{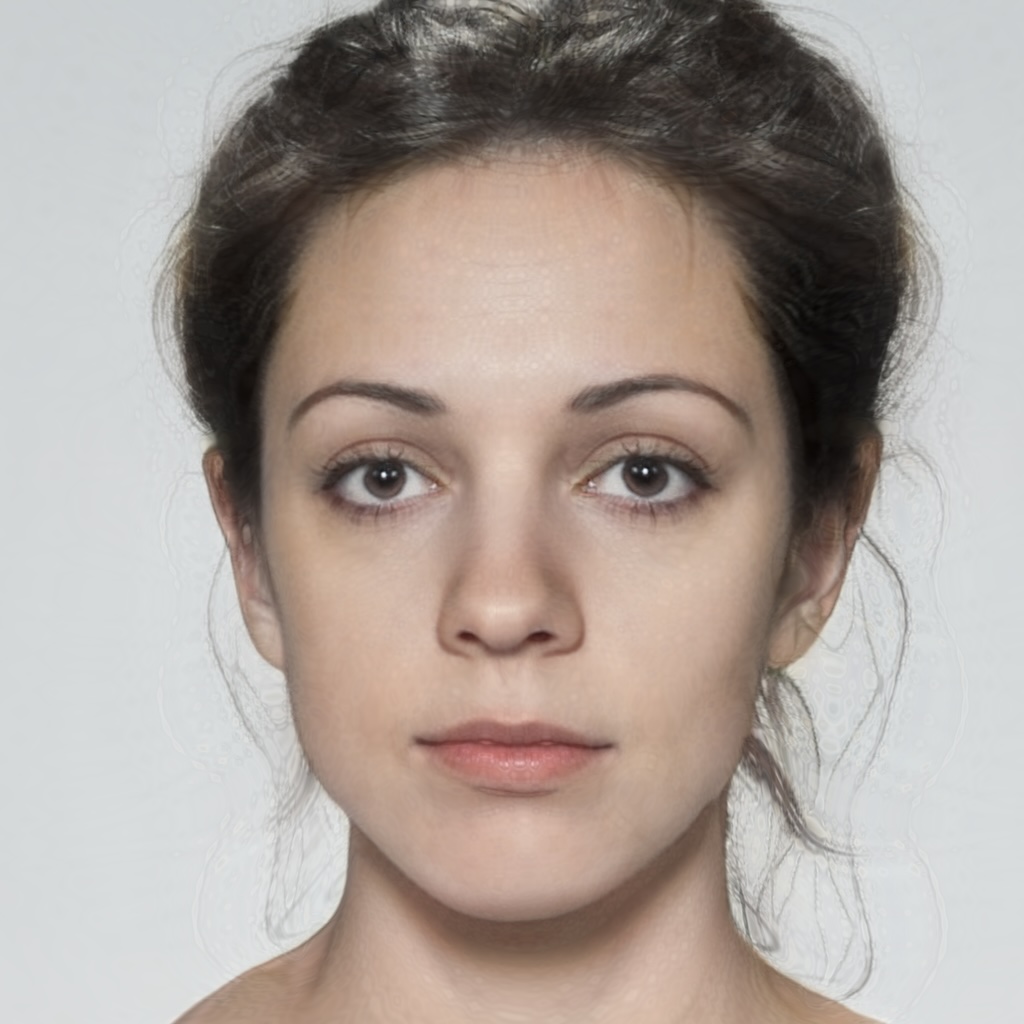}
  \includegraphics[width=0.19\columnwidth]{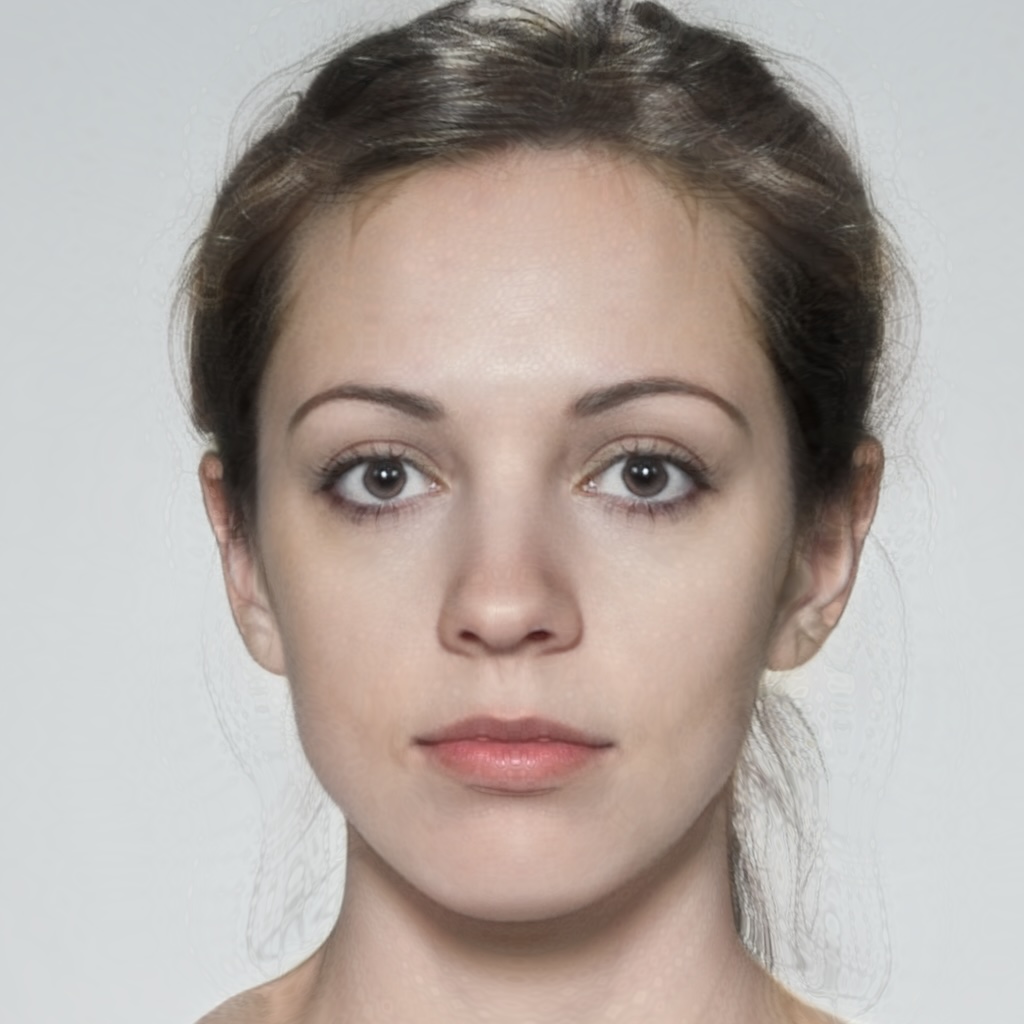}
  \includegraphics[width=0.19\columnwidth]{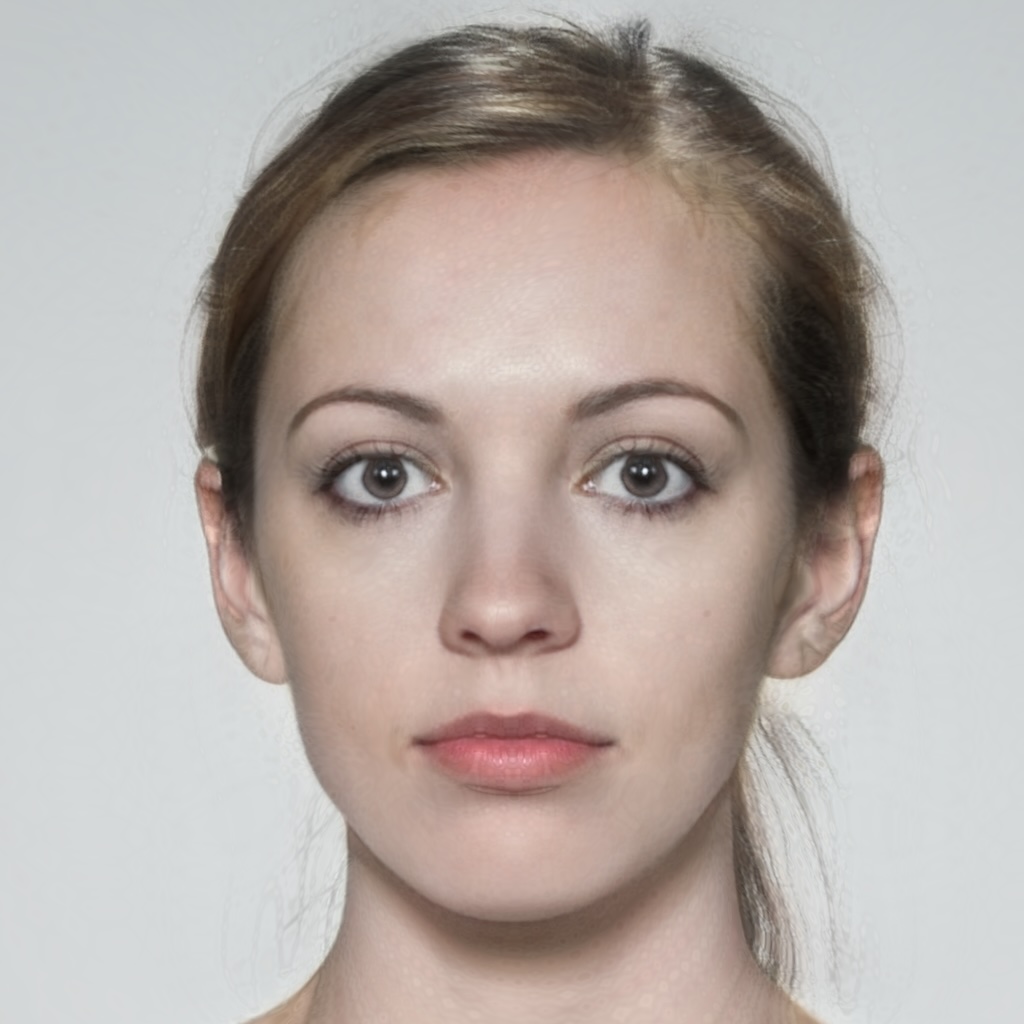}
  \includegraphics[width=0.19\columnwidth]{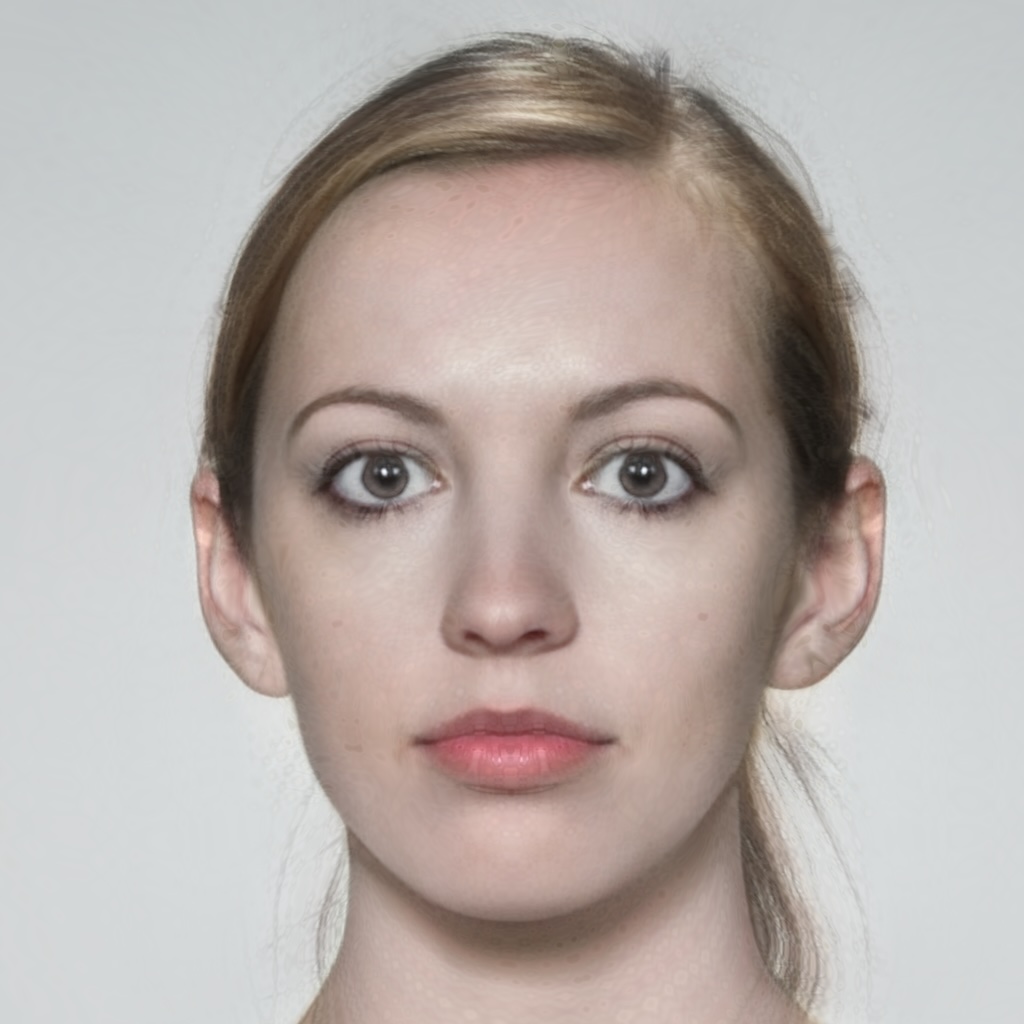}
  \vspace{-0.3cm}
  \caption{Morphing comparisons of our method and generative approaches (neural warping + linear blending, diffAE, neural warping + diffAE, and StyleGAN3). Columns 1 and 5 are the target faces, while the three middle columns are blendings for $t=0.25, 0.5, 0.75$.
  The original images are the ends of Line 1.}
  \label{fig:morphing_comparisons}
  \vspace{-0.7cm}
\end{figure}

We assess our approach regarding the visual quality of both warping/blending of faces. Fig~\ref{fig:morphing_comparisons} shows our neural warping with linear blending, diffAE with FFHQ alignment, neural warping and diffAE, and StyleGAN3 with FFHQ alignment.
Note that unlike StyleGAN3, diffAE provides a close, although blurred, reconstruction of the target.

In Fig~\ref{fig:morphing_comparisons}, diffAE (Line 2) produces a shadow in the forehead/hair transition area for images with $t=0.5,0.75$.

\noindent It also creates a hole in the subject's left earlobe. These issues are missing when using our neural warping for alignment (Line 3).
Another point of note is the face similarity between neural warping + diffAE (Line 3) and neural warping + linear blending (Line 1).
This is due to the temporal coherence added by time-dependent alignment given by the warping.
Thus, the generative morphing produces intermediate faces closer to the targets when compared to employing FFHQ alignment.
Moreover, since StyleGAN3 does not reproduce the target faces from the latent code projections, the blendings are generating faces unrelated to the originals.

\begin{figure}[hh]
    \text{\footnotesize diffAE}\\
    \includegraphics[width=\columnwidth]{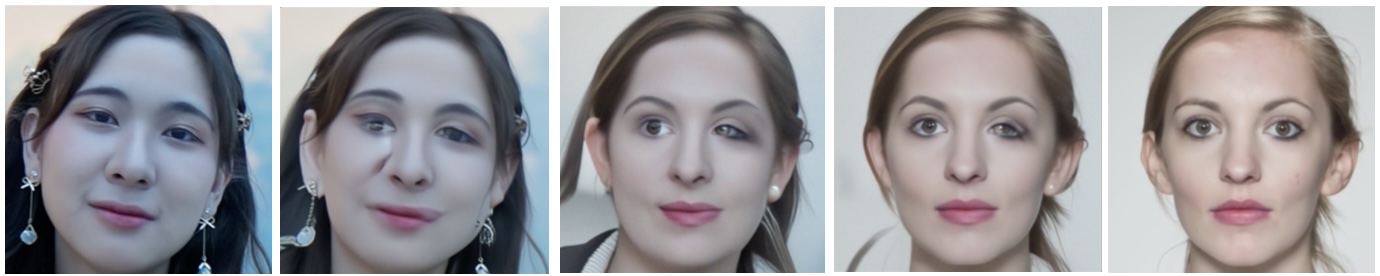}
    \text{\footnotesize Neural warping + linear blending [Ours]}\\
    \includegraphics[width=\columnwidth]{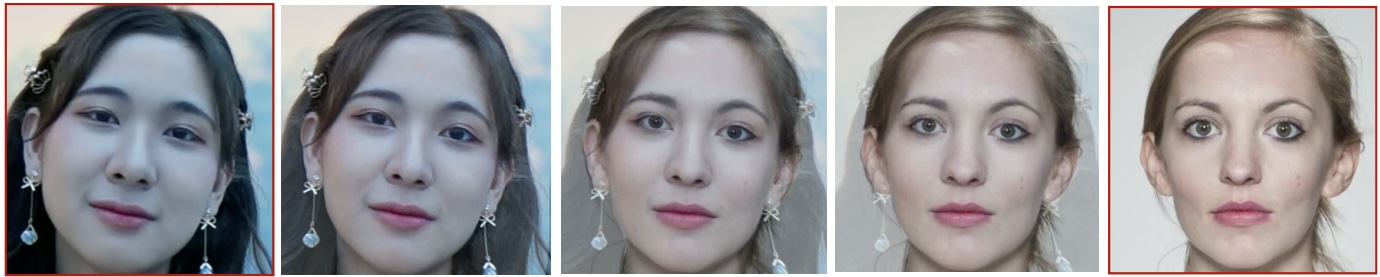}
    \text{\footnotesize Neural warping + diffAE [Ours]}\\
    \includegraphics[width=\columnwidth]{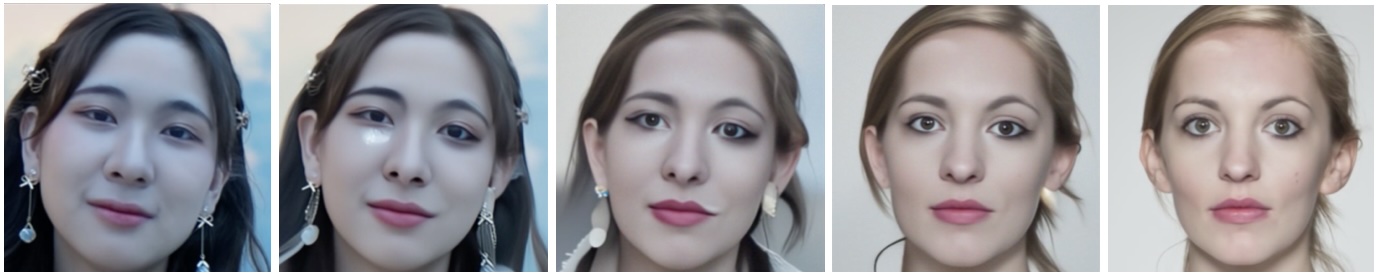}
    \caption{Morphings between unaligned faces. Columns 1 and 5 are the target images (in red). Columns 2, 3, and 4 are morphings at $t=0.25, 0.5, 0.75$.
    Line 1 shows diffAE blending where the target images were cropped to contain mostly the face. Line 2 shows our neural warping and linear blending, and Line 3 shows our neural warping and diffAE blending.
    Note that the diffAE adds a blurring to the reconstructed images.}
  \label{fig:roto-translations}
  \vspace{-0.2cm}
\end{figure}

As shown in Fig~\ref{fig:diffae_pure_warp}, FFHQ alignment is necessary for interpolating faces; otherwise, it produces visual artifacts. This is because generative models do not perform warping of facial features; instead, they blend them. Thus, we cannot use such methods for morph faces in different poses. However, we observe that we can use neural warping for this task.
Fig~\ref{fig:roto-translations} displays morphings between faces in different positions. As expected, diffAE cannot blend the faces (Line 1). Thus, we consider our neural warping (Line 2) and pass it as input to diffAE, resulting in better interpolations (Line 3).

\begin{figure}[hh]
\centering
  \includegraphics[width=0.93\columnwidth]{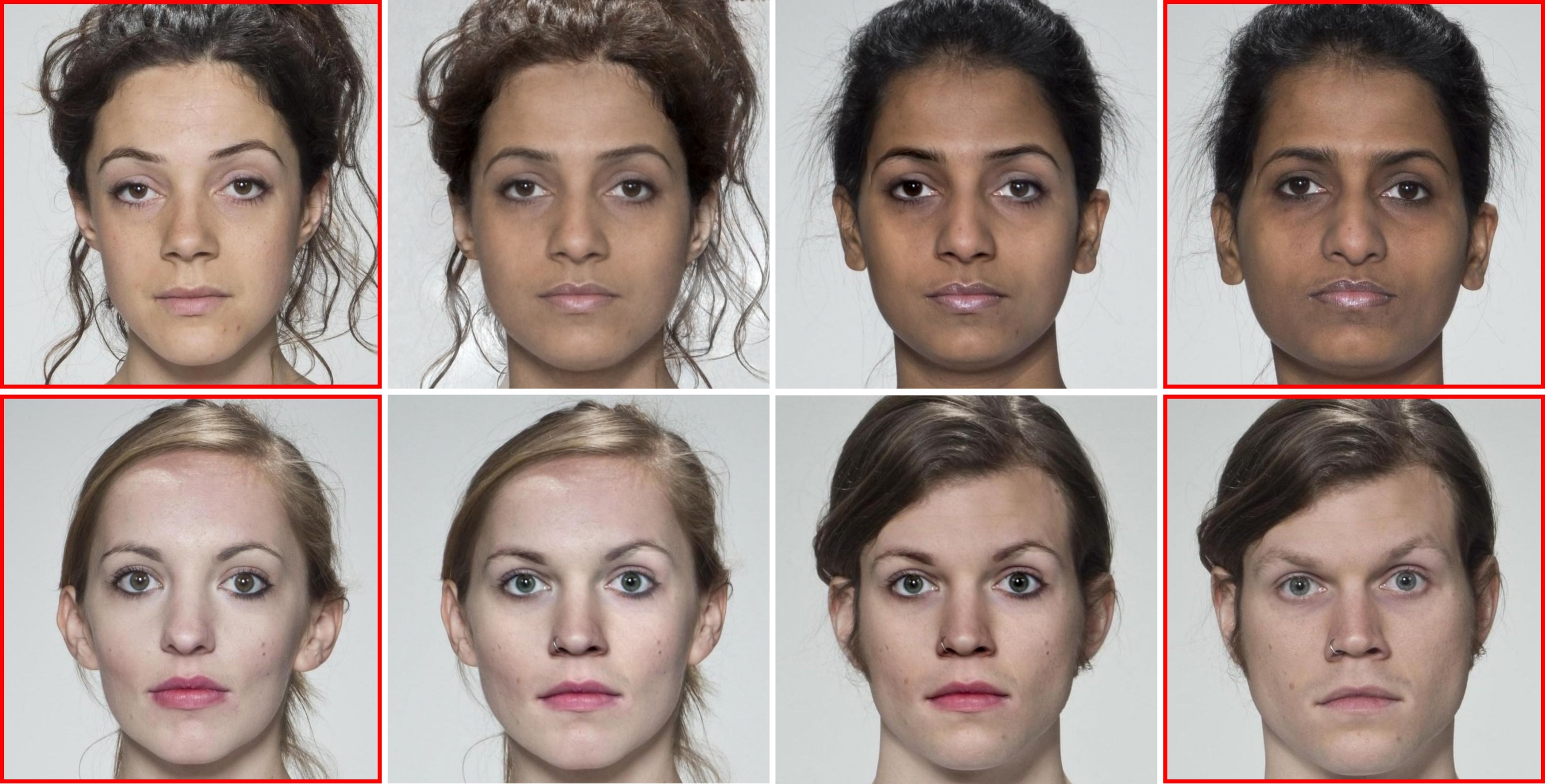}
  \footnotesize{\text{\,\quad Source}\quad\quad\;\;\text{Seamless mix}\qquad\text{Seamless mix}\qquad\;\;\,\text{Target}\;\;\;}
  \vspace{-0.2cm}
  \caption{Morphings between faces of different ethnicities (Line 1) and genders (Line 2). Columns 1 and 4 show the target faces. We blend them using seamless mix, at $t=0.5$, and either the source image as base (Column 2), or the target image as base (Column 3). In both case we employed our neural warping/blending.}
  \label{fig:unusual-morphings}
  \vspace{-0.2cm}
\end{figure}

Our approach also handles faces with varying genders/ethnicities, resulting in high-quality morphings, as shown in Fig~\ref{fig:unusual-morphings}. It shows that our method learns effective alignments, enabling seamless blendings to preserve details. Morphing in this context is challenging due to feature alignment, and blending skin colors/textures~\cite{neubert2018stirtrace}. Additional examples are shown in the supp. material.

\pagebreak

\begin{figure}[h!]
  \centering
  \includegraphics[width=0.95\columnwidth]{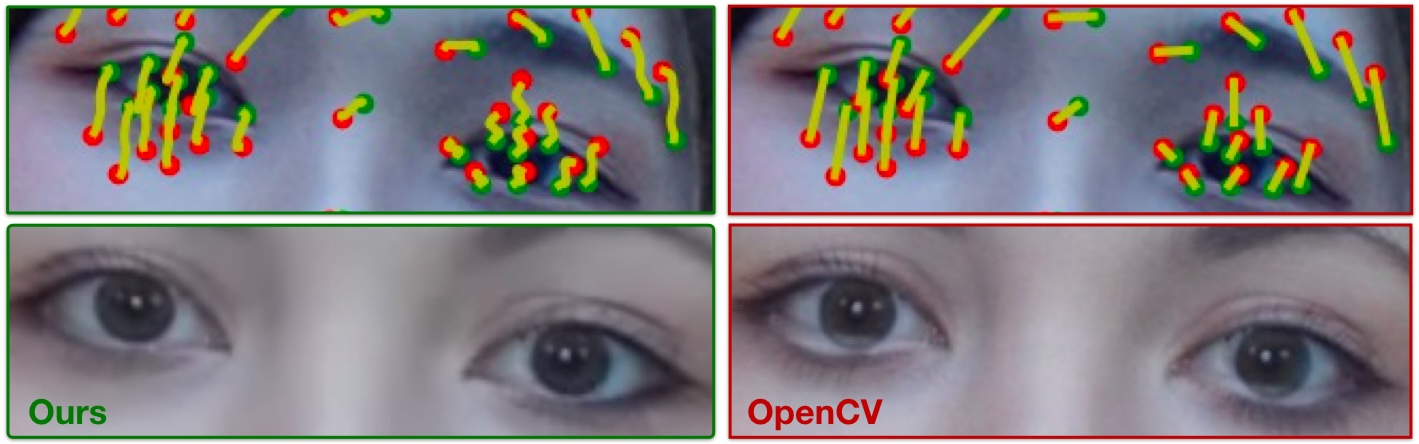}
  \caption{Comparison between our warping (left) and OpenCV (right) and the resulting blendings (bottom row).}
  \label{fig:warping-paths}
  \vspace{-0.2cm}
\end{figure}

Fig~\ref{fig:warping-paths} shows an example of the warping paths (top) and the linear blending of both images (bottom) created by our method (left) and classic OpenCV warping (right). The creation of a non-linear path lead to a better alignment, and thus a blending with less ghosting artifacts.

\begin{figure}[h!]
  \centering
  \includegraphics[width=\columnwidth]{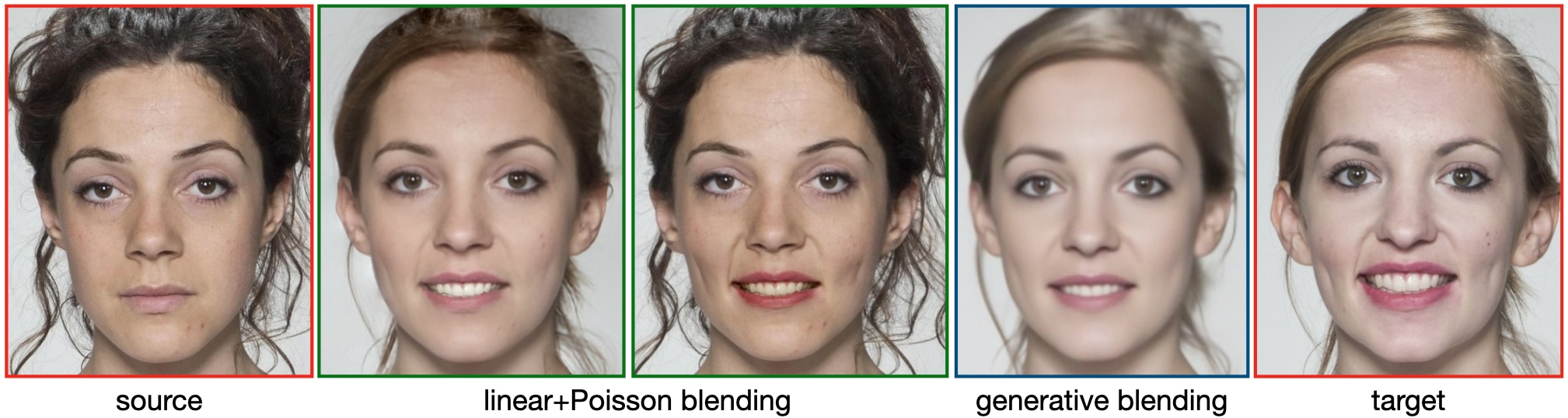}
  \includegraphics[width=\columnwidth]{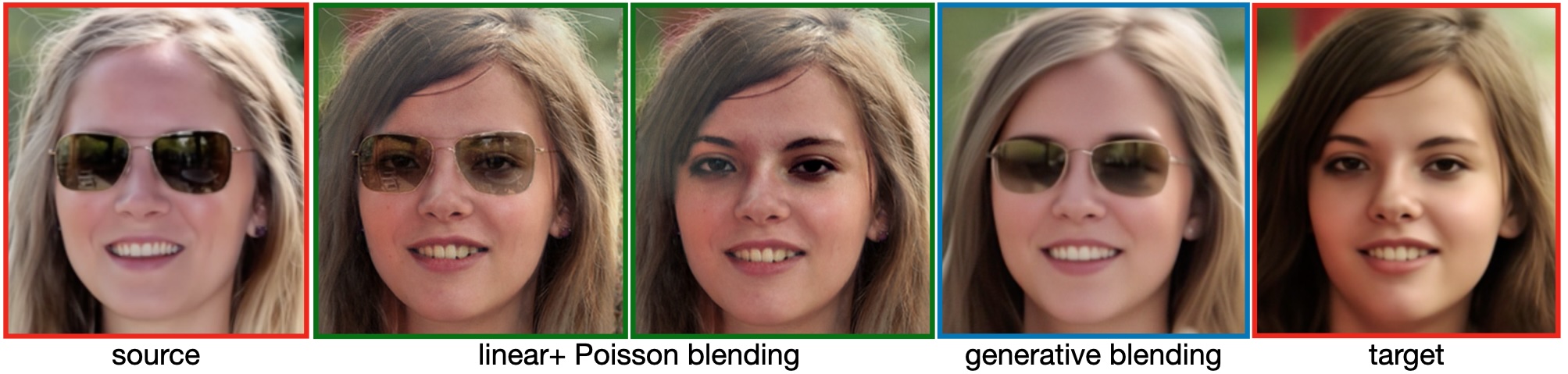}
  \vspace{-0.7cm}
  \caption{Morphings between subjects with different expressions (top) and, with partial occlusion and faces in the wild (bottom).}
  \label{fig:expressions-occlusion}
\end{figure}

\begin{figure}[h!]
  \centering
  \includegraphics[width=\columnwidth]{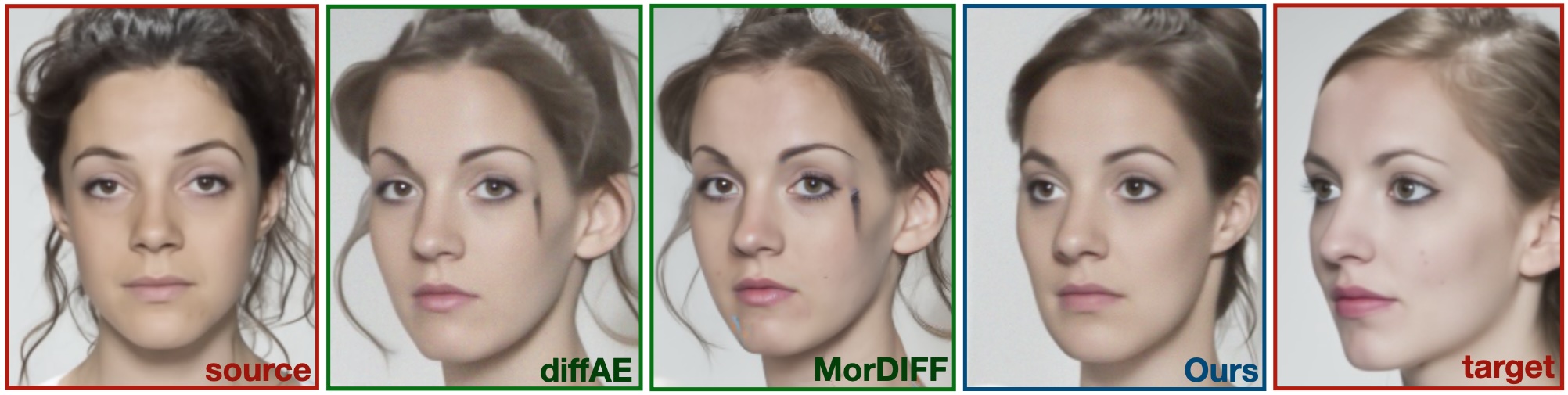}
  \vspace{-0.7cm}
  \caption{Morphings between faces with different poses.}
  \label{fig:poses}
\end{figure}

Additionally, our method handles morphing between faces with different expressions (Fig~\ref{fig:expressions-occlusion}, top row), partial occlusions (Fig~\ref{fig:expressions-occlusion}, bottom row) and, poses (Fig~\ref{fig:poses}). In Fig~\ref{fig:expressions-occlusion}, we employ linear, our neural Poisson, and diffAE blendings, while in Fig~\ref{fig:poses} we compare diffAE and MorDiff~\cite{MorDIFF} with our generative blending.

\vspace{-0.5cm}
\subsubsection*{Feature transfer using neural warping/blending}
\begin{figure}[hh]
  \centering
  \includegraphics[width=0.70\columnwidth,valign=c]{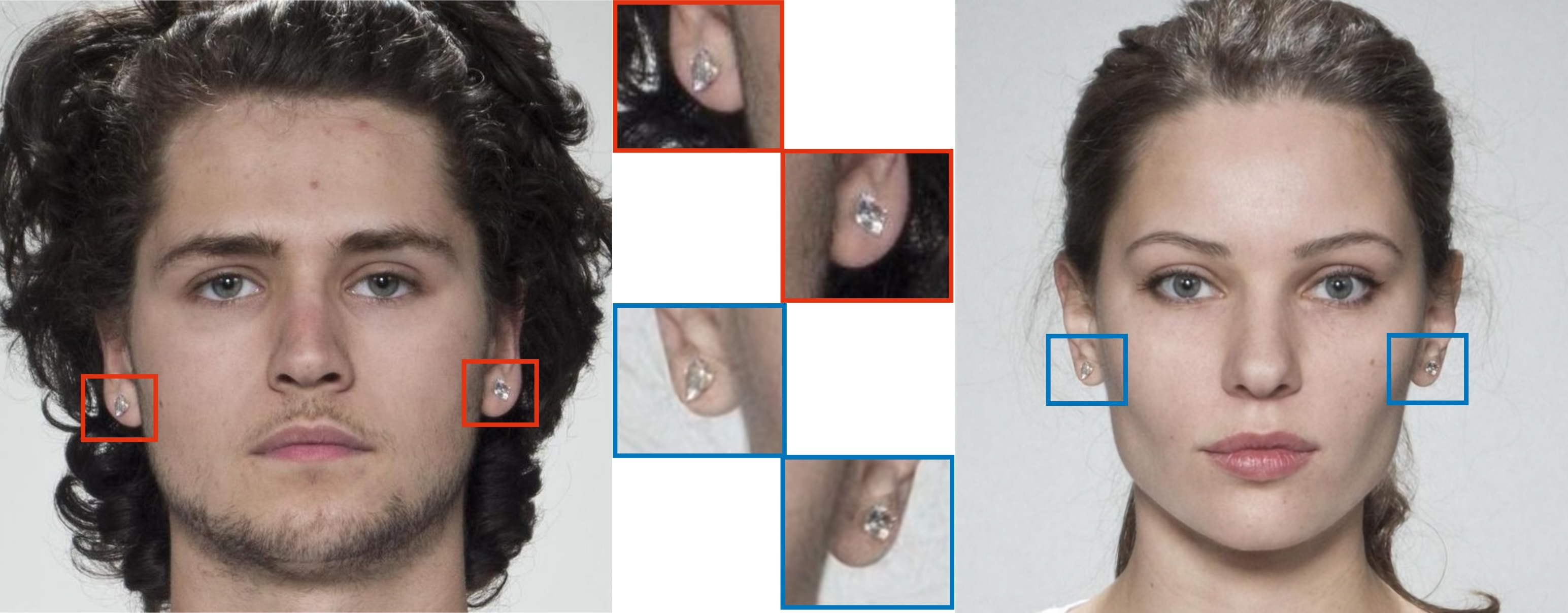}
  \includegraphics[width=0.28\columnwidth,valign=c]{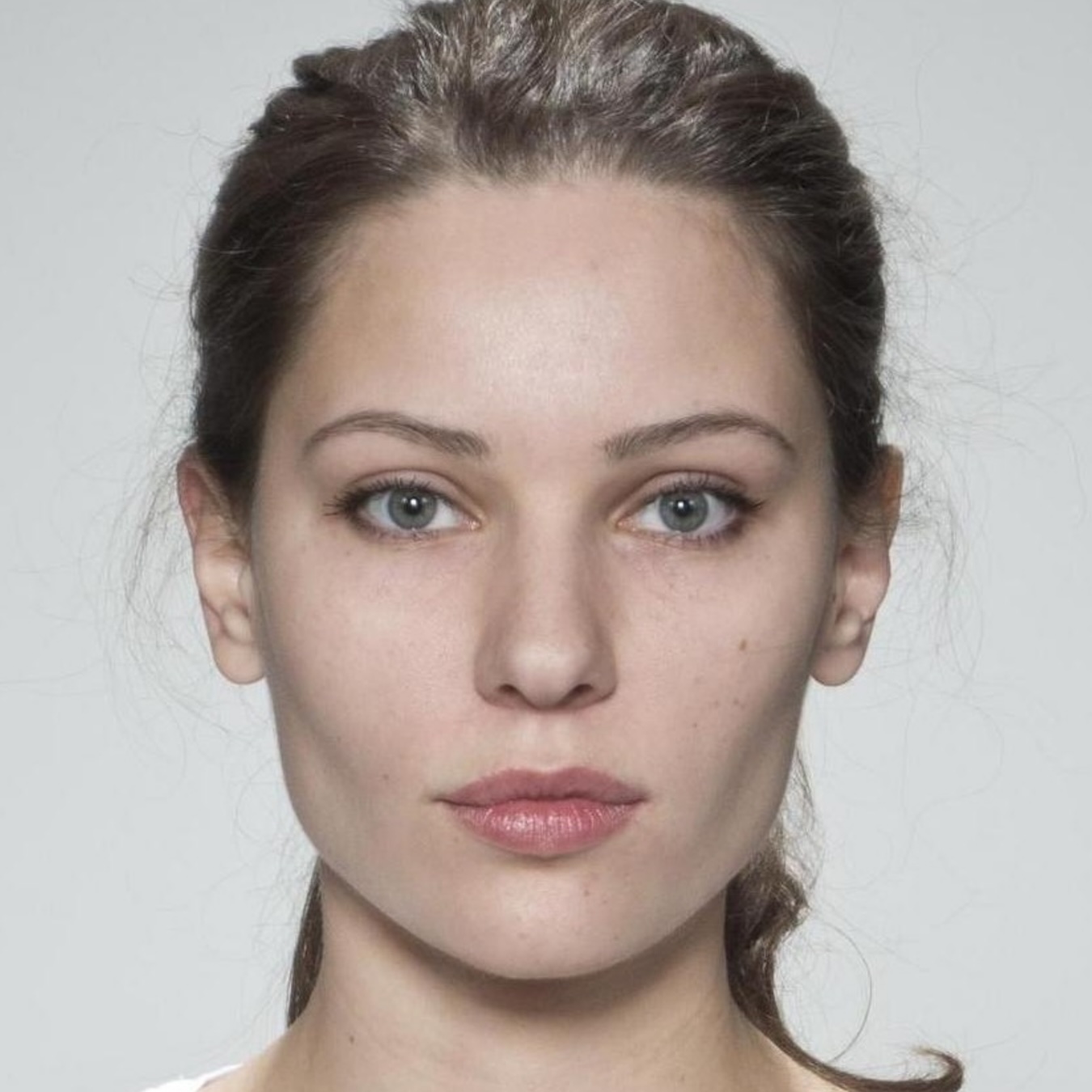} \\
  \includegraphics[width=0.70\columnwidth,valign=c]{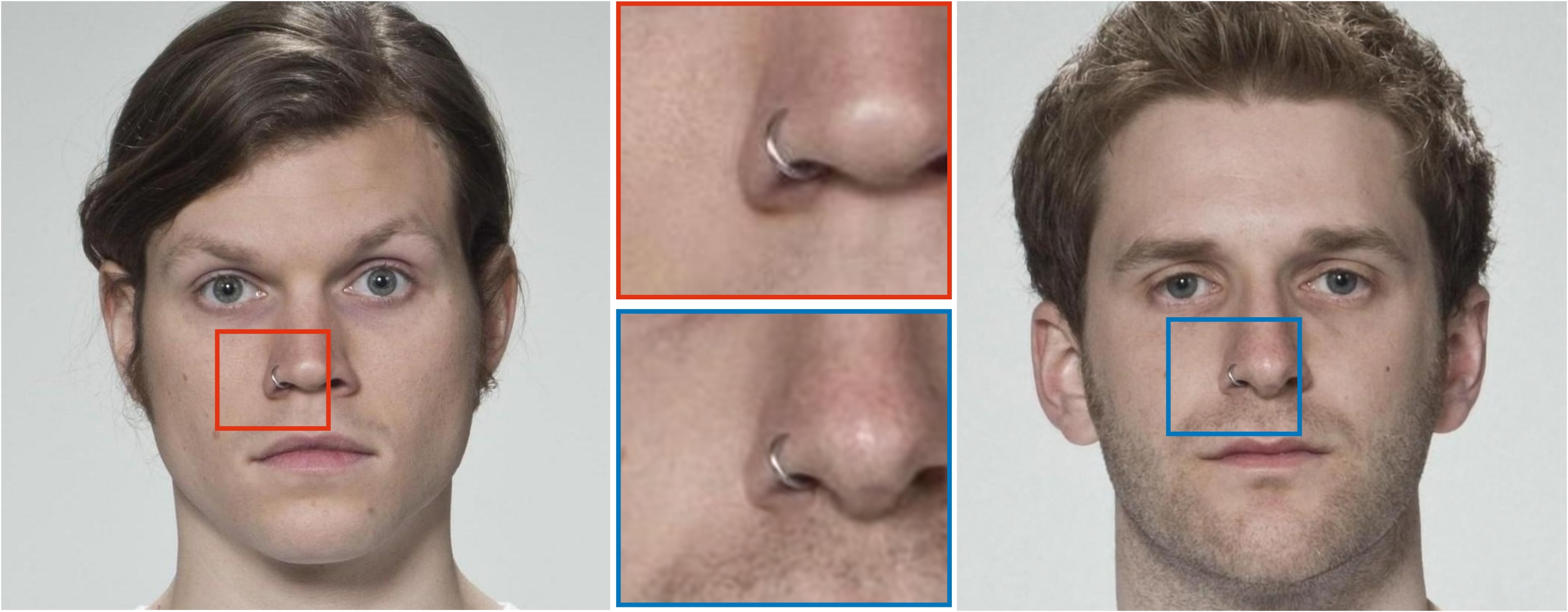}
  \includegraphics[width=0.28\columnwidth,valign=c]{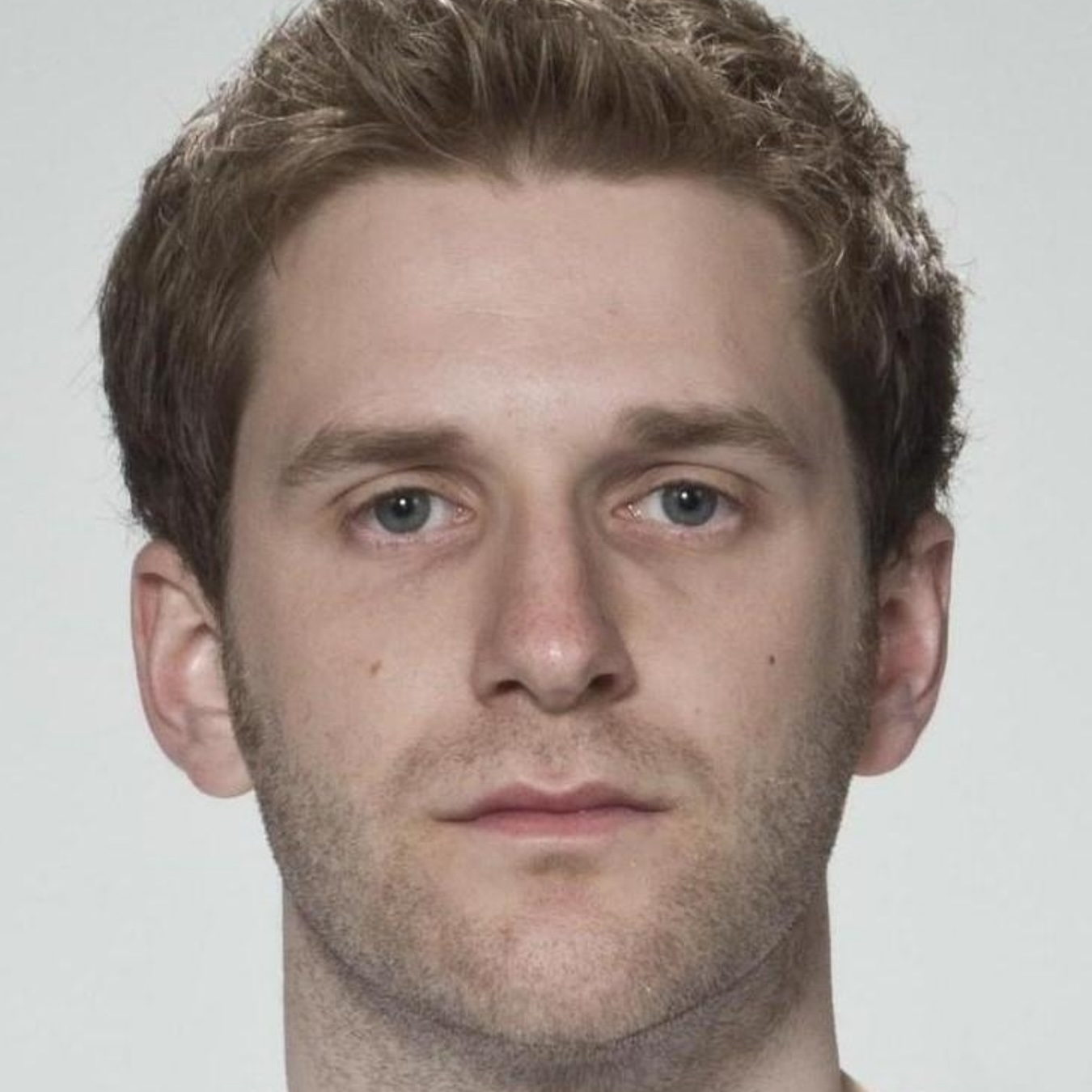}
  \includegraphics[width=0.70\columnwidth,valign=c]{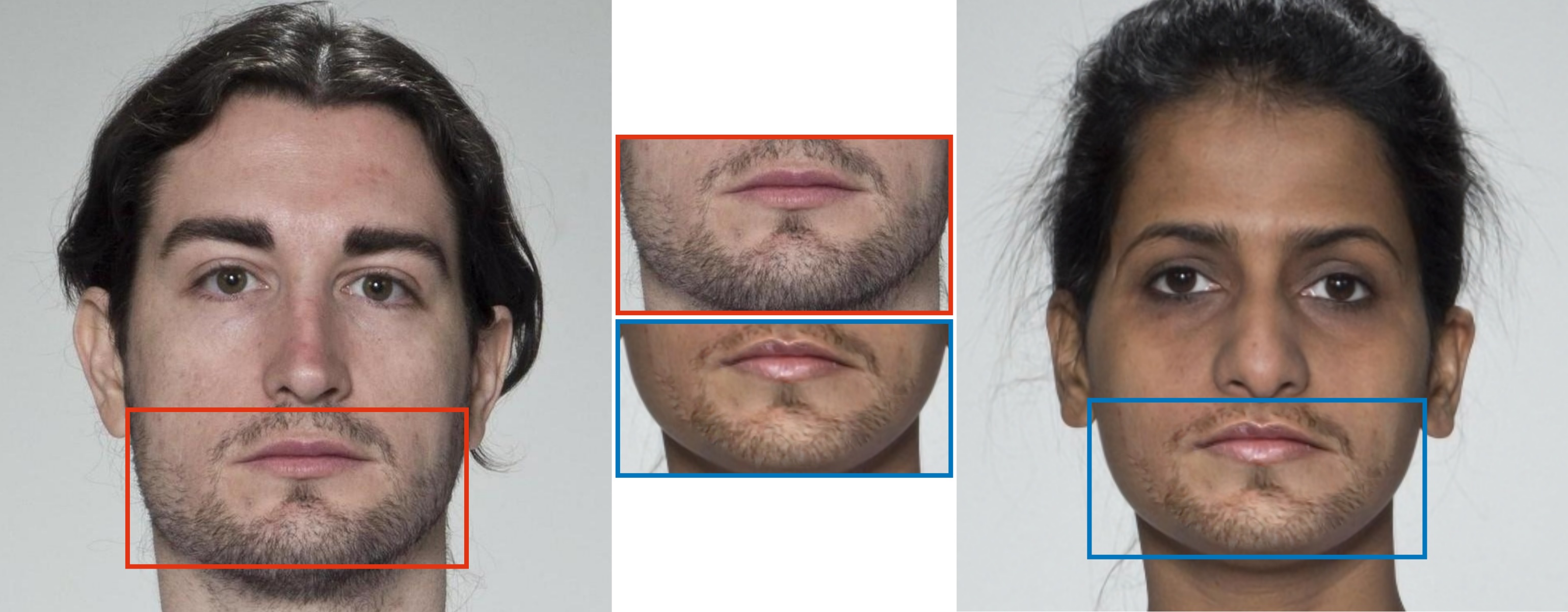}
  \includegraphics[width=0.28\columnwidth,valign=c]{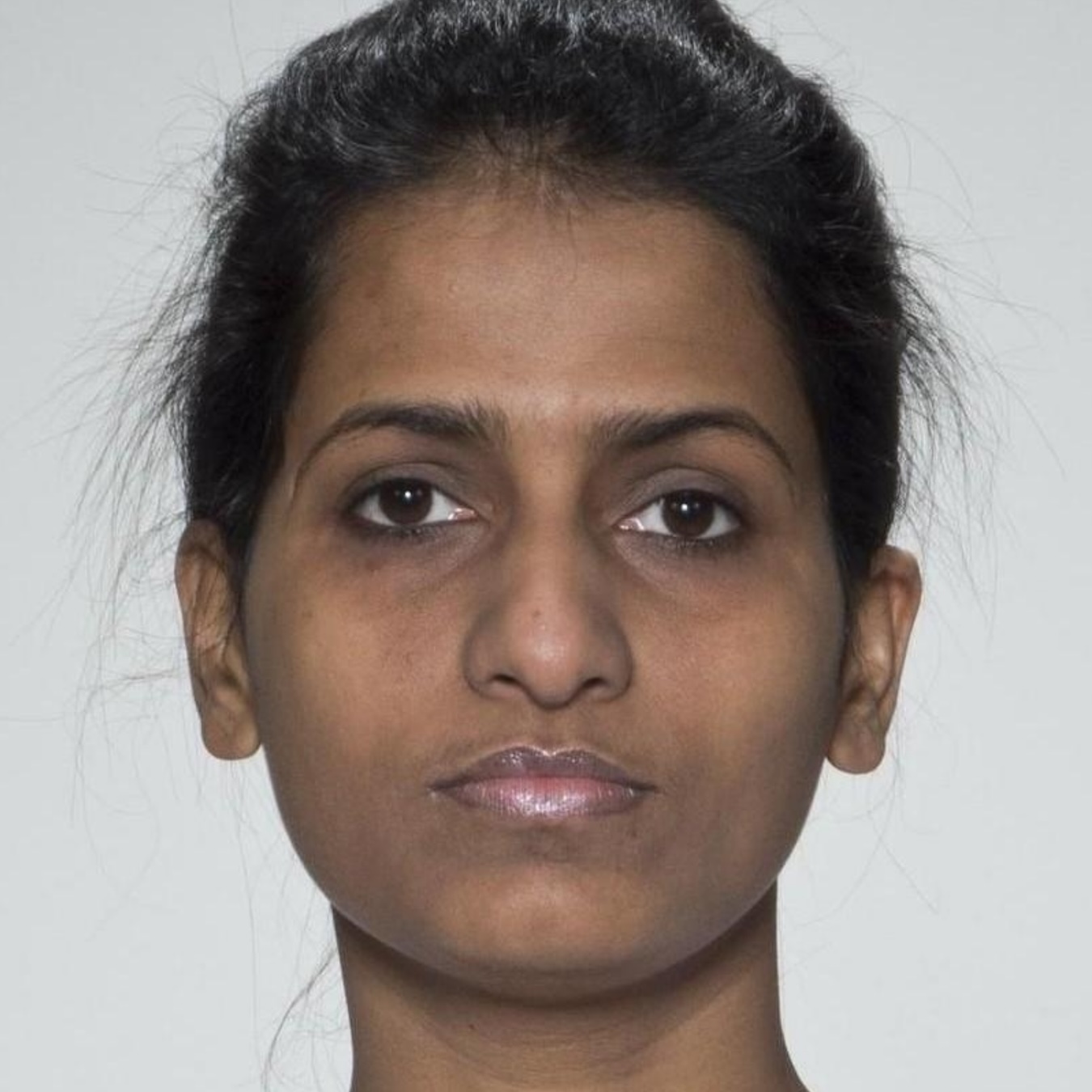}
  \caption{Transference of features between images. Columns 1 and 4 present the source/target faces, Column 2 shows the region containing the desired feature(s) and Column 3 shows the feature(s) transferred to the target image.}
  \label{fig:feature-transfer}
\end{figure}

\vspace{-0.18cm}
Our method can be used to transfer features between faces, as shown in Fig~\ref{fig:feature-transfer}. To transfer features, we train a warping between two faces, select the region with a desired feature, warp the source face to match the target face, and blend only that region in the gradient domain (Sec~\ref{s-blending-grad-domain}).

\vspace{-0.1cm}
\subsection{Quantitative comparisons}
\label{sec:quantitative-results}
\vspace{-0.2cm}
We compare our approach with StyleGAN3, diffAE, and the classic OpenCV procedure. We assess the performance of our neural warping with different blendings: linear, seamless cloning, and mixing. From the $102$ images of the FRLL dataset~\cite{debruine2017frll}, we generated $1220$ morphings following the protocol in~\cite{sarkar2020landmarks}, thus resulting in morphings of similar faces (i.e., same gender, similar ethnicity). Moreover, we used the FFHQ alignment, provided as a stand-alone script by the diffAE\footnote{\scriptsize \url{https://github.com/phizaz/diffae/blob/master/align.py}} to post-process the images (both original and morphed), cropping and resizing them to $256\!\times\! 256$ pixels.

To assess the visual fidelity, we used \textit{Fréchet inception distance} (FID)~\cite{heusel2017fid} and \textit{learned perceptual image patch similarity} (LPIPS)~\cite{zhang2018perceptual}. FID is employed by generative methods to measure the proximity between the distributions of real and generated images~\cite{lucic2018fid}. Lower FID values mean that the distributions are close, thus the generated images are close to the original. LPIPS calculates the similarity of two images by splitting them into patches passed through an image network and measuring their activation similarity. The final LPIPS of the two images is the mean LPIPS of their patches. The FID metric is calculated using \texttt{pytorch-fid v0.3.0}~\cite{seitzer2020fid}, while LPIPS uses \texttt{lpips v0.1.4}~\cite{zhang2018perceptual}.

Table~\ref{tab:fid-lpips} shows the FID and LPIPS scores of the techniques. Here, the target images are I$_0$ and I$_1$, and I is the morphing between then at $t=0.5$. We split the LPIPS score between (I$_0$, I) and (I, I$_1$), since the seamless-\{clone,mix\} blending transfers the warped features of I$_1$ to I$_0$, thus leading to a higher similarity between (I$_0$, I) compared to (I, I$_1$).
Our warping with seamless mix blending achieves higher visual fidelity according to FID and better perceptual similarity to the source image, as indicated by LPIPS~(I$_0$, I), while our method with linear blending obtained LPIPS~(I, I$_1$) comparable to generative methods.

\begin{table}[hh]
\centering
\caption{FID and LPIPS for OpenCV, StyleGAN3/diffAE, and our warping with different blendings.}
\vspace{-0.2cm}
\footnotesize{
\begin{tabular}{lrrr}
\toprule
\textbf{Morphing Type} & \!\!\textbf{\small FID} $\downarrow$ & \!\!\!\!\textbf{\small LPIPS} (I$_0$, I) $\downarrow$ & \!\!\!\!\textbf{\small LPIPS} (I, I$_1$) $\downarrow$\\ \midrule
  OpenCV                 & 68.234 & 0.275  & 0.281                  \\
  StyleGAN3              & 35.653 & 0.174  & 0.173                  \\
  diffAE                 & 41.356 & 0.183  & 0.186                  \\
  Ours (linear)          & 31.950 & 0.158  & \textbf{0.164}         \\
  Ours (S. Clone)        & 25.290 & 0.093  & 0.234                  \\
  Ours (S. Mix)          & \textbf{22.604} & \textbf{0.081} & 0.241 \\
  Ours (diffAE)          & 40,224 &  0.175 & 0.176                  \\
  \bottomrule
\end{tabular}}
\label{tab:fid-lpips}
\end{table}

The results in Table~\ref{tab:fid-lpips} show that by improving the warping, the morphing quality increases (see Lines 1 and 4) such that the resulting images surpass generative methods w.r.t. perceptual metrics. Further improvements in the blending lead to morphings with a natural appearance, and more similar to one of the target images. Additionally, see the morphing-attack-detection (MAD) results in supp. material.

\vspace{-0.6cm}
\paragraph{Hardware used} The images and morphing networks were trained using an NVIDIA GeForce RTX 3090 GPU, with 24GB of memory. The system has a AMD Ryzen Threadripper PRO 5965WX CPU and 256GB of DDR4 memory.

\vspace{-0.52cm}
\paragraph{Ethical Issues} One of the problems with face morphing is its use to create fake appearances for official purposes or defamation of individuals. This raises concerns in both the community and the authors. We hope that by exposing our method to the community, we ensure that other colleagues can create detection models to counteract such threats.

\vspace{-0.52cm}
\paragraph{Limitations} Our method builds a functional representation of the warping to align the features of two faces. It encodes the direct/inverse transformations required in morphing in a single network. Thus, requesting the learning of a non-invertible transformation may lead to inconsistencies. For example, if a particular region of the image collapses during warping, it cannot be inverted.
Nevertheless, we can still represent such a transformation with the inverse part of the neural warping or using its direct counterpart.

\vspace{-0.3cm}
\section{Conclusions}
\vspace{-0.2cm}
We proposed a face morphing by leveraging coord-based neural networks. We exploited their smoothness to add energy functionals to warp and blend target images seamlessly without the need of derivative discretizations.

Our method ensures continuity in both space and time coordinates, resulting in a smooth transition between images. By operating on a smooth representation of the underlying images, we eliminate the need for pixel interpolation/resampling.
The seamless blending of the target images is achieved through the integration of energy functionals, ensuring their harmonious clone. The resulting morphs exhibit a high level of visual fidelity and maintain the overall structure and appearance of the target faces, even when morphing between different genders or ethnicities. Finally, our neural warping offers a versatile framework being easily integrated with generative methods, opening up possibilities for applications in computer graphics and digital entertainment.

In the future, we aim to create morphing datasets using our method to improve MAD models, thus limiting any potential negative impact. We intend to extend it to other type of images, and operate on surfaces as well~\cite{novello2023neural,liu2022learning,novello2022,sitzmann2020implicit}.

\vspace{-0.55cm}
\paragraph{Acknowledgements} The authors would like to thank projects UIDB/00048/2020\footnote{DOI: \url{https://doi.org/10.54499/UIDB/00048/2020}} and 150991/2023-1 (CNPq and FAPERJ) for partially funding this work. We also thank Daniel Perazzo for helping with the landmark editing~UI.

%%%%%%%%%%%%%%%%%%%%%%%%%%%%%%%%%%%%%%%%%%%%%%%%%%%%%%%%%%%%

%===================================================
\bibliographystyle{ieeenat_fullname}
\bibliography{refs}
%======
\end{document}

\maketitle

\begin{figure*}[htbp]
  \centering
  \includegraphics[width=1.0\textwidth]{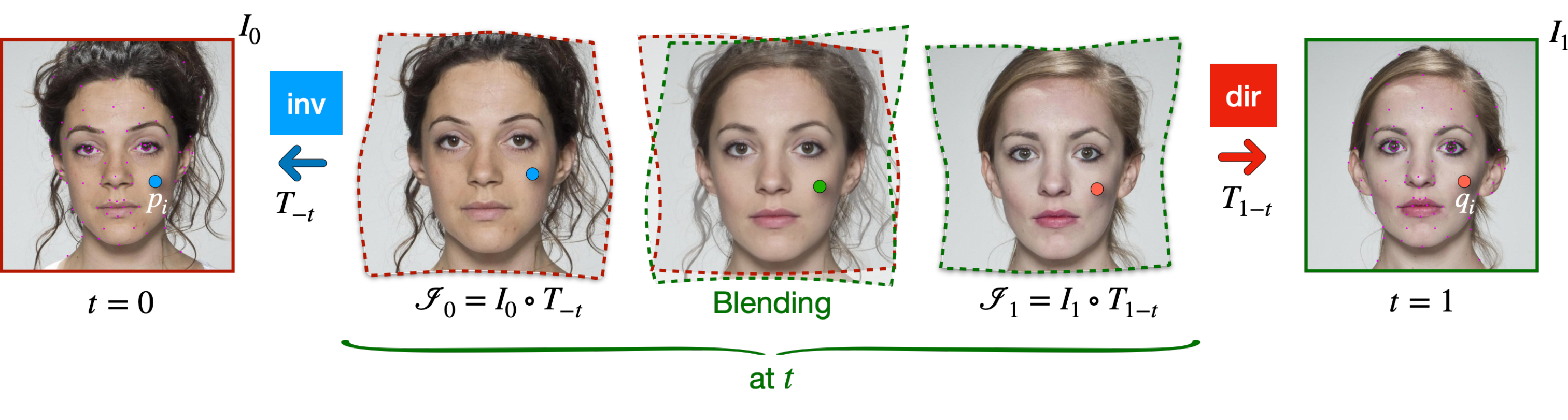}
  \vspace{-0.3cm}
  \caption{Overview of the proposed morphing approach. Let $\text{I}_0$ and $\text{I}_1$ be two face images represented as sinusoidal MLPs. We position a series landmarks (illustrated as blue and red points in $\text{I}_0$ and $\text{I}_1$), which are inputs of a warping network, also a (shallow) sinusoidal MLP. The result of this step is a transform $T$, that when composed with $I_{\{0,1\}}$ results in are two aligned and warped images $\mathscr{I_{\{0,1\}}}$. Afterwards, the warpings may be used to create different blendings. We've focused on linear, Poisson and generative blendings.}% Notice that performing an inverse transform on $\mathscr{I}_0$, by composing $I_0 \circ T_{-t}$, we obtain the original image $I_0$. Additionally, performing a direct transform by composing $\I_1 \circ T_{1-t}$, we obtain $I_1$.}
\end{figure*}

\section{Morphing attack detection performance}
\label{sec:mad_performance}
In addition to the quality metrics of morphings, we present our results from the perspective of morphing attack detection (MAD) performance metrics.
This experiment was conducted under the same protocol defined in Sec.~4.2 in the main work.
The morphing type is different for protocols (but follow the same identity pairing list) while the original images remain the same.
The morphed images were tested using the MAD approach proposed in~\cite{medvedev:2022}.

\begin{figure}[H]
  \centering
  \includegraphics[width=0.9\columnwidth]{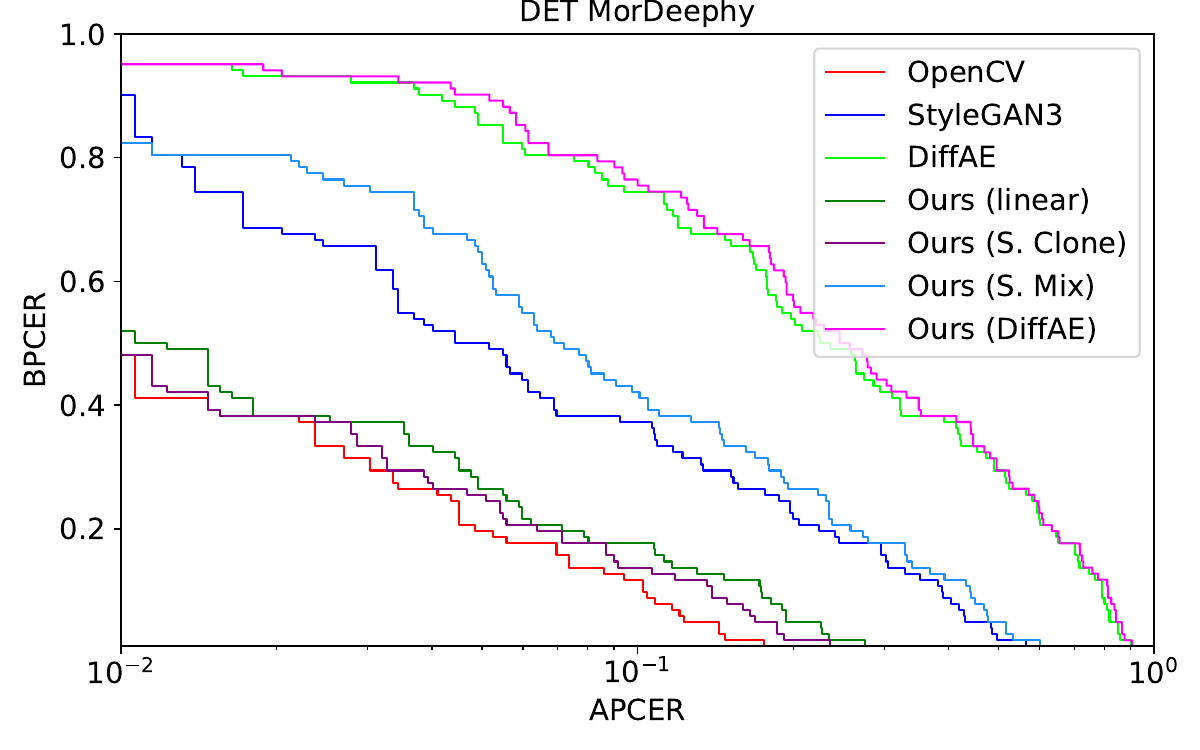}
  \vspace{-0.3cm}
  \caption{MAD performance of MorDeephy approach \cite{medvedev:2022} on various types of face morphing.}
  \label{fig:mdf_performance}
\end{figure}

The primary metric for assessing the detection performance of single image morphing relies on the relationship between the Bona fide Presentation Classification Error Rate (BPCER) and the Attack Presentation Classification Error Rate (APCER)(according to ISO/IEC 30107-3~\cite{REFERENCE_OF_ISO_30107}), which may be plotted as a Detection Error Trade-off (DET) curve. We report the performance by $BPCER@APCER$ = $(0.1, 0.01)$
(see Table~\ref{tab:APCER_BPCER_results}) and DET curve (see Fig~\ref{fig:mdf_performance}).

% TABLE benchmarks APCER@BPCER
\begin{table}[H]
\centering
\caption{
APCER@BPCER = (0.1, 0.01) of MAD method \cite{medvedev:2022} in different protocols. Lower means better detector performance against the morphing method.
\vspace{-0.3cm}
}
\footnotesize{
\begin{tabular}{l|rr}
\toprule
\multirow{2}{*}{\textbf{Morphing Type}} & \multicolumn{2}{c}{$BPCER@APCER=\delta$} \\ \cline{2-3}
                               & $\delta = 0.1$     & $\delta = 0.01$     \\ \midrule
OpenCV                         & 0.117              & 0.951               \\
StyleGAN3                      & 0.264              & 0.666               \\
DiffAE                         & 0.745              & 0.951               \\
Ours (Linear)                  & 0.176              & 0.519               \\
Ours (S. Clone)                & 0.137              & 0.500               \\
Ours (S. Mix)                  & 0.421              & 0.824               \\
Ours (diffAE)                  & 0.754              & 0.950               \\ \bottomrule
\end{tabular}}
\label{tab:APCER_BPCER_results}
\end{table}

%Discussion
By evaluating the MAD performance, we can gauge the detection complexity inherent in our morphs. Note that these results come with a disclaimer: the effectiveness of the chosen detector is not universal and may be influenced by certain unknown biases inherent to the detector itself~\cite{medvedev:2022}.

Based on the results, the ``Cloning'' method demonstrates comparable performance to ``OpenCV''. The ``Mix'' morphs are harder to detect than other interpolation based morphs and they appear to be more challenging than StyleGAN3. DiffAE morphs are challenging to detect and indeed represent an unseen type of attack for this detector instance.

\section{Ablation Studies}
We present experiments showing the effects of each loss term (Sec.~\ref{sec:ablation-loss}) and network width (Sec.~\ref{sec:ablation-netwidth}). For these experiments, we employed the landmarks detected using DLib~\cite{kazemi2014cvpr,sagonas2016}, and images from the provided with the FRLL dataset~\cite{debruine2017frll}. Additionally, we used the network initialization procedure proposed in~\cite{sitzmann2020implicit}.

\subsection{Loss function study}
\label{sec:ablation-loss}
In this section, we measure the impact of each constraint of our loss function in the warping/blending. Recall that our loss function $\mathscr{L}(\theta) = \mathscr{W}(\theta)+\mathscr{D}(\theta)+\mathscr{T}(\theta)$ (Eq. 2 in the paper), has three terms: Warping ($\mathscr{W}$), data ($\mathscr{D}$), and thin-plate energy ($\mathscr{T}$).
The warping constraint is composed by the identity and inverse constraints (Eq. 3 in the paper):
\begin{align}\small
    \mathscr{W}\!(\theta)\!\!=\!\!\!\underbrace{\int\limits_{\R^2}\!\!\norm{\gt{T}(x,0)\!-\!x}^2\!\!dx}_{\text{Identity constraint}} \!+\!\!\!\! \underbrace{\int\limits_{\R^2\times \R}\!\!\!\!\!\norm{\gt{T}\big(\gt{T}(x,t),-t\big)\!\!-\!x}^2\!\!\!dxdt}_{\text{Inverse constraint}}.
\end{align}

Table~\ref{tab:fid-lpips-loss} presents the FID and LPIPS (I$_0 \rightarrow$ I, and I $\rightarrow$ I$_1$) for these experiments. We used linear blending for all cases, and hidden-layer width of 128 neurons.

\begin{table}[H]
    \centering
    \small
    \caption{FID and LPIPS for our warping with linear blending with different loss terms deactivated.}
    \begin{tabular}{lrrr}
        \toprule
        Experiment      & \textbf{FID} $\downarrow$ & \textbf{LPIPS} (I$_0$, I) $\downarrow$ & \textbf{LPIPS} (I, I$_1$) $\downarrow$ \\
        \midrule
        No $\mathscr{D}$ & 113.599 & 0.215 & 0.246 \\
        No $\mathscr{T}$ & 499.289 & 0.594 & 0.607 \\
        No Id.           & 118.201 & 0.306 & 0.314 \\
        No Inv.          & 113.224 & 0.278 & 0.285 \\
        Original         &  31.950 & 0.158 & 0.164 \\
        \bottomrule
    \end{tabular}
    \label{tab:fid-lpips-loss}
\end{table}

From Table~\ref{tab:fid-lpips-loss} we conclude that $\mathscr{T}$ has the most impact in our results, followed by the identity constraint (Id.), $\mathscr{D}$, and the inverse constraint (Inv.). $\mathscr{T}$ regularizes the loss, ensuring that the alignment is smooth during training, while Id. ensures that the warping will match the original landmark positions at $t=0$ and $t=1$. Additionally, $\mathscr{D}$ ensures that the source and target landmarks match for every $t$.

\subsection{Network width}
\label{sec:ablation-netwidth}
\begin{figure}[H]
  \centering
  \includegraphics[width=0.22\columnwidth]{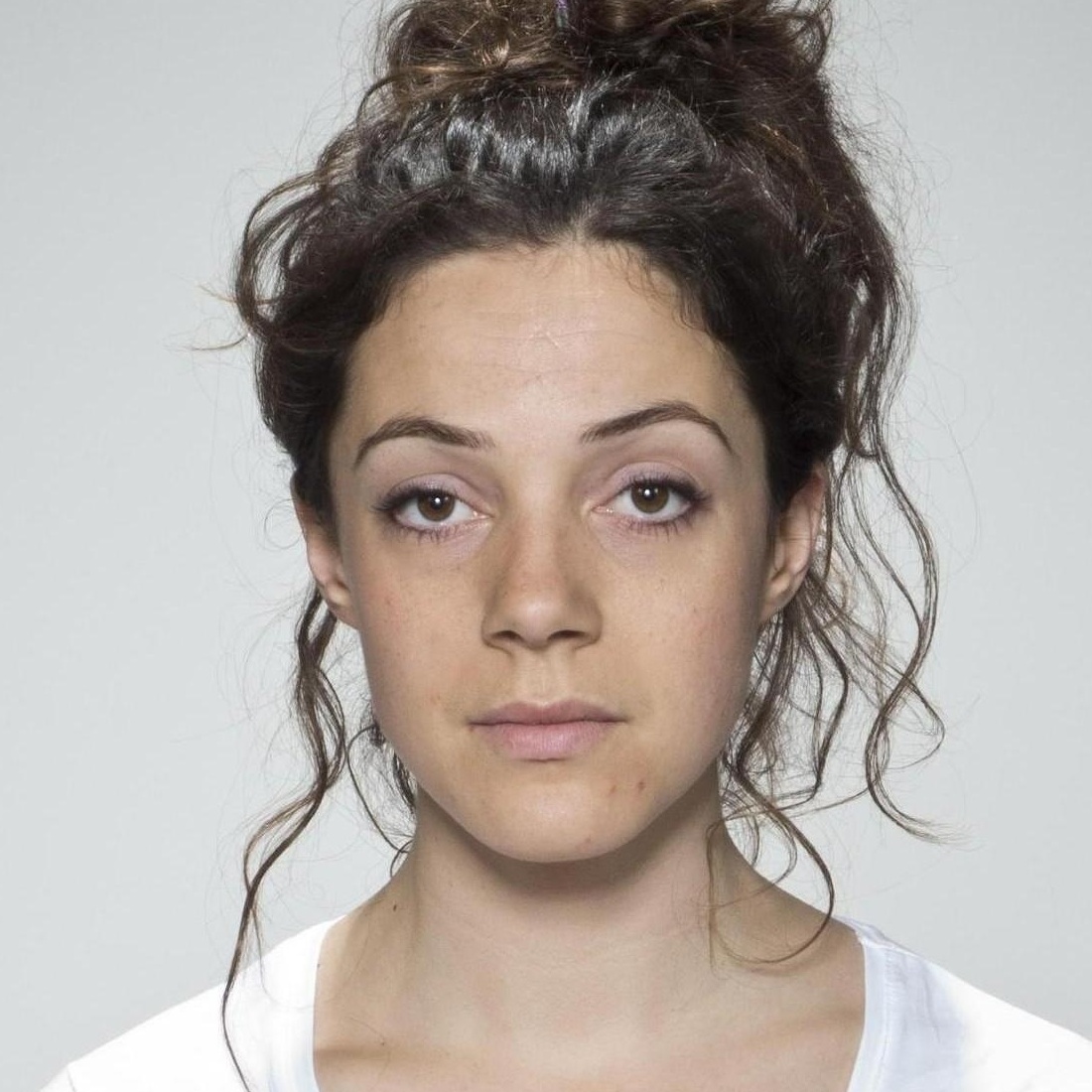}
  \includegraphics[width=0.22\columnwidth]{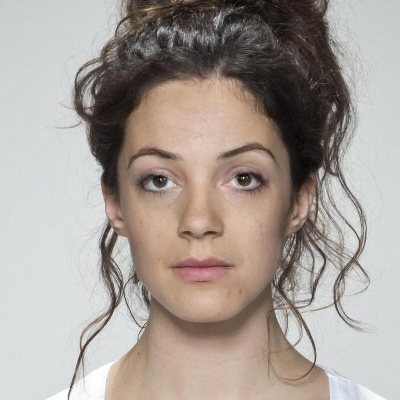}
  \includegraphics[width=0.22\columnwidth]{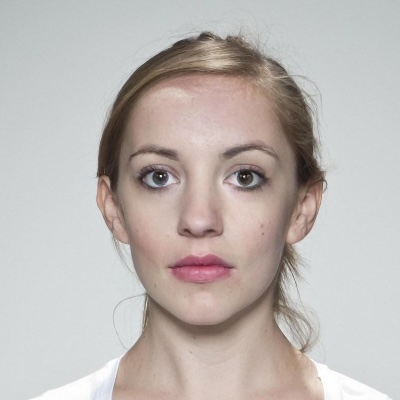}
  \includegraphics[width=0.22\columnwidth]{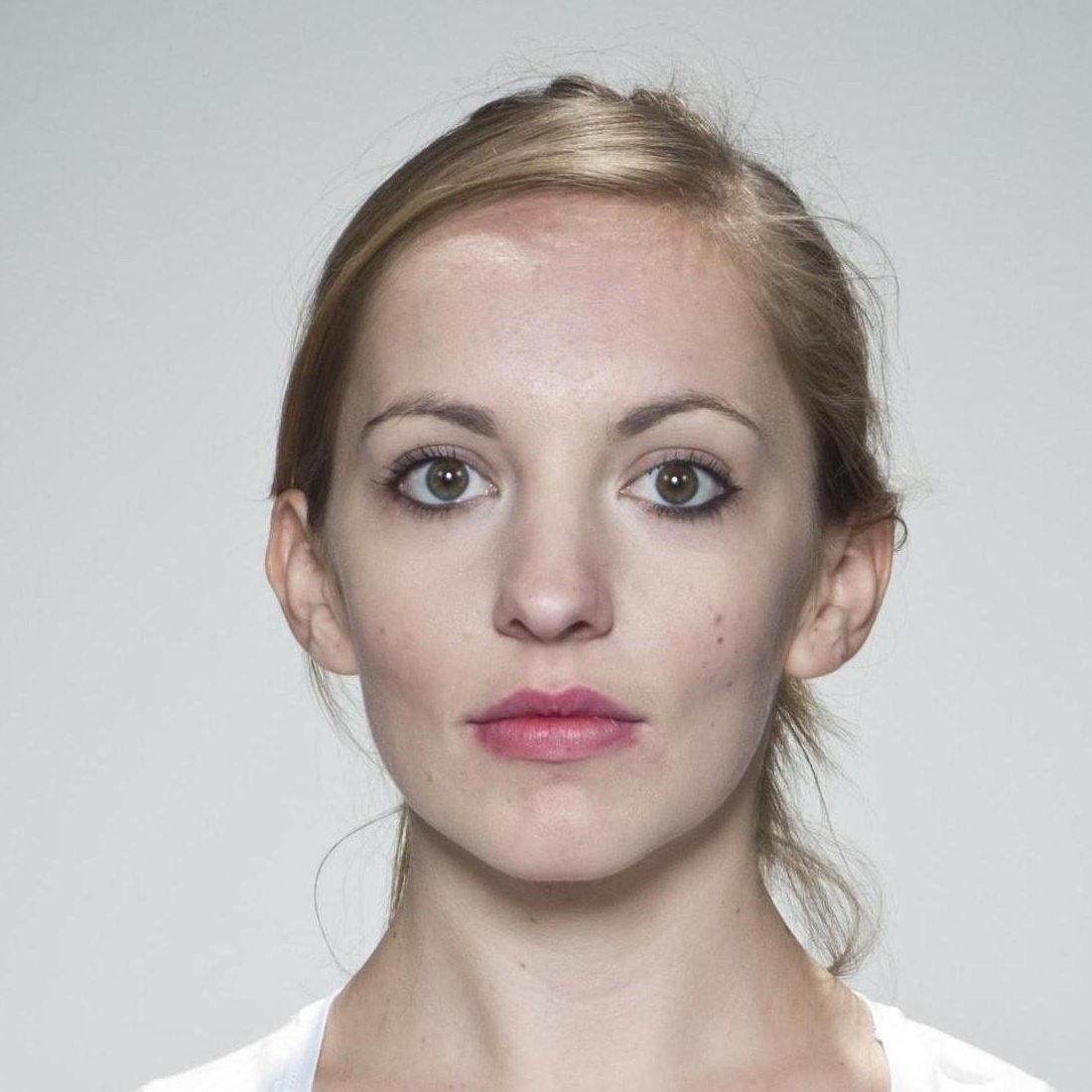} \\

  \includegraphics[width=0.22\columnwidth]{001_03.jpg}
  \includegraphics[width=0.22\columnwidth]{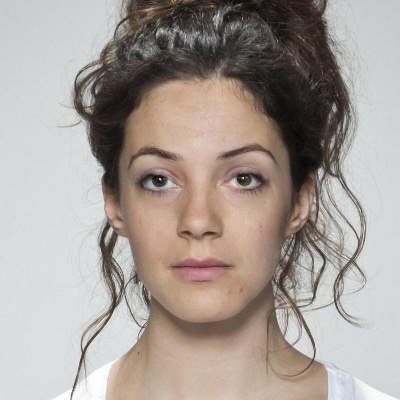}
  \includegraphics[width=0.22\columnwidth]{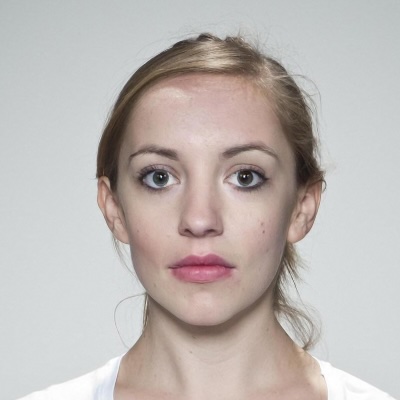}
  \includegraphics[width=0.22\columnwidth]{002_03.jpg} \\

  \includegraphics[width=0.22\columnwidth]{001_03.jpg}
  \includegraphics[width=0.22\columnwidth]{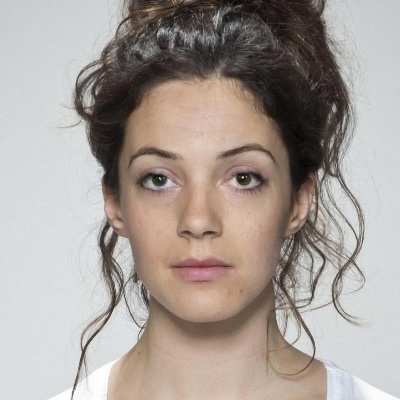}
  \includegraphics[width=0.22\columnwidth]{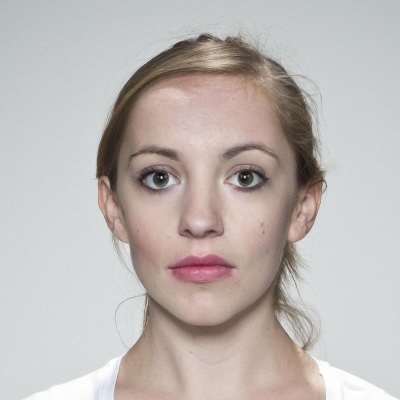}
  \includegraphics[width=0.22\columnwidth]{002_03.jpg}
  \vspace{-0.3cm}
  \caption{Samples of results drawn with varying network widths. The left and right columns show I$_0$ and I$_1$ respectively. The second column shows a morphing from I$_0$ to I$_1$ at $t=0.4$, while the third column shows morphing from I$_1$ to I$_0$ at $t=0.6$.}
  \label{fig:ablation-samples}
\end{figure}

For the neural warping, we vary the network width of its single hidden layer by 32 and 64 neurons. Fig~\ref{fig:ablation-samples} shows reconstructions using mixed cloning with varying values of $t$ and morphing directions  (either I$_0$ to I$_1$ or I$_1$ to~I$_0$).

Table~\ref{tab:fid-lpips-width} shows the FID and LPIPS scores for varying hidden-layer widths. The result quality improves dramatically as the network size increases as well. However, at 128 neurons, the results reach a good trade-off between quality and training speed. Thus, we chose this size for the experiments in the main paper.

\begin{table}[htb]
    \centering
    \caption{FID and LPIPS for our warping with linear blending using 32, 64 or 128 neurons for the hidden layer. The 128 neuron results were reproduced from Tab. 1 in the main paper.}
    \small{
    \begin{tabular}{lrrr}
        \toprule
        \textbf{Net. Width }& \textbf{FID} $\downarrow$ & \textbf{LPIPS} (I$_0$, I) $\downarrow$ & \textbf{LPIPS} (I, I$_1$) $\downarrow$ \\
        \midrule
        32  & 101.034 & 0.211 & 0.222 \\
        64  & 112.270 & 0.215 & 0.224 \\
        128 &  31.950 & 0.158 & 0.164 \\
        \bottomrule
    \end{tabular}}
    \label{tab:fid-lpips-width}
\end{table}

\section{Additional Results}
We show additional variations of ethnicities and genders to illustrate the flexibility of our approach (Sec.~\ref{sec:vari-gend-ethn}). Furthermore, we show experiments using automatic face landmark detection via DLib (Sec.~\ref{sec:empl-autom-landm}) and how the morphings may be improved with additional manual adjustments to these landmarks (Sec.~\ref{sec:manu-doct-landm}).

\subsection{Variations of Gender and Ethnicity}
\label{sec:vari-gend-ethn}
\begin{figure}[H]
  \centering
  \includegraphics[width=0.22\columnwidth]{001_03.jpg}
  \includegraphics[width=0.22\columnwidth]{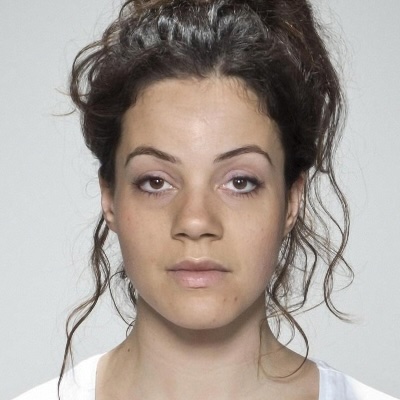}
  \includegraphics[width=0.22\columnwidth]{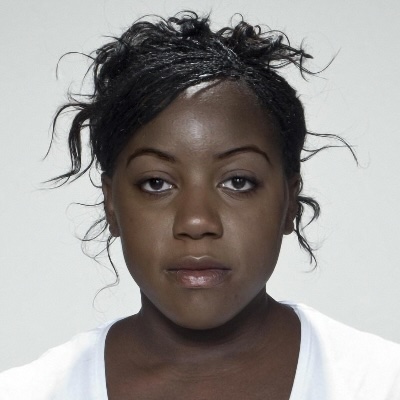}
  \includegraphics[width=0.22\columnwidth]{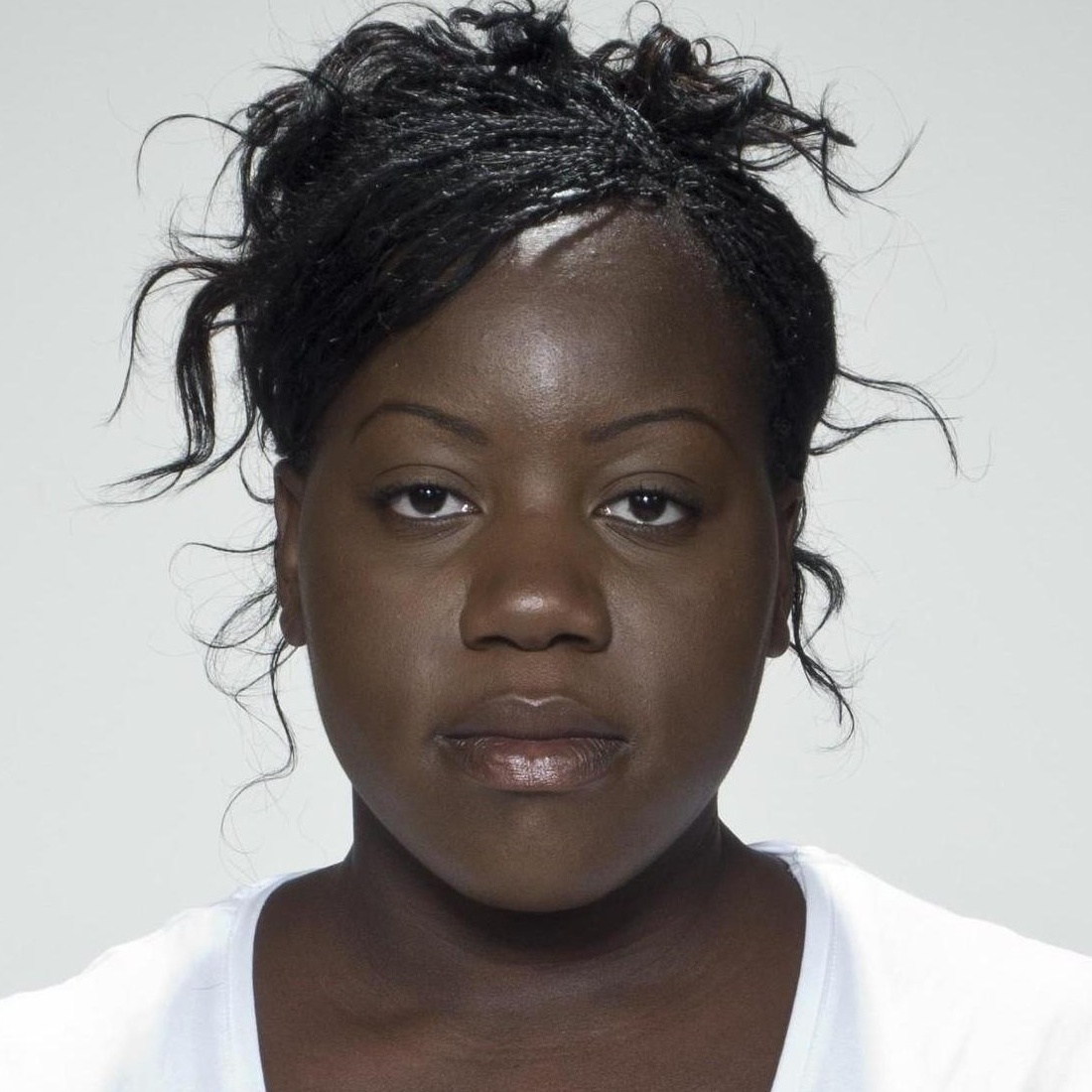} \\

  \includegraphics[width=0.22\columnwidth]{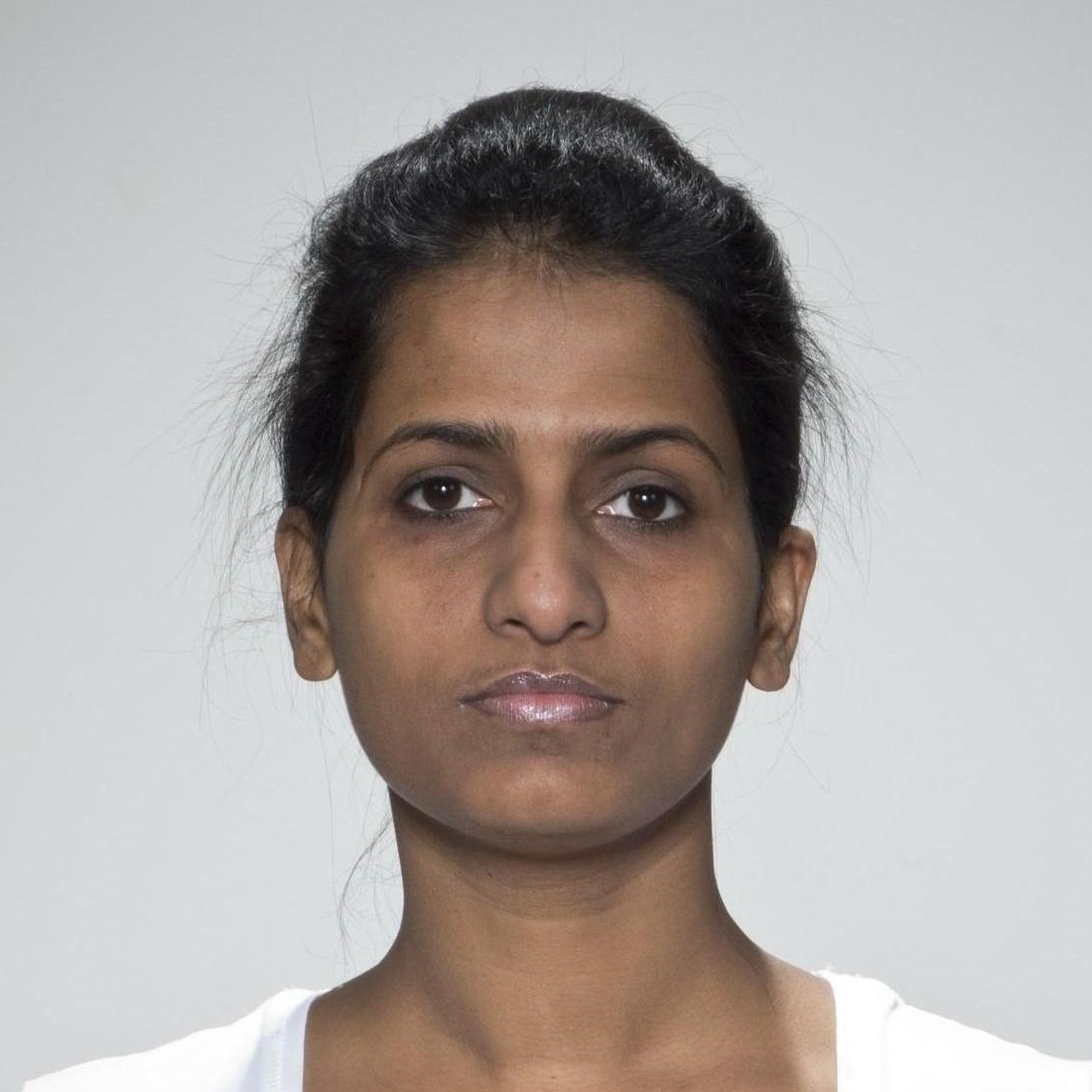}
  \includegraphics[width=0.22\columnwidth]{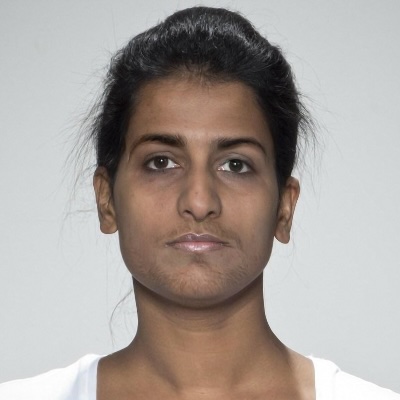}
  \includegraphics[width=0.22\columnwidth]{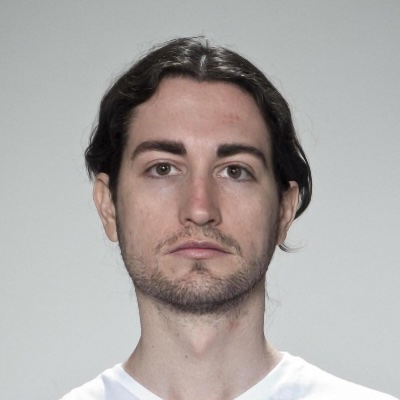}
  \includegraphics[width=0.22\columnwidth]{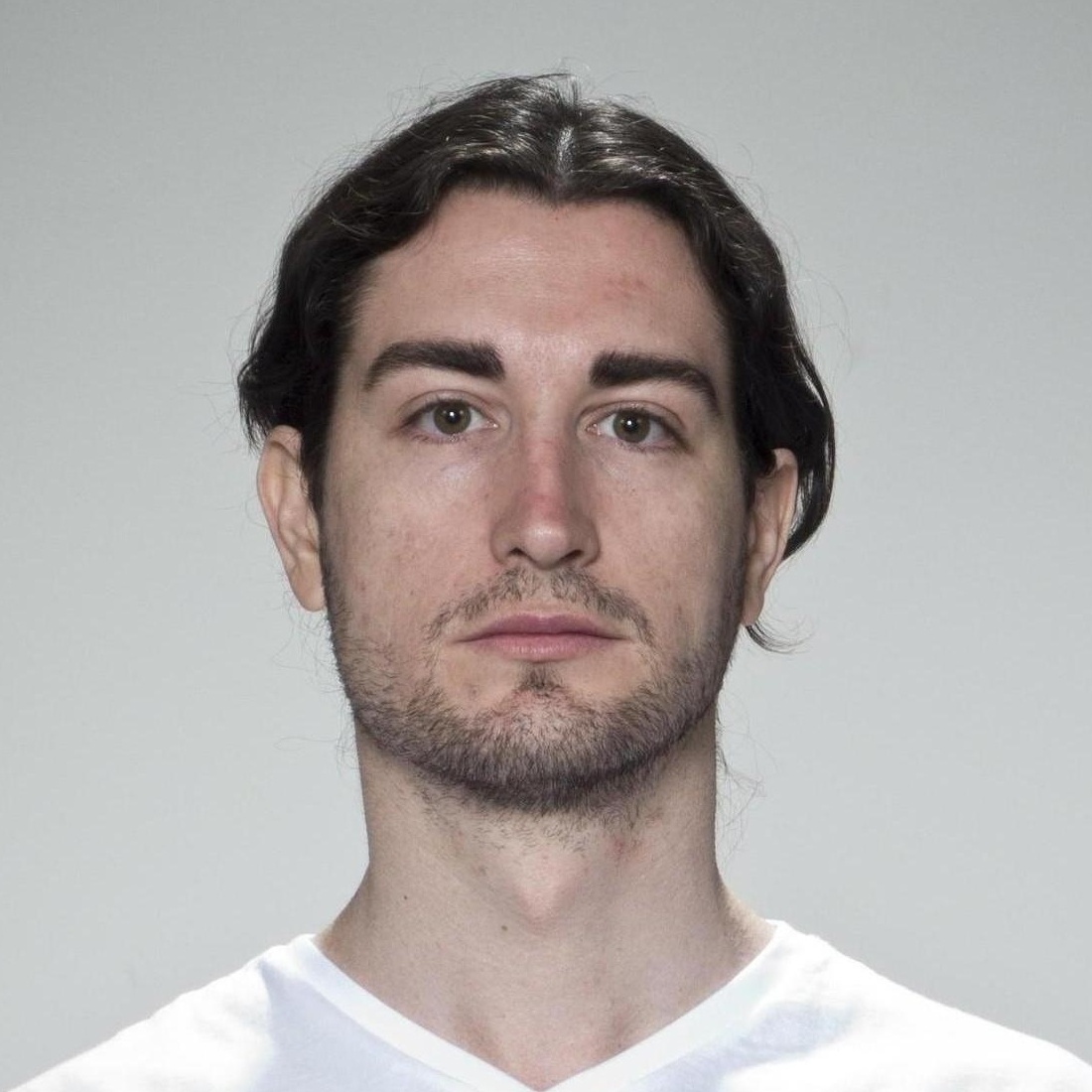} \\

  \includegraphics[width=0.22\columnwidth]{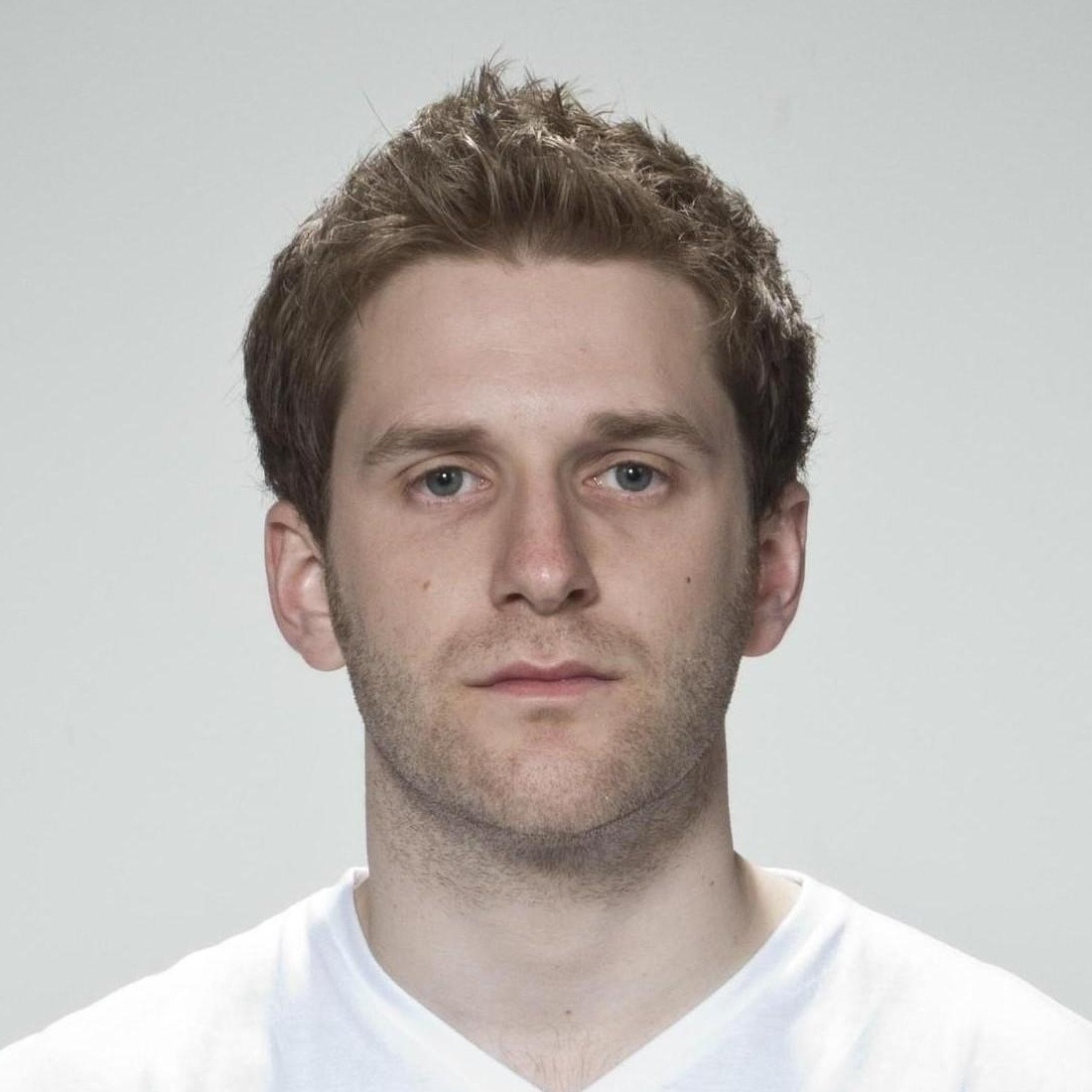}
  \includegraphics[width=0.22\columnwidth]{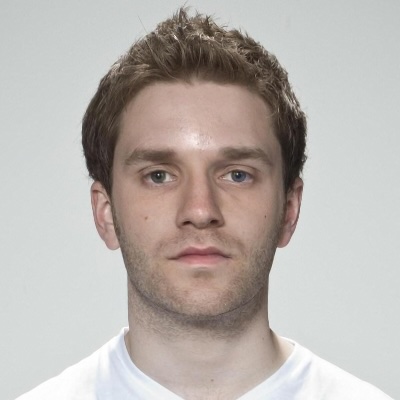}
  \includegraphics[width=0.22\columnwidth]{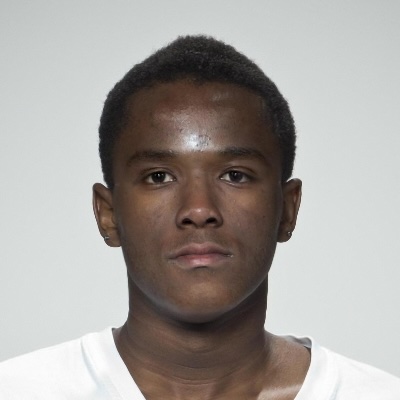}
  \includegraphics[width=0.22\columnwidth]{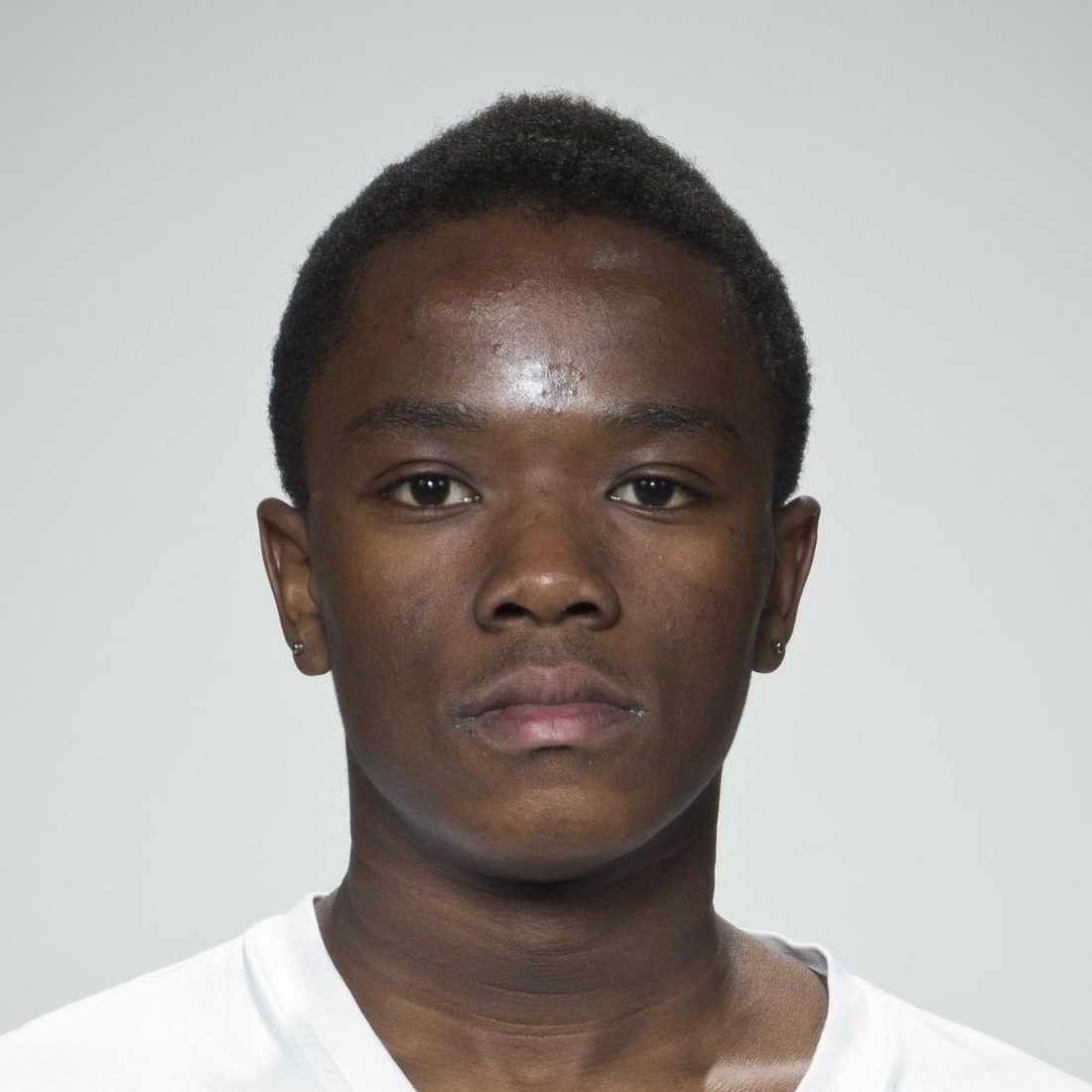} \\

  \includegraphics[width=0.22\columnwidth]{001_03.jpg}
  \includegraphics[width=0.22\columnwidth]{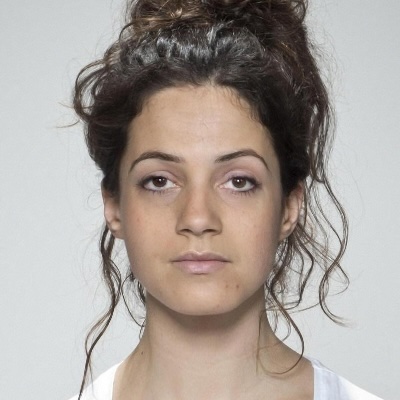}
  \includegraphics[width=0.22\columnwidth]{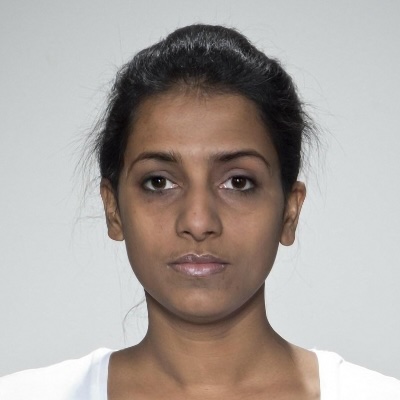}
  \includegraphics[width=0.22\columnwidth]{006_03.jpg} \\

  \includegraphics[width=0.22\columnwidth]{004_03.jpg}
  \includegraphics[width=0.22\columnwidth]{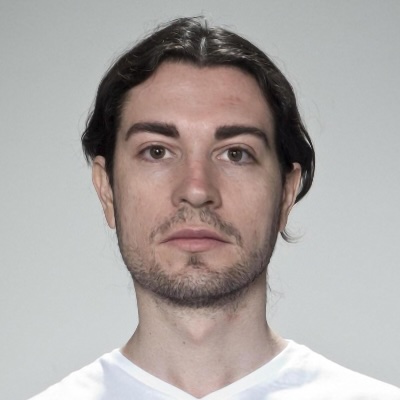}
  \includegraphics[width=0.22\columnwidth]{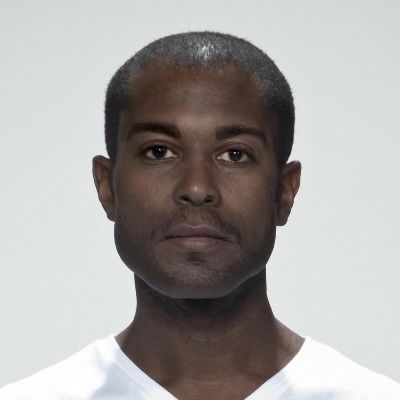}
  \includegraphics[width=0.22\columnwidth]{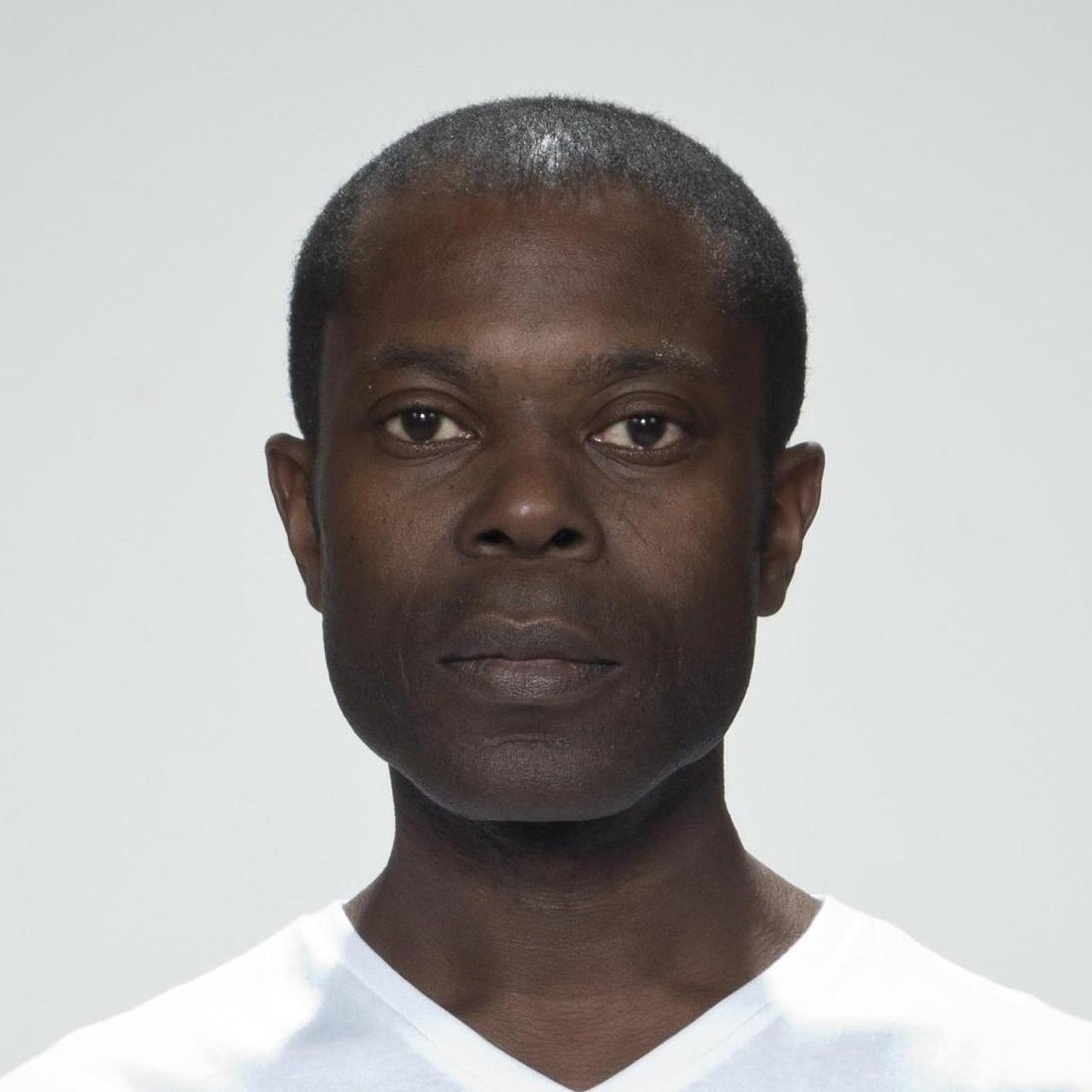} \\

  \includegraphics[width=0.22\columnwidth]{062_03.jpg}
  \includegraphics[width=0.22\columnwidth]{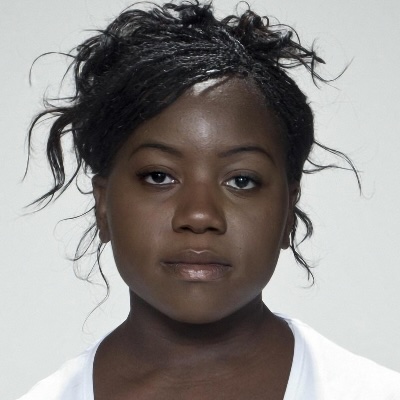}
  \includegraphics[width=0.22\columnwidth]{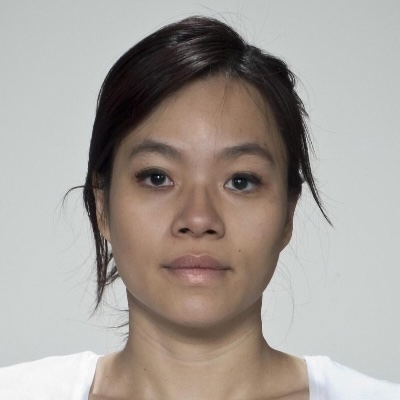}
  \includegraphics[width=0.22\columnwidth]{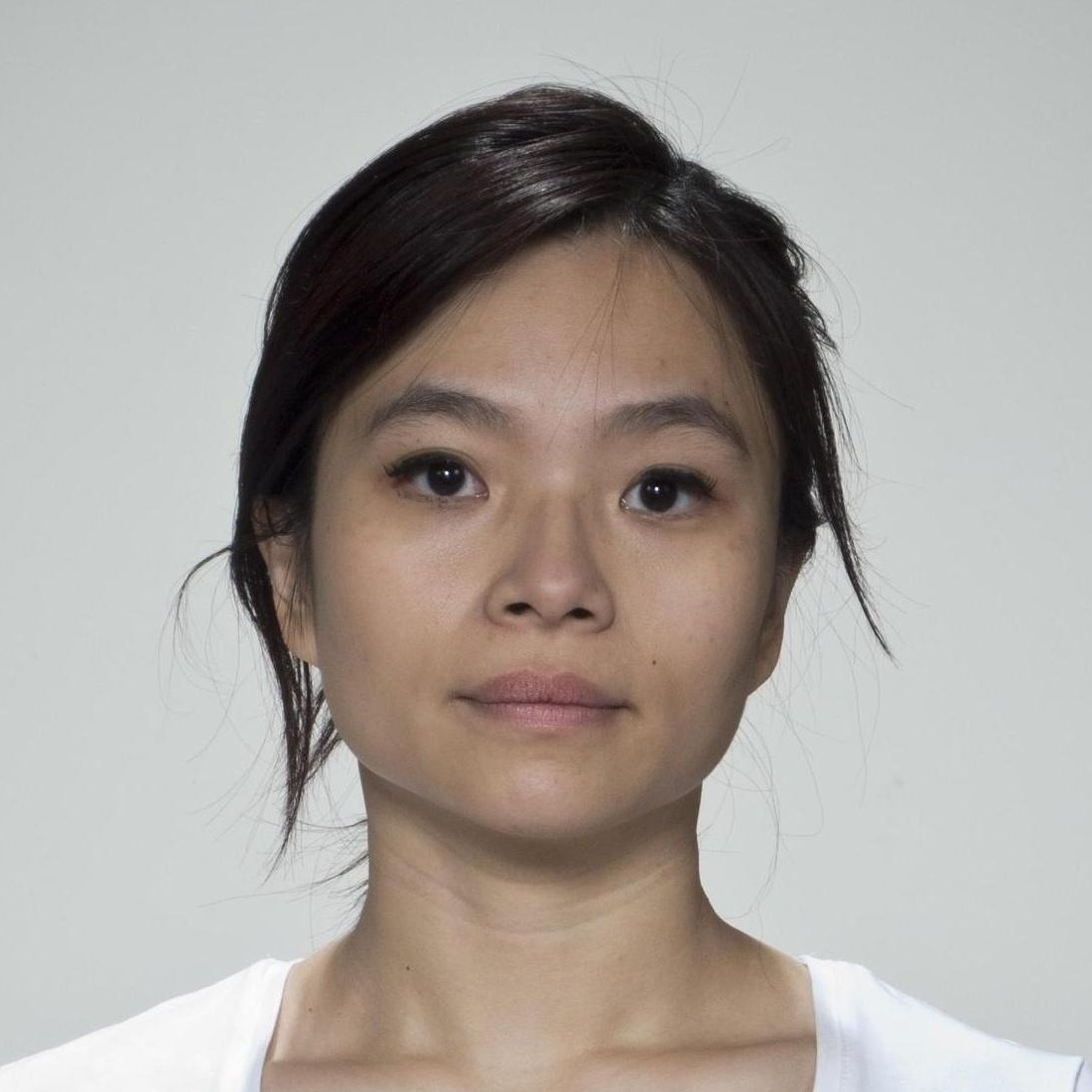} \\

  \includegraphics[width=0.22\columnwidth]{001_03.jpg}
  \includegraphics[width=0.22\columnwidth]{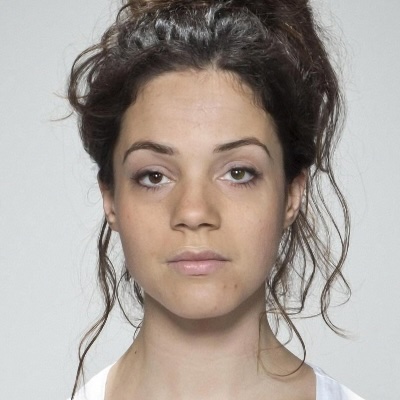}
  \includegraphics[width=0.22\columnwidth]{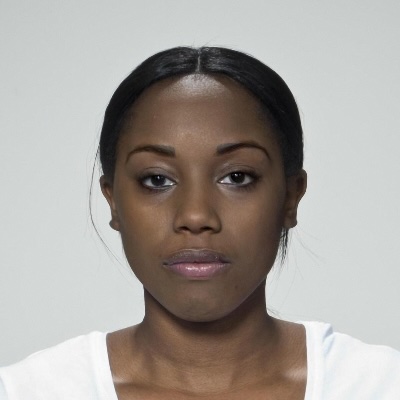}
  \includegraphics[width=0.22\columnwidth]{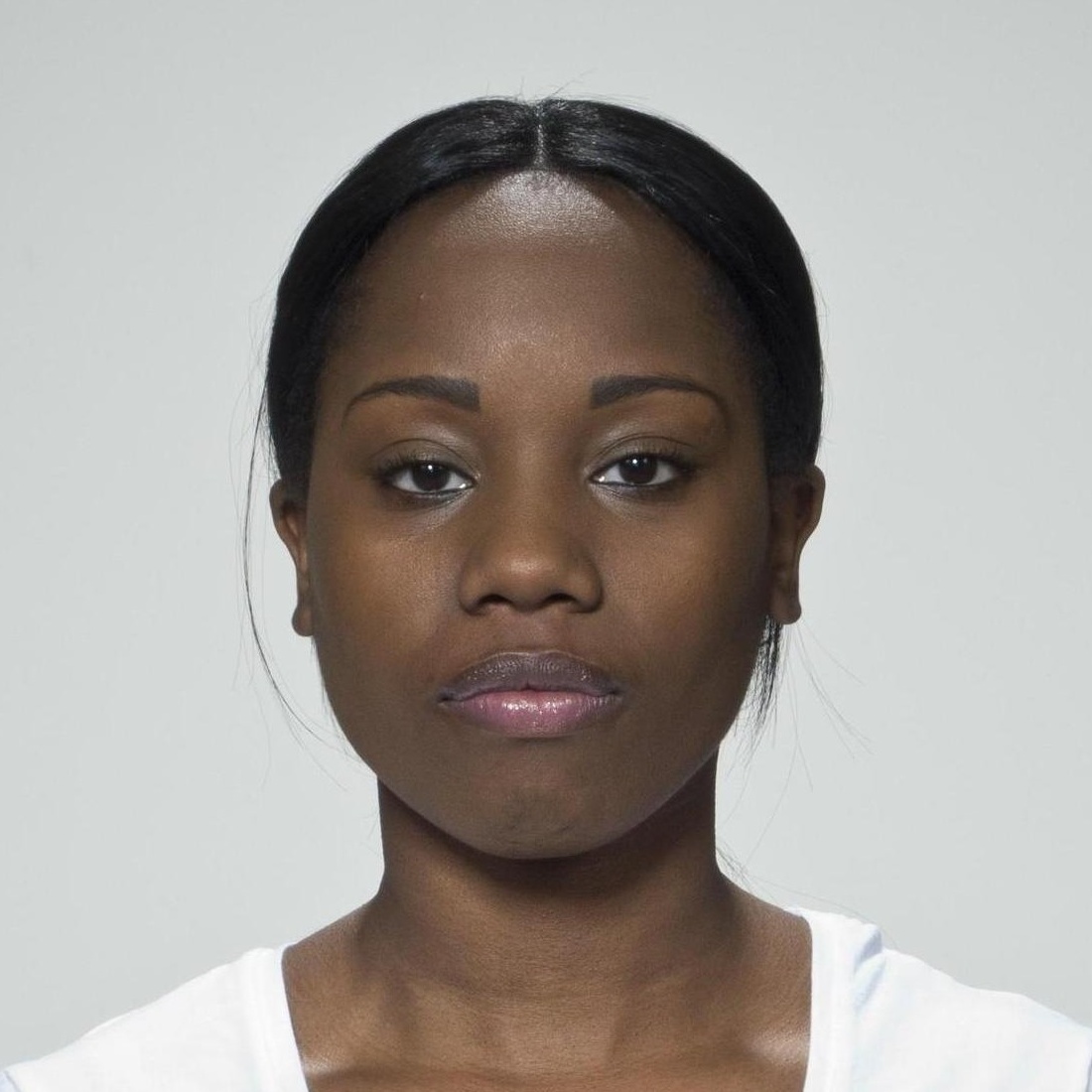}
 \vspace{-0.3cm}
  \caption{Morphings between people of varying ethnicities and genders. The left column shows the source image (I$_0$), followed by the morphing of I$_0$ to I$_1$ at $t=0.6$, morphing of I$_1$ to I$_0$ at $t=0.4$, and the target image (I$_1$).}
  \label{fig:eth-gender}
\end{figure}

One of the main limitations of the current state-of-the-art methods is generating credible morphings of people of different genders and ethnicities. Fig~\ref{fig:eth-gender} shows results of these morphings using our approach. We employed the landmarks provided with the FRLL dataset for the neural warping.

\subsection{Employing Automatic Landmark Detection}
\label{sec:empl-autom-landm}
\begin{figure}[htb]
  \centering
  \includegraphics[width=0.32\columnwidth]{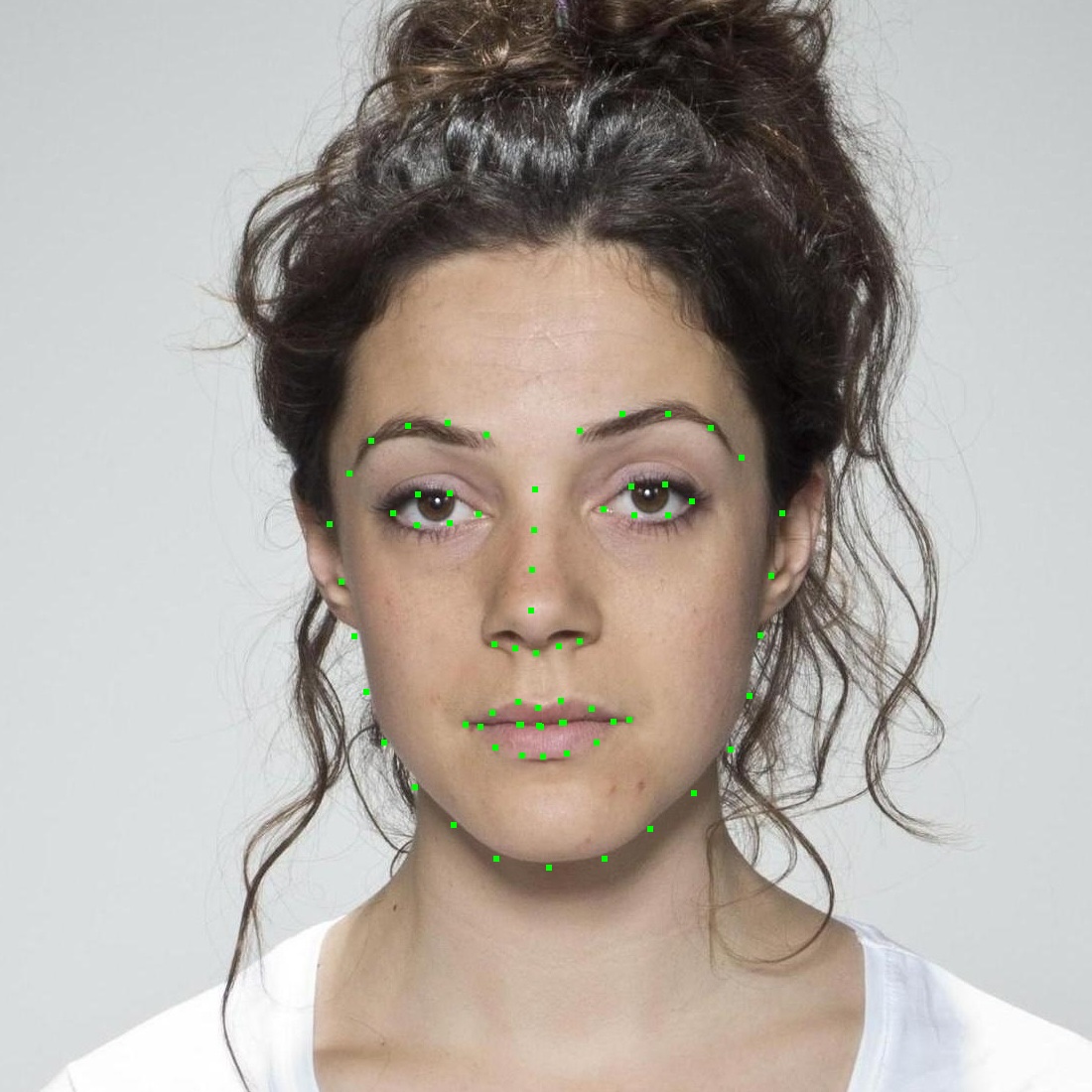}
  \includegraphics[width=0.32\columnwidth]{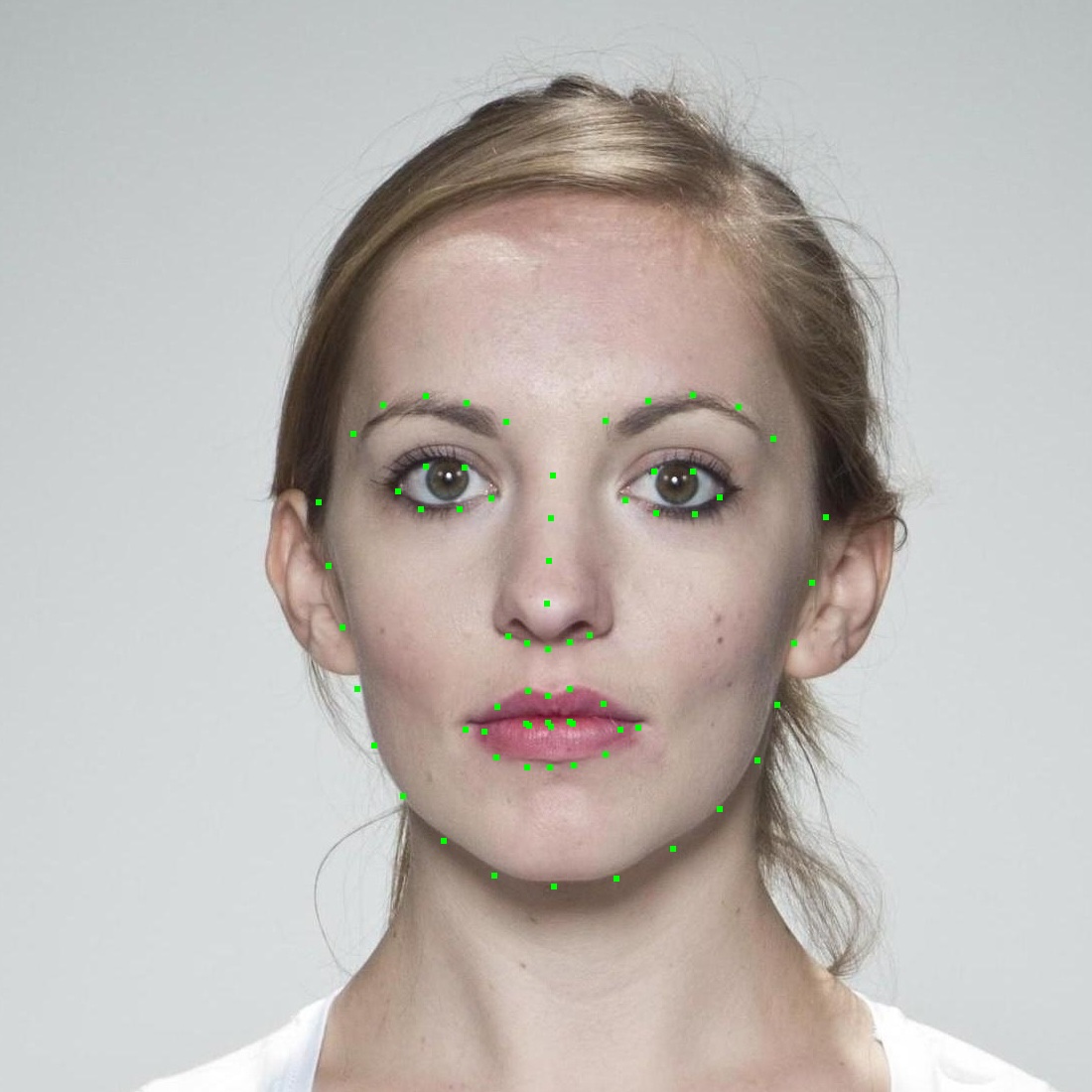}
  \includegraphics[width=0.32\columnwidth]{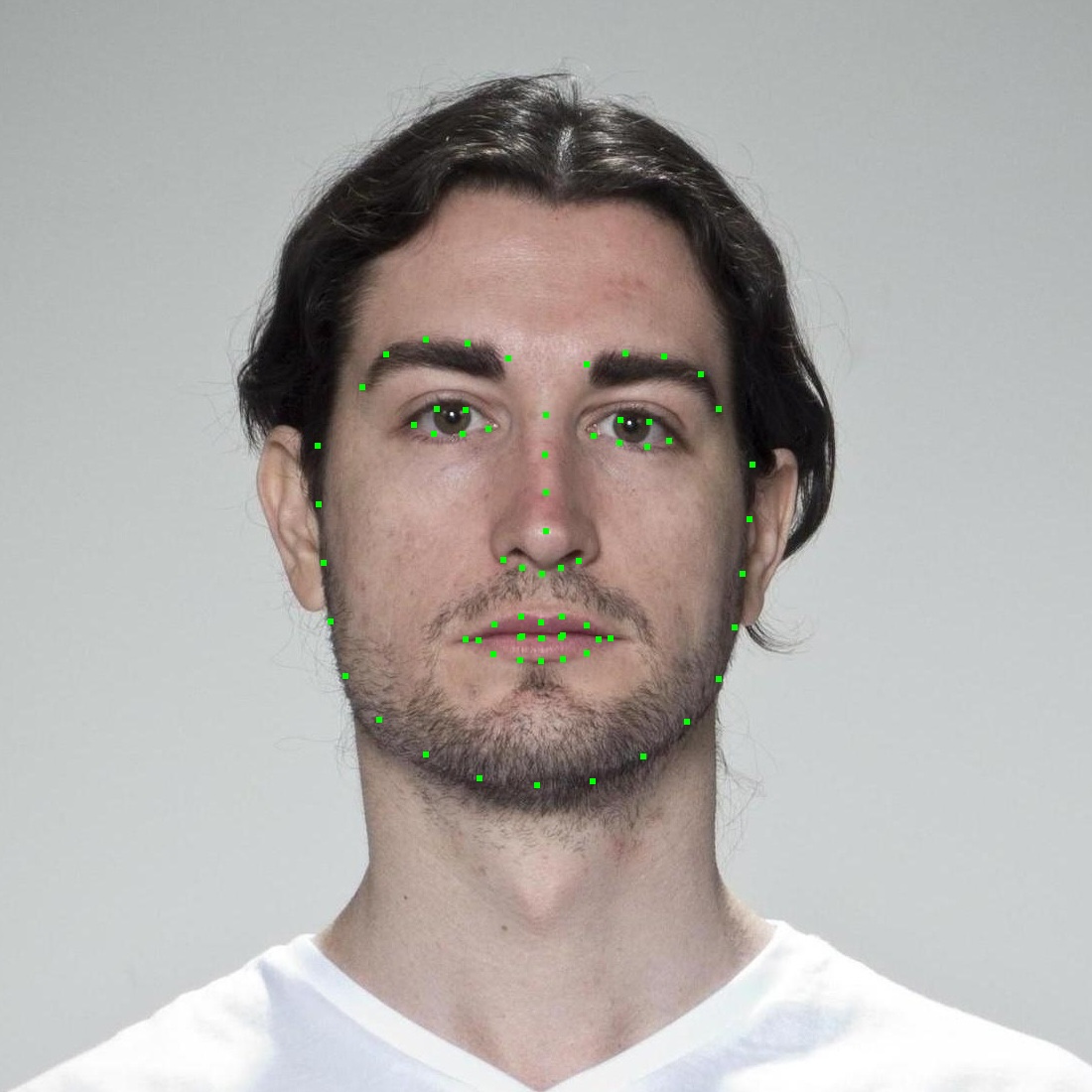}
  \includegraphics[width=0.32\columnwidth]{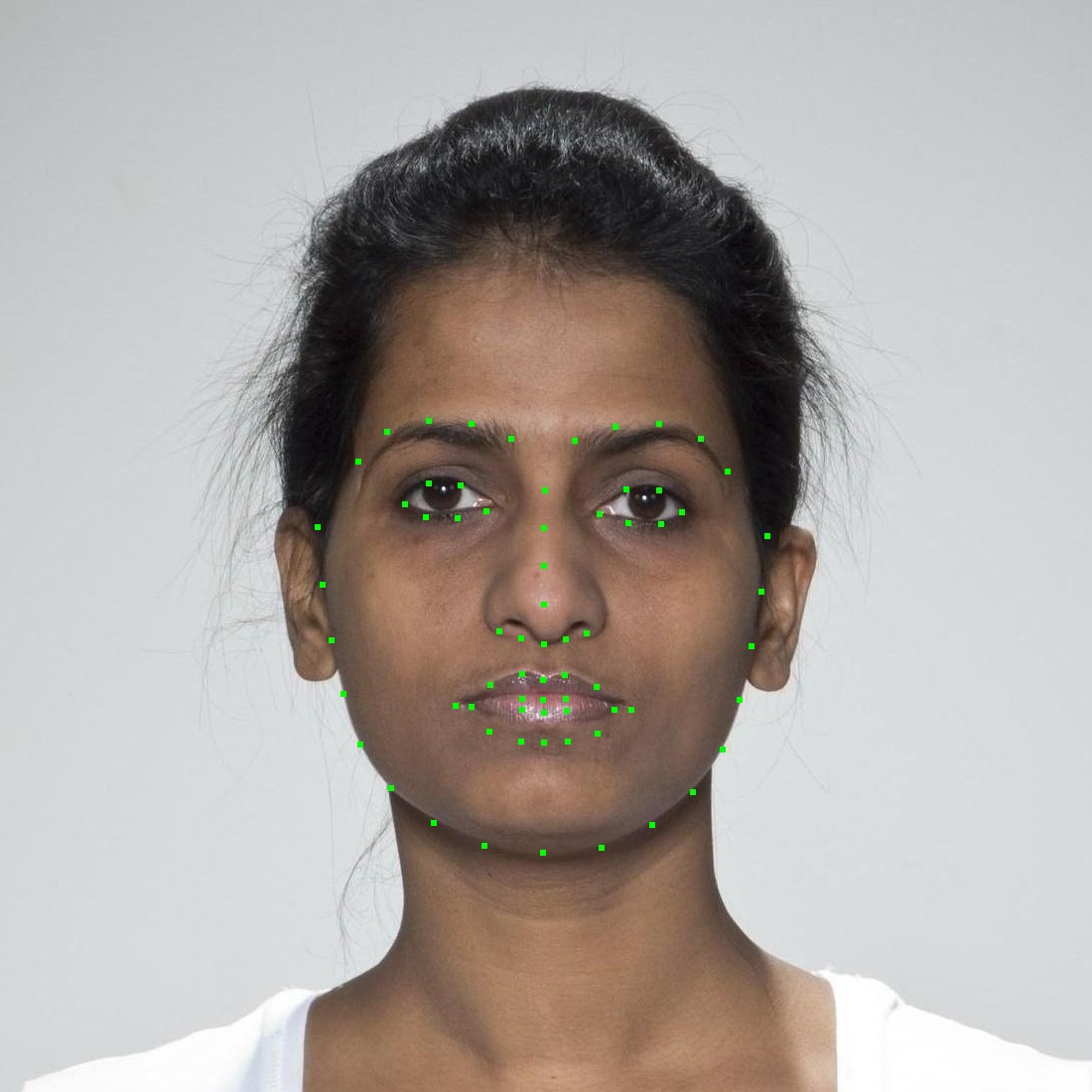}
  \includegraphics[width=0.32\columnwidth]{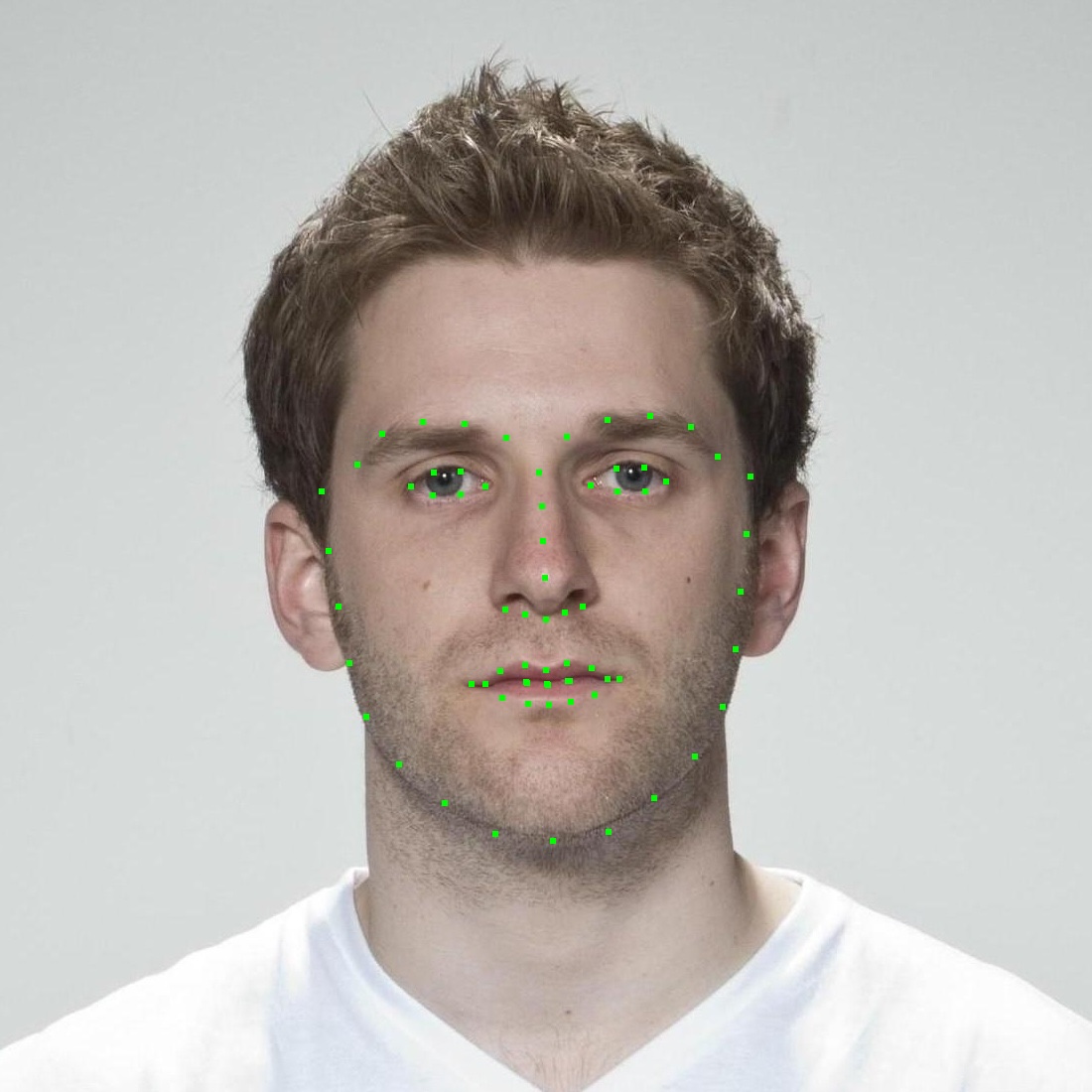}
  \includegraphics[width=0.32\columnwidth]{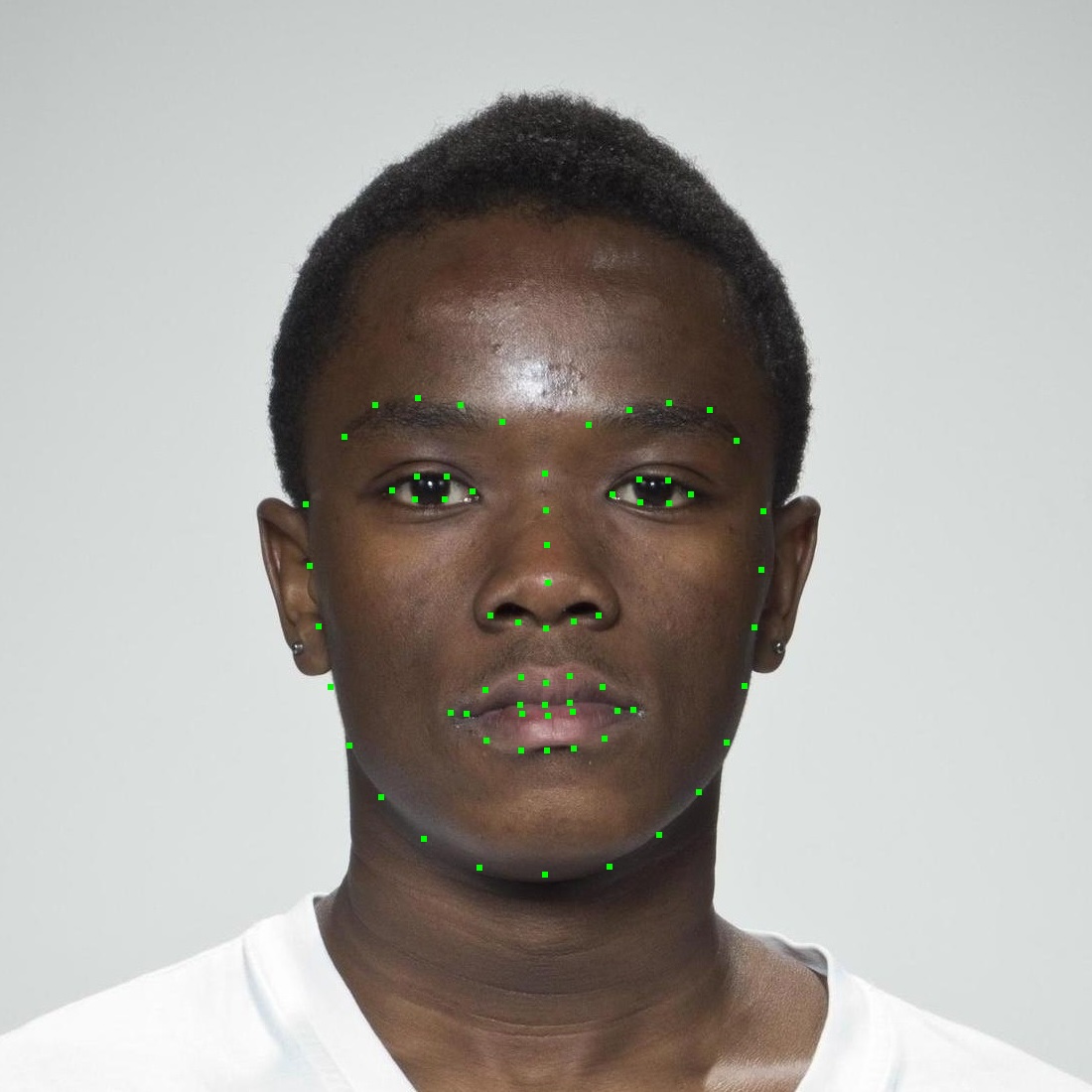}
  \includegraphics[width=0.32\columnwidth]{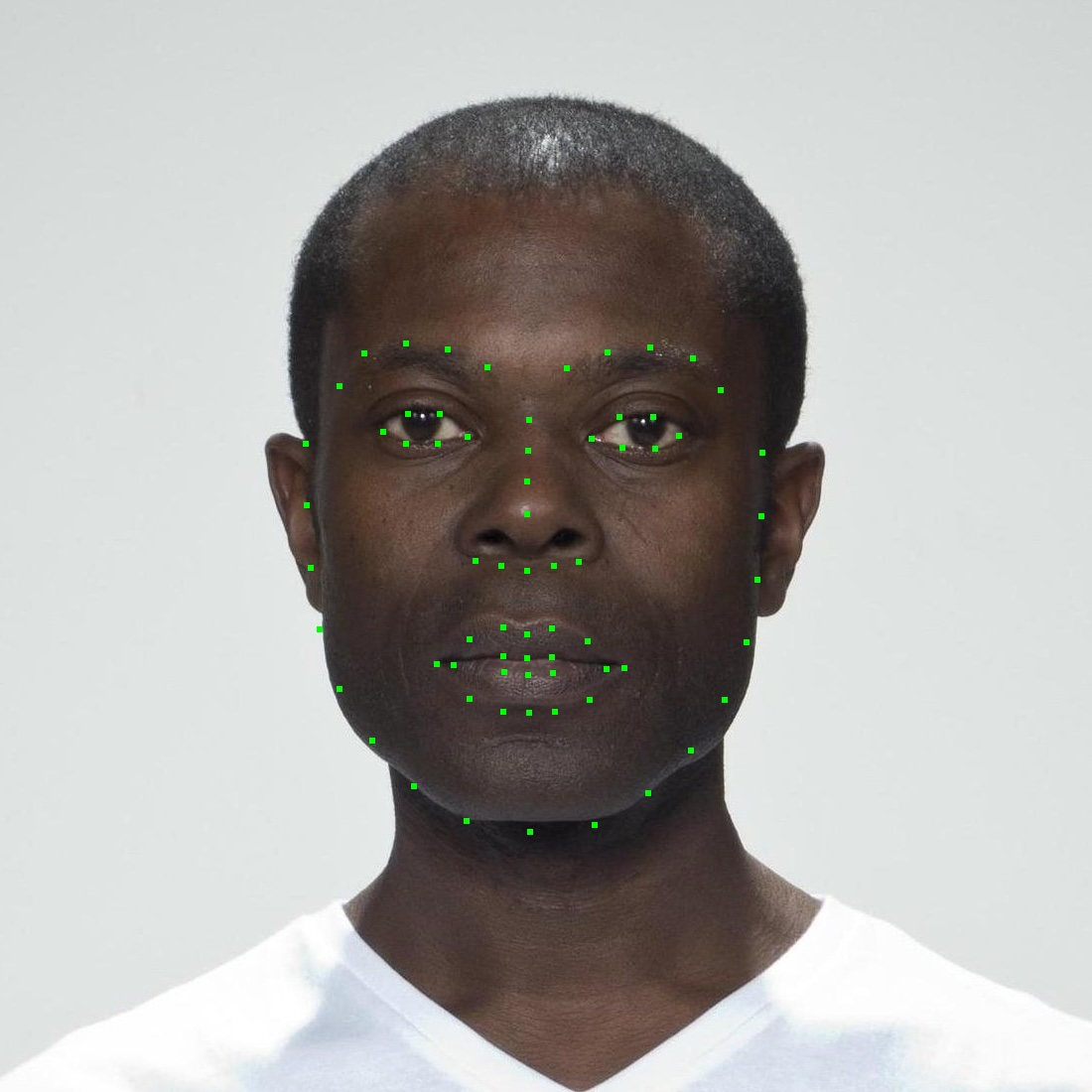}
  \includegraphics[width=0.32\columnwidth]{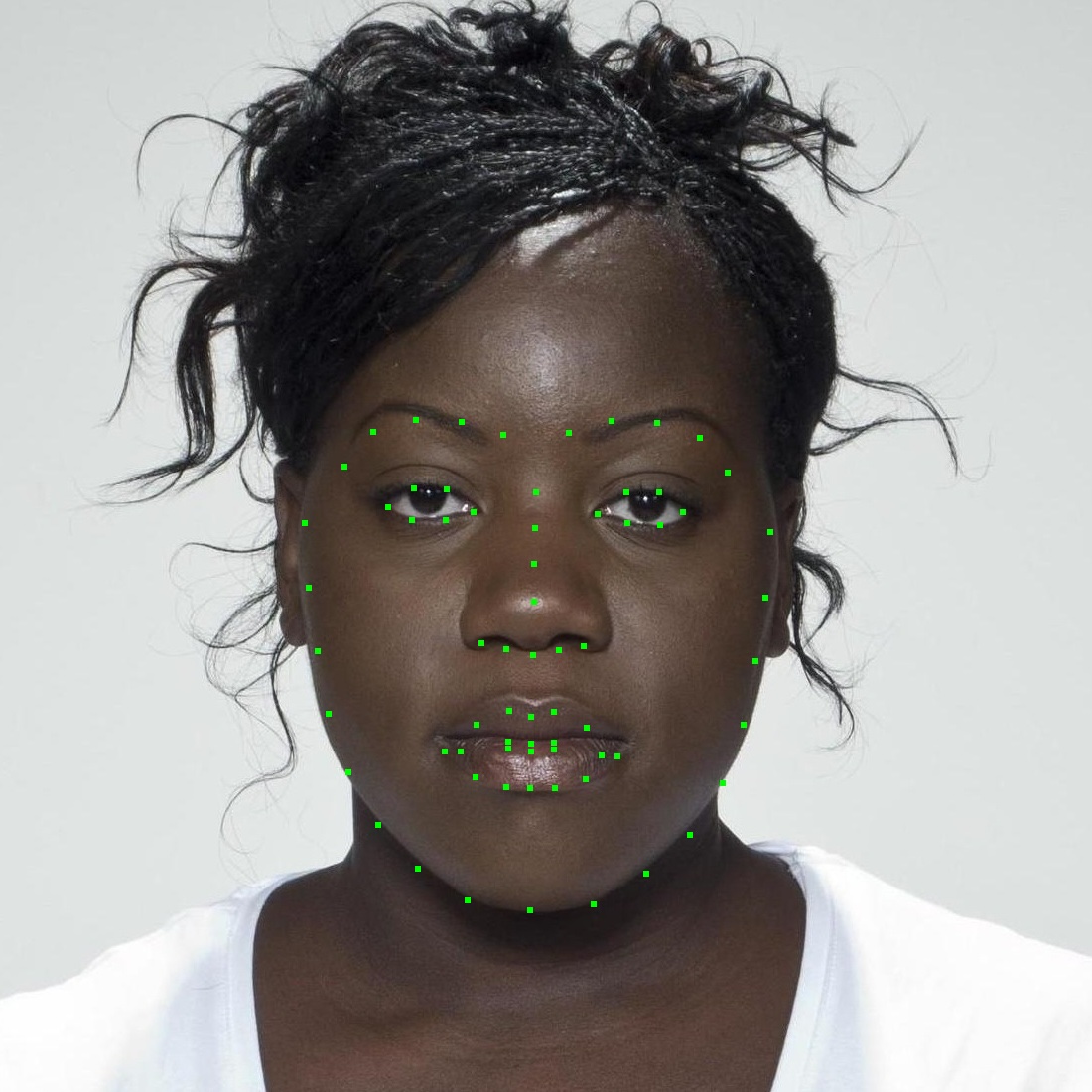}
  \includegraphics[width=0.32\columnwidth]{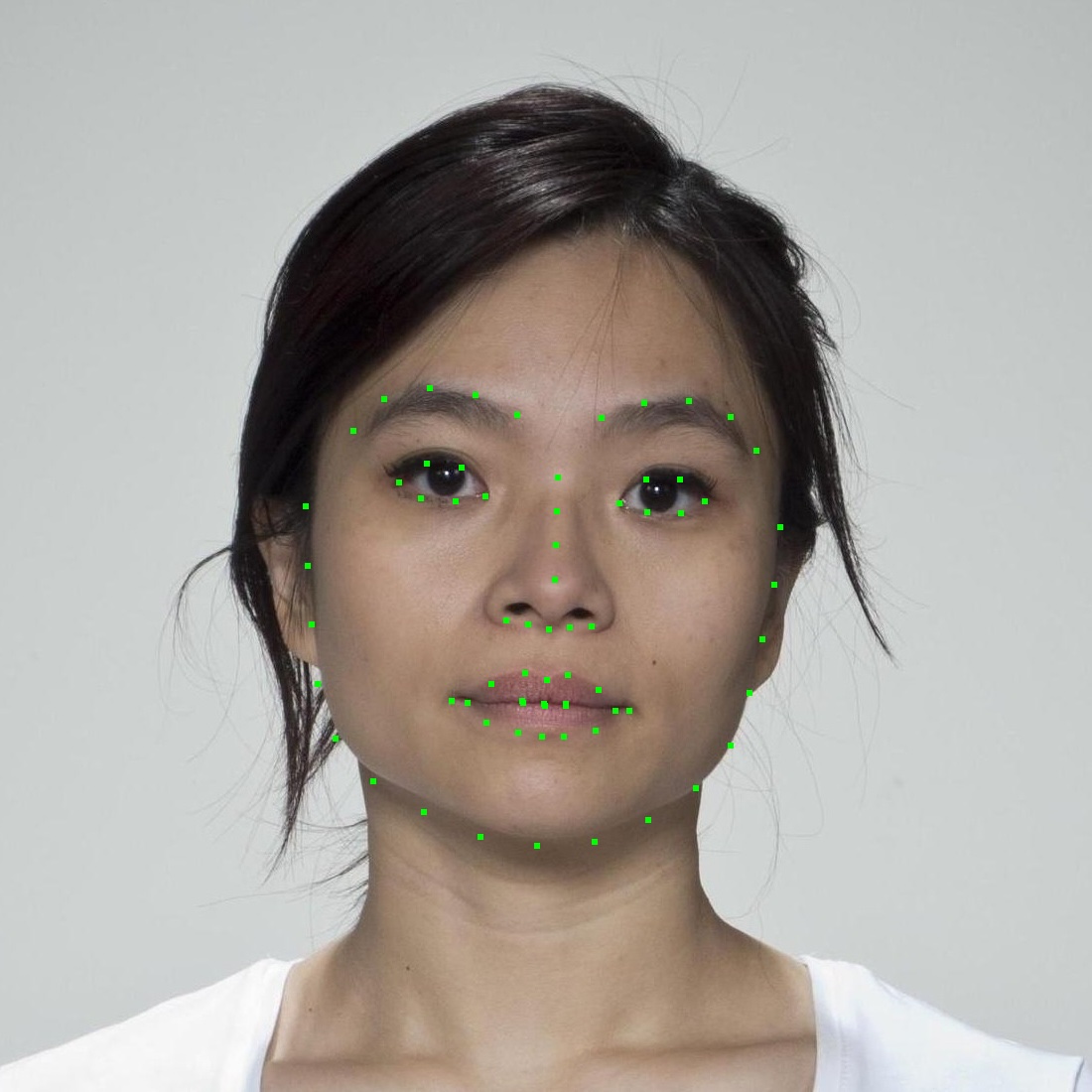}
  \includegraphics[width=0.32\columnwidth]{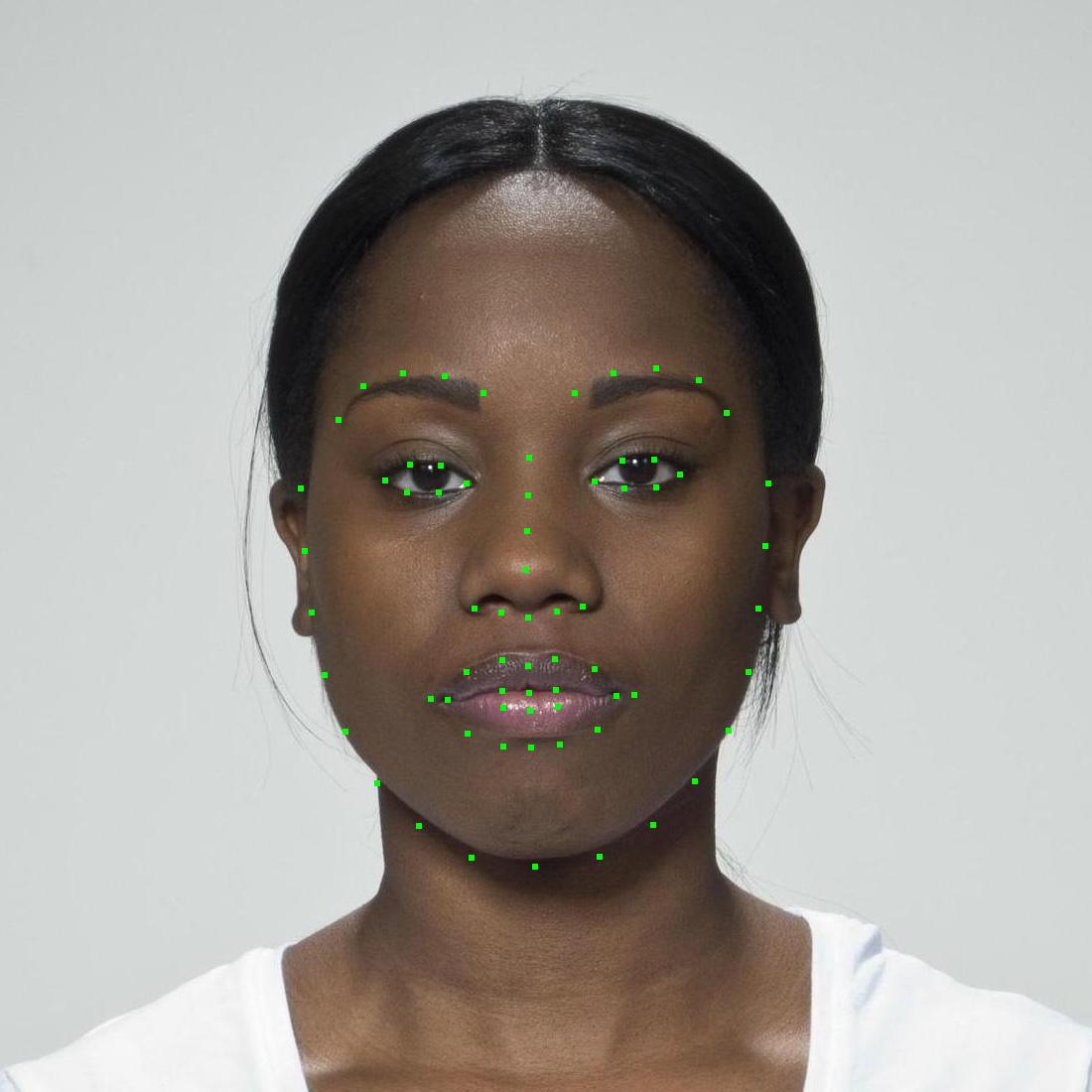}
  \vspace{-0.3cm}
  \caption{Landmarks found by DLib 68 landmarks overlaid on sample faces of the FRLL dataset.}
  \label{fig:landmarks-dlib}
\end{figure}

\begin{figure}[htb]
  \centering
  \includegraphics[width=0.22\columnwidth]{001_03.jpg}
  \includegraphics[width=0.22\columnwidth]{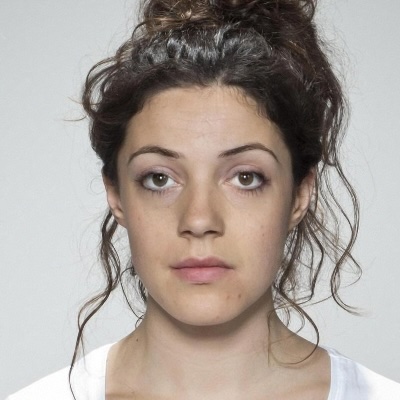}
  \includegraphics[width=0.22\columnwidth]{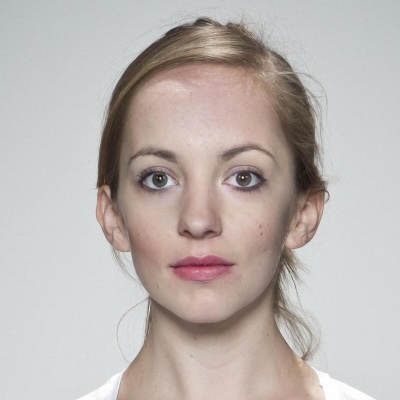}
  \includegraphics[width=0.22\columnwidth]{002_03.jpg} \\

  \includegraphics[width=0.22\columnwidth]{001_03.jpg}
  \includegraphics[width=0.22\columnwidth]{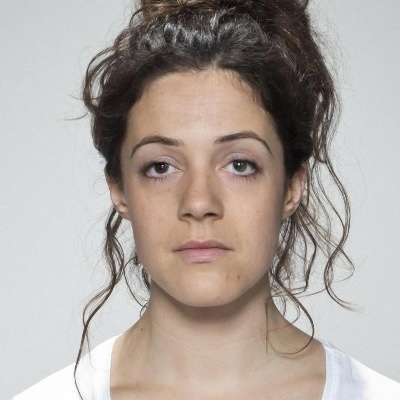}
  \includegraphics[width=0.22\columnwidth]{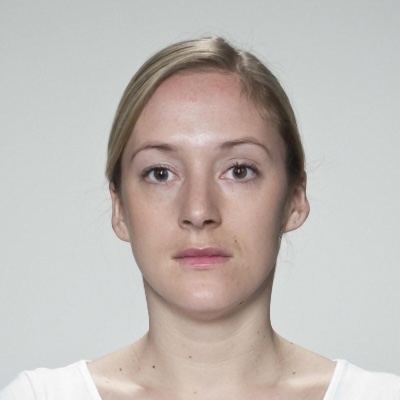}
  \includegraphics[width=0.22\columnwidth]{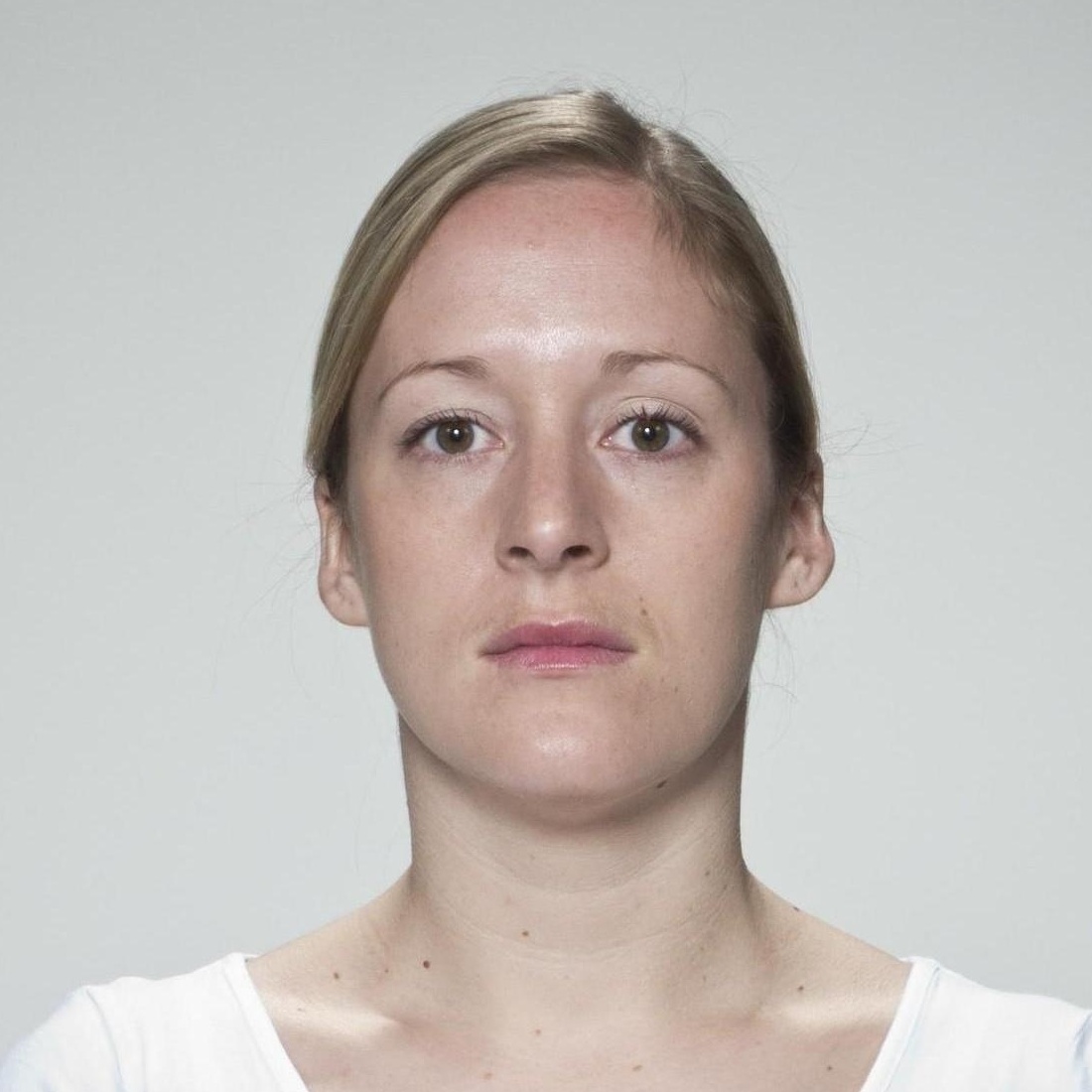} \\

  \includegraphics[width=0.22\columnwidth]{004_03.jpg}
  \includegraphics[width=0.22\columnwidth]{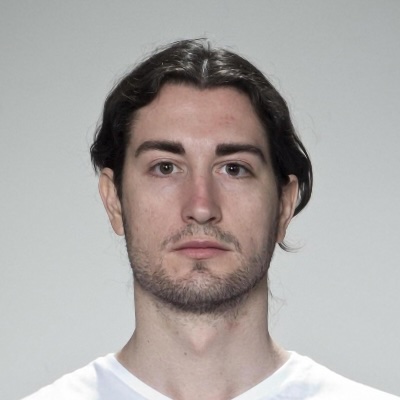}
  \includegraphics[width=0.22\columnwidth]{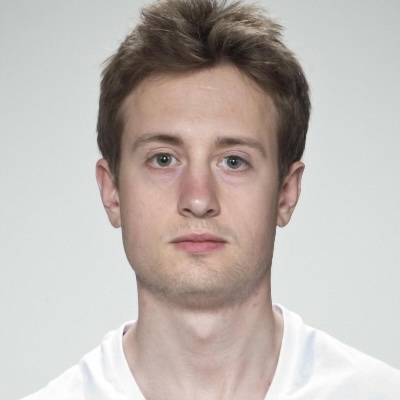}
  \includegraphics[width=0.22\columnwidth]{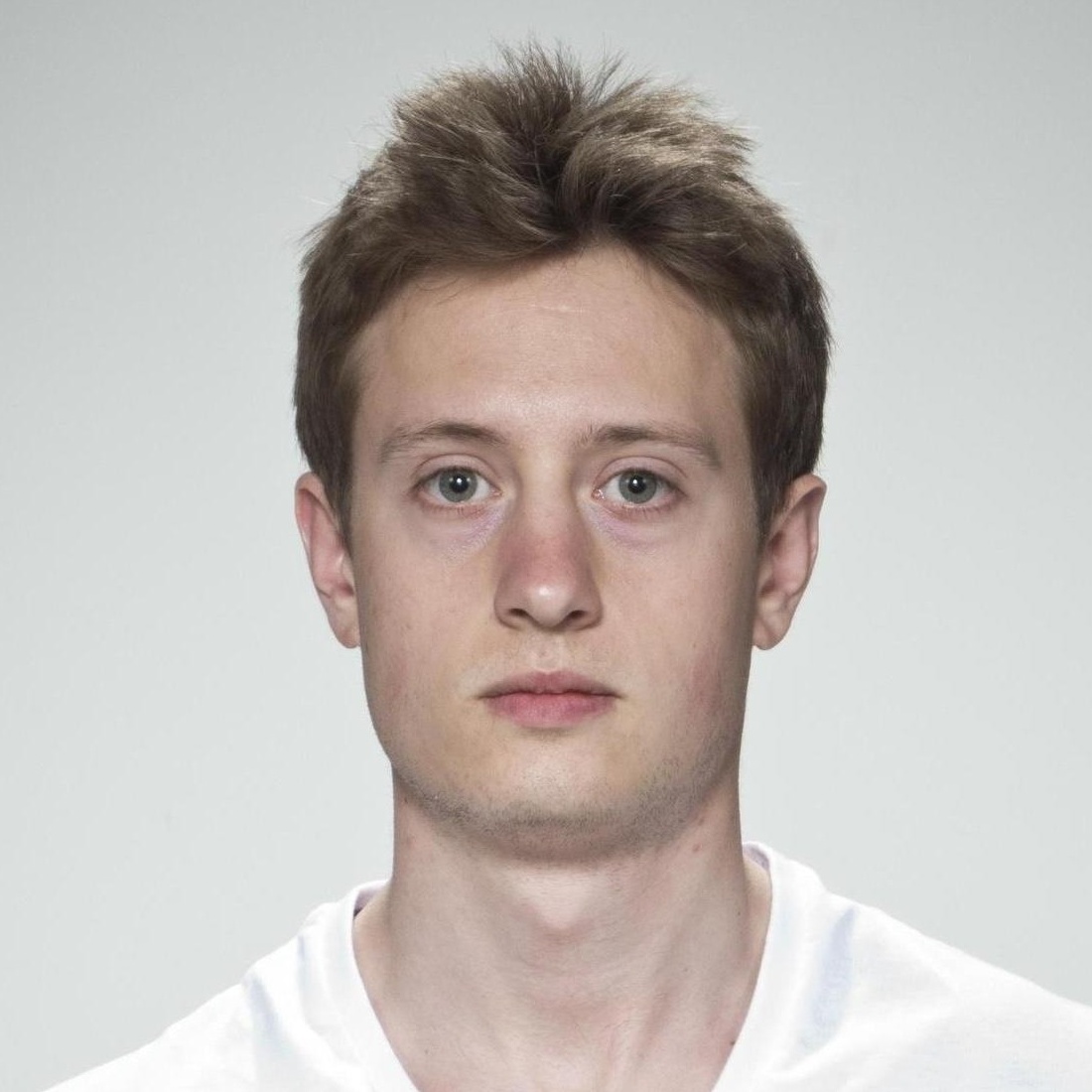} \\

  \includegraphics[width=0.22\columnwidth]{004_03.jpg}
  \includegraphics[width=0.22\columnwidth]{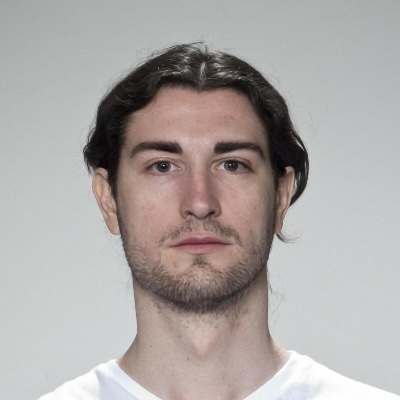}
  \includegraphics[width=0.22\columnwidth]{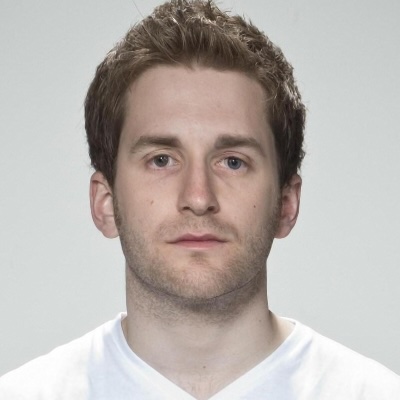}
  \includegraphics[width=0.22\columnwidth]{029_03.jpg}
\vspace{-0.3cm}
  \caption{Morphings created using our neural warping approach with landmarks obtained with DLib 68 landmarks model. From left to right: Source image (I$_0$), followed by the morphing of I$_0$ to I$_1$ at $t=0.6$, morphing of I$_1$ to I$_0$ at $t=0.4$, and target image (I$_1$).}
  \label{fig:dlib-results}
\end{figure}

The landmarks used for the experiments were provided with the dataset, however, other approaches for facial landmark detection may be employed and attain equally good results. For the experiments below, we used DLib with the 68 landmark model~\cite{kazemi2014cvpr,sagonas2016}. Fig~\ref{fig:landmarks-dlib} shows the landmarks overlaid on sample images of the FRLL dataset, while Fig~\ref{fig:dlib-results} shows the results of morphings using our method with said landmarks.

\subsubsection{Manual Doctoring of Landmarks}
\label{sec:manu-doct-landm}
\begin{figure}[ht]
  \centering
  \includegraphics[width=0.24\columnwidth]{001_03.jpg}
  \includegraphics[width=0.24\columnwidth]{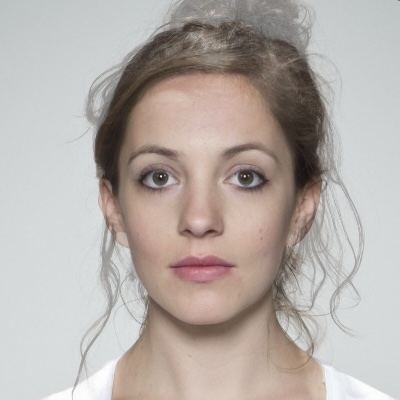}
  \includegraphics[width=0.24\columnwidth]{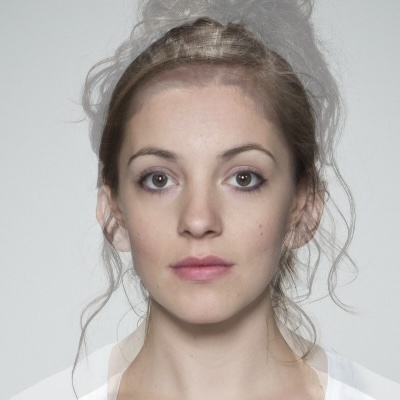}
  \includegraphics[width=0.24\columnwidth]{002_03.jpg} \\

  \includegraphics[width=0.24\columnwidth]{001_03.jpg}
  \includegraphics[width=0.24\columnwidth]{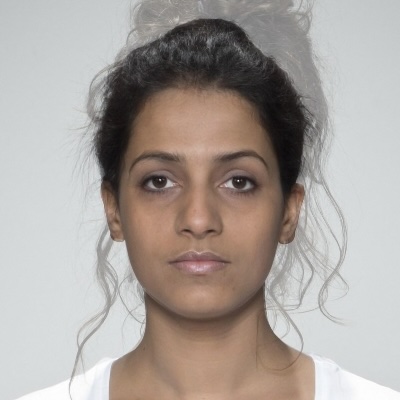}
  \includegraphics[width=0.24\columnwidth]{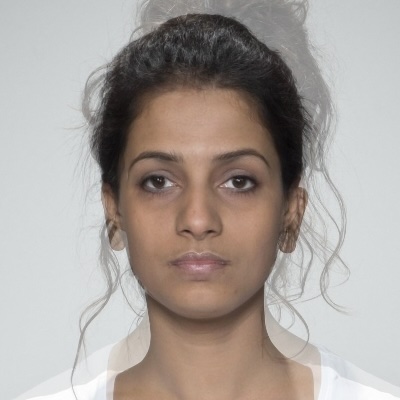}
  \includegraphics[width=0.24\columnwidth]{006_03.jpg}
\vspace{-0.3cm}
  \caption{Morphing with automatic and manually placed landmarks. Left and right columns show I$_0$ and I$_1$. The second column shows a linear blending at $t=0.6$ with manually positioned landmarks in addition to the ones detected by DLib. The third column shows a linear blending after our neural warping using only DLib landmarks. Notice how, besides the face, other areas such as neck and hairline are aligned on the second column, compared to the third.}
  \label{fig:manual-landmarks}
\end{figure}

In addition to automatic landmark detection, we may also manually edit landmarks to constrain other regions during the warping, such as clothing, and hairline. This improves the morphing on areas other than the face. Fig~\ref{fig:manual-landmarks} shows a visual comparison between morphing using only automatic landmark detection and manual landmark adjustments.

\bibliographystyle{abbrvnat}
\bibliography{refs}